\pdfoutput=1

\documentclass[11pt]{article}

\usepackage[final]{acl}

\usepackage{times}
\usepackage{latexsym}

\usepackage[T1]{fontenc}

\usepackage[utf8]{inputenc}

\usepackage{microtype}

\usepackage{inconsolata}

\usepackage{graphicx}

\usepackage{dsfont}

\usepackage{enumitem}
\usepackage{rotating}

\usepackage{booktabs}
\usepackage{multirow}
\usepackage{tcolorbox}
\usepackage{subcaption}

\usepackage[normalem]{ulem}

\usepackage{arydshln}
\makeatletter
\newcommand{\adl@rowsL}{}
\makeatother

\usepackage{amsthm}
\usepackage{amsmath}
\usepackage{amssymb}
\usepackage{xspace}
\newcommand{\narrowtextsc}[1]{\textls[-50]{\textsc{#1}}}
\newcommand{\lm}[1]{\texttt{#1}}
\newcommand{\sys}[1]{\narrowtextsc{#1}}
\newcommand{\data}[1]{\textsf{#1}}

\usepackage{xcolor,colortbl}
\usepackage[dvipsnames]{xcolor}

\usepackage{soul}
\definecolor{lightblue}{RGB}{239,247,250}
\definecolor{blue}{RGB}{224,230,255}
\definecolor{lighpurple}{RGB}{226, 218, 246}
\definecolor{purple}{RGB}{193, 175, 236}
\definecolor{deepred}{RGB}{183,26,26}
\definecolor{deepgreen}{RGB}{4,98,10}
\definecolor{lightred}{RGB}{242,207, 194}
\definecolor{gray}{RGB}{216,216,216}

\usepackage{verbatim}

\usepackage{pifont}
\usepackage{adjustbox}
\usepackage{booktabs,multirow}
\usepackage{makecell}
\usepackage{xcolor}
\definecolor{morandiRed}{RGB}{190,90,90}   
\definecolor{morandiGreen}{RGB}{90,140,120} 
\usepackage{xstring}
\usepackage{comment}
\usepackage{soul}
\definecolor{lightRed}{RGB}{255,200,200}
\sethlcolor{lightRed}
\usepackage{xcolor}

\definecolor{aigcHLBase}{RGB}{190,90,90}   
\colorlet{aigcHL}{aigcHLBase!25}           
\sethlcolor{aigcHL}

\usepackage{fvextra}
\RecustomVerbatimEnvironment{verbatim}{Verbatim}{
  breaklines=true,      
  breakanywhere=true,   
  fontsize=\small,      
  xleftmargin=1em,      
  xrightmargin=1em,
  breaksymbolleft={},   
  breaksymbolright={}
}

\usepackage{tcolorbox}

\usepackage{needspace}
\usepackage{ragged2e}
\usepackage{tabularx}
\usepackage{array}
\usepackage[table]{xcolor}
\definecolor{headergray}{gray}{0.9}

\newcolumntype{L}{>{\raggedright\arraybackslash}p{1.5cm}}
\newcolumntype{Y}{>{\raggedright\arraybackslash}X}

\title{\raisebox{-0.15\height}{\includegraphics[width=0.5cm]{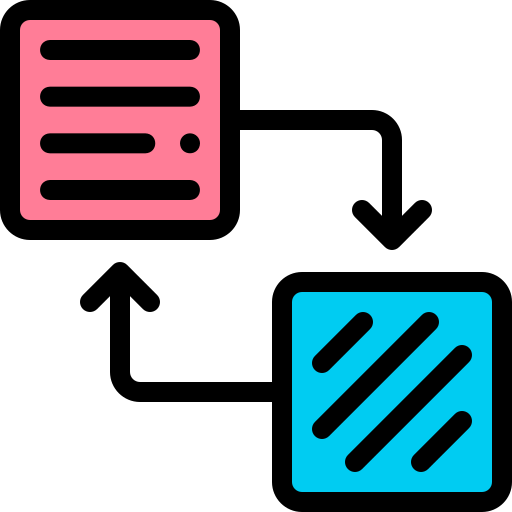}} \textit{i}\texttt{Flip}: \\ Iterative Feedback-driven Counterfactual Example Refinement}

\newcommand{\affilsup}[1]{\rlap{\textsuperscript{\normalfont#1}}}

\author{
    Yilong Wang\affilsup{1,\footnotemark[1]}
    \qquad
    Qianli Wang\affilsup{1,3,\footnotemark[1],\footnotemark[2]}
    \qquad
    Nils Feldhus\affilsup{2,3,\footnotemark[3]}\\
    $^1$Technische Universit\"at Berlin
    \qquad
    $^2$University of Groningen\\
    $^3$German Research Center for Artificial Intelligence (DFKI)
    \\
    \small{\footnotemark[2]\textbf{Correspondence:} \texttt{\href{mailto:qianli.wang@tu-berlin.de}{qianli.wang@tu-berlin.de}}}
}

\begin{document}

\maketitle
\renewcommand{\thefootnote}{\fnsymbol{footnote}}
\footnotetext[1]{Equal contribution and shared first authorship.}
\footnotetext[3]{Work done at TU Berlin.}
\renewcommand*{\thefootnote}{\arabic{footnote}}

\begin{abstract}
Counterfactual examples are minimal edits to an input that alter a model's prediction. They are widely employed in explainable AI to probe model behavior and in natural language processing (NLP) to augment training data. However, generating valid counterfactuals with large language models (LLMs) remains challenging, as existing single-pass methods often fail to induce reliable label changes, neglecting LLMs' self-correction capabilities. To explore this untapped  potential, we propose \textit{i}\texttt{Flip}\footnote{\url{https://github.com/YilongWangBerlin/iFlip}}, an iterative refinement approach that leverages three types of feedback, including model confidence, feature attribution, and natural language. Our results show that \textit{i}\texttt{Flip} achieves a more favorable validity-minimality balance than the three state-of-the-art baselines. The user study further provides complementary evidence that \textit{i}\texttt{Flip} is preferred over the baselines in \textit{completeness}, \textit{overall satisfaction}, and \textit{feasibility}. In addition, ablation studies demonstrate that three components are paramount for \textit{i}\texttt{Flip} to generate valid counterfactuals: leveraging an appropriate number of iterations, pointing to highly attributed words, and early stopping. Finally, counterfactuals generated by \textit{i}\texttt{Flip} enable effective counterfactual data augmentation, substantially improving model performance and robustness.

\end{abstract}

\section{Introduction}
\begin{figure*}[!t]
    \centering
    \includegraphics[width=\textwidth]{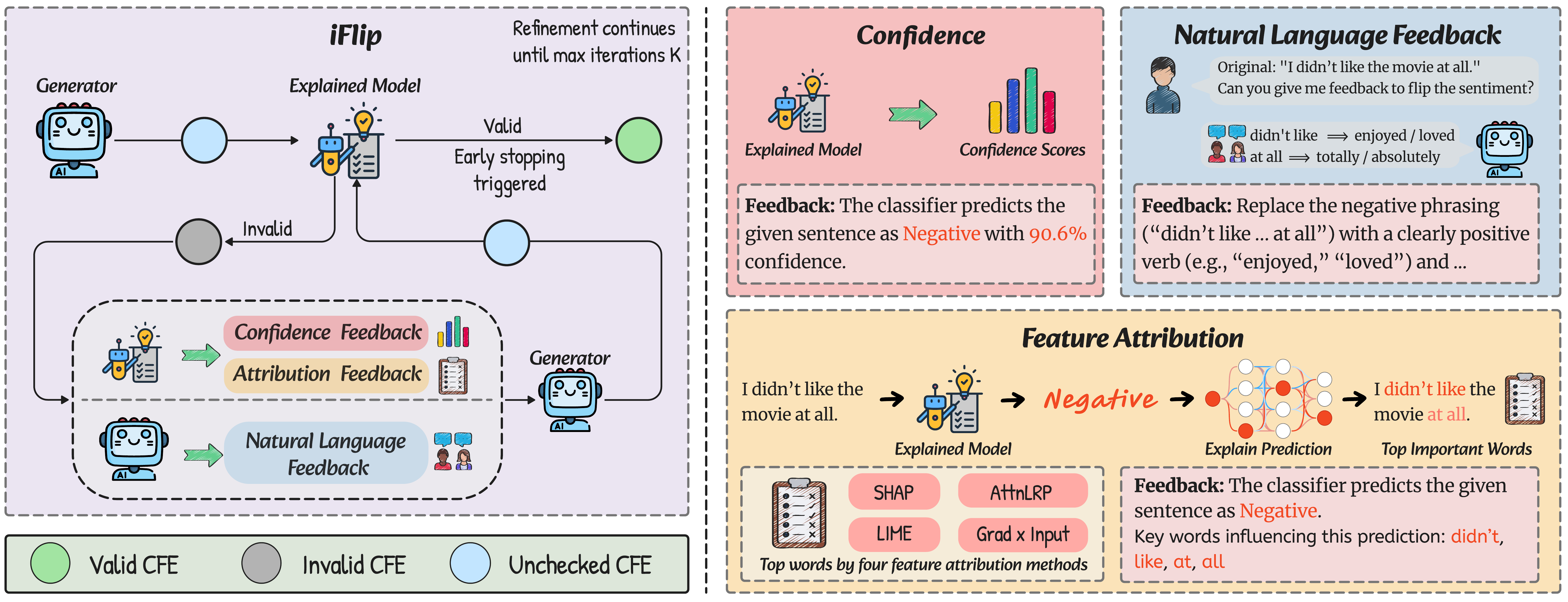}
    \caption{\textit{i}\texttt{Flip} framework overview. \textbf{\textit{Left}}: The core iterative loop of our method, which includes a generator $\mathcal{G}$ and an explained model $\mathcal{M}$. $\mathcal{G}$ generates and refines counterfactuals. 
    $\mathcal{M}$ evaluates counterfactuals under our validity criterion: a counterfactual is considered \emph{valid} if $\mathcal{M}$'s predicted label is equal to the target label $\tilde{y}$. Otherwise, \textit{invalid} counterfactuals are iteratively refined until the validity criterion is met to avoid unnecessary iterations or the maximal number of refinement steps is reached. \textbf{\textit{Right}}: Examples of the three types of feedback we employed for \textit{i}\texttt{Flip}: \textit{Confidence}, \textit{Feature Attribution}, and \textit{Natural Language Feedback}.}    
    \label{fig:aigc}
    
\end{figure*}

Counterfactual examples are minimally modified inputs that cause a model to alter its prediction \cite{madsen-2022-survey, wang-etal-2024-survey, zhao-etal-2024-xai, wang2026macroenhancingmultilingualcounterfactual}. Applications include elucidating opaque LLMs through contrastive causal analysis \cite{ross-etal-2021-explaining, treviso-etal-2023-crest, nguyen-etal-2024-llms, wang-etal-2026-parallel} and enhancing model performance and robustness \cite{kaushik2020learning, dixit-etal-2022-core, qiu-etal-2024-paircfr}. 
However, most recent approaches \cite{bhan-etal-2023-tictec, bhattacharjee-etal-2024-zero, nguyen2025guidingllmsgeneratehighfidelity} rely on single-pass generation, often yielding invalid counterfactuals that are insufficient to shift the model's original prediction. 
This neglects LLMs' inherent self-correction capabilities, despite their proven efficacy in other domains and downstream tasks \cite{madaan2023selfrefine, gou2024critic, rahmani2025selfcorrectinglargelanguagemodels}. To overcome these limitations, we propose \textit{i}\texttt{Flip} (Figure~\ref{fig:aigc}), a framework that iteratively refines counterfactuals via diverse feedback signals to enhance validity. 

\textbf{First}, we evaluate \textit{i}\texttt{Flip} across three datasets and three LLMs using confidence, feature attribution, and natural language feedback. 
Compared to three LLM-based baselines -- CGG~\citep{nguyen2025guidingllmsgeneratehighfidelity}, \sys{FIZLE} \cite{bhattacharjee-etal-2024-zero}, and Causal What-Ifs \citep{10.1007/978-981-95-4367-0_13} -- \textit{i}\texttt{Flip} improves validity by 78.8\% at the cost of 4.95\% lower similarity, with natural language feedback proving most effective.

\textbf{Second}, a user study confirms \textit{i}\texttt{Flip} with natural language feedback is preferred over FIZLE and CGG in \textit{completeness}, \textit{overall satisfaction}, and \textit{feasibility}, achieving the most substantial improvement in \textit{overall satisfaction} (83.21\%). Error analysis reveals that the iterative refinement process effectively addresses both incomplete and contextually inappropriate edits in counterfactuals (\S\ref{subsec:user_study_and_laaj}).


\textbf{Third}, we perform systematic ablation studies to assess the contribution of each component in \textit{i}\texttt{Flip}: (1) the number of refinement iterations, (2) the choice of feedback signal, and (3) the early stopping criteria. We find that counterfactual validity can be noticeably improved across iterations, even within a limited number of rounds (e.g., $\mathcal{K}=5$). Feedback-guided generation considerably boosts performance over the without-feedback baseline. Moreover, we establish that early stopping upon achieving validity is imperative to prevent valid counterfactuals from being erroneously overturned in subsequent iterations.

\textbf{Lastly}, we show that counterfactuals generated by \textit{i}\texttt{Flip} enable effective counterfactual data augmentation (CDA), improving model performance and robustness, measured by accuracy on both the original and the human-annotated counterfactual test sets. 
Additionally, our analysis confirms that the quality of the generated counterfactuals positively correlates with the CDA performance gains.

\section{Related Work}
\paragraph{LLM-based Counterfactual Generation.}
\sys{FIZLE} employs LLMs for zero-shot counterfactual generation to explain and evaluate black-box classifiers \citep{bhattacharjee-etal-2024-zero}. \citet{bhattacharjee2024llmguidedcausalexplainabilityblackbox} propose a three-step pipeline that leverages instruction-tuned LLMs to extract latent features and their associated input words.
Several recent methods incorporate feature attribution signals to guide counterfactual generation.
ZeroCF improves zero-shot counterfactual generation by leveraging feature attribution methods to guide text edits \citep{wang-etal-2025-fitcf}. Similarly, \sys{CGG} inserts token attributions into few-shot prompts to guide LLMs in generating high-validity CFEs \citep{nguyen2025guidingllmsgeneratehighfidelity}.
\citet{10.1007/978-981-95-4367-0_13} propose an agentic framework that iteratively refines counterfactuals through self-reflection but does not incorporate \textit{feedback} or \textit{early stopping}, which we find crucial for enhancing counterfactual validity (\S\ref{subsubsec:feedback}, \S\ref{subsubsec::early_stopping}). 
Unlike these approaches, \textit{i}\texttt{Flip} explicitly incorporates three types of feedback: feature attribution, confidence scores, and natural language feedback, and iteratively refines counterfactuals based on feedback, with early stopping to prevent unnecessary iterations as soon as a valid counterfactual is identified. Both aspects are particularly critical for counterfactual generation, as evidenced by our ablation studies (\S\ref{subsec:ablation}).

\paragraph{Iterative Refinement with Feedback.}
\citet{xi-etal-2023-self} and \citet{liu2024intrinsicselfcorrectioncapabilityllms} empirically observe that model performance tends to improve over successive rounds of self-correction, wherein LLMs refine their answers using either intrinsic \cite{madaan2023selfrefine, xu-etal-2024-sayself} or external \cite{welleck2023generating} feedback. In the field of explainability, \sys{SR-NLE} improves the faithfulness of post-hoc explanations through an iterative critique and refinement process \cite{wang2025selfcritiquerefinementfaithfulnatural}. \sys{Cross-Refine} enhances the quality of natural language explanations via tandem learning through feedback and critiques in an iterative manner \cite{wang-etal-2025-cross}. In this work, we employ three types of feedback signals to guide iterative refinement of generated counterfactuals across multiple rounds.

\section{Methodology}

\subsection{\textit{i}\texttt{Flip}}
\label{subsec:method_pipeline}
As illustrated in Figure~\ref{fig:aigc}, \textit{i}\texttt{Flip} employs iterative refinement driven by the interaction of two components: the counterfactual generator $\mathcal{G}$ and the explained model $\mathcal{M}$. The explained model validates the counterfactual candidates produced by the generator. When a candidate is unsuccessful in changing the model's prediction, it is sent back to the generator, accompanied by specific feedback signals (\S\ref{subsec:feedback}), to guide the next generation step. \textit{i}\texttt{Flip} is structured into three primary stages:


\paragraph{Step 1: Generation.}
Given the original input $x$ and the explained model $\mathcal{M}$, we first obtain the original model prediction $y = \mathcal{M}(x)$, a label in a classification task,
then define a target label $\tilde{y} \neq y$. Generator $\mathcal{G}$ produces an initial counterfactual candidate:
\begin{equation}
\tilde{x}_0 \gets \mathcal{G}(x, y, \tilde{y})
\tag{1}
\end{equation}

\paragraph{Step 2: Verification.}
Explained model $\mathcal{M}$ validates the candidate $\tilde{x}_0$.
If $\mathcal{M}(\tilde{x}_0) = \tilde{y}$, i.e., the desired target label is achieved, we return $\tilde{x}_0$ (early stopping); otherwise, we proceed to \textbf{refinement}.

\paragraph{Step 3: Refinement.}
If the initial counterfactual candidate $\tilde{x}_0$ fails to induce the model prediction to $\tilde{y}$, we perform up to  $\mathcal{K}$ iterative refinement steps, indexed by $k\in\{1, 2,\cdots,\mathcal{K}\}$, each guided by feedback signals $f_k$: 
\begin{equation}
\tilde{x}_k \gets \mathcal{G} (x,\tilde{x}_{k-1}, y, \tilde{y}, f_k)
\tag{2}
\end{equation}
Upon generating a new counterfactual candidate $\tilde{x}_k$, we return to the validation phase (\textbf{Step 2}). This iterative process continues until either the validation succeeds ($\mathcal{M}(\tilde{x}_k) = \tilde{y}$) or the maximum iteration count $\mathcal{K}$ is attained.

\subsection{Feedback Signals}
\label{subsec:feedback}

In our experiments, we cover three feedback signals $ F = \{ f_{\text{Conf}}, f_{\text{Attr}}, f_{\text{NL}} \}$ that are commonly used in the literature: \textit{confidence}, \textit{feature attribution}, and \textit{natural language feedback} (Figure~\ref{fig:aigc}).

\paragraph{Confidence (Conf).} The self-correction capability of language models comprises two components: \textit{confidence}, defined as the ability to retain correct answers, and \textit{critique}, defined as the ability to revise incorrect answers \cite{yang-etal-2025-confidence}. In our framework, we use the explained model's predicted-label probability, i.e., the maximum softmax score across all classes, as a feedback signal for confidence. This choice is lightweight and easy to integrate into iterative refinement. 

\paragraph{Feature Attribution.} Following prior work that established guiding LLMs for counterfactual generation using important words \cite{bhan-etal-2023-tictec, wang-etal-2025-fitcf, nguyen2025guidingllmsgeneratehighfidelity}, we apply established four feature attribution methods: SHAP \cite{lundberg-etal-2017-shap}, AttnLRP \cite{achtibat2024attnlrp}, LIME \cite{ribeiro-etal-2016-lime}, and Gradient × Input   \cite{shrikumar2017justblackboxlearning}, which determine token importance in an input text.

\paragraph{Natural Language Feedback (NL).} LLM-generated feedback in free-text form has been shown to be effective for improving both downstream task performance and explanation generation \cite{madaan2023selfrefine, wang-etal-2025-cross}. Accordingly, we instruct the LLMs to provide natural language feedback on their own counterfactuals.

\paragraph{Combined Feedback.} Beyond these individual signals, we also consider a combined feedback that integrates confidence, SHAP and natural language feedback. We discuss the comparison of individual and combined feedback in \S\ref{sec:automatic_evaluation}.

\section{Experimental Setup}
\label{sec:exp_setup}
\subsection{Datasets \& Models}
\paragraph{Datasets.} We test \textit{i}\texttt{Flip} on three NLP datasets that are widely adopted in the counterfactual generation literature \cite{bhattacharjee2024llmguidedcausalexplainabilityblackbox, wang-etal-2025-fitcf, nguyen2025guidingllmsgeneratehighfidelity}.
\footnote{Dataset examples, label distributions, and dataset sources are provided in Appendix~\ref{app:dataset}.} \data{IMDb} comprises diverse movie reviews labeled as \textit{positive} or \textit{negative} sentiment \cite{maas-etal-2011-imdb}. \data{\data{AG News}} contains news articles categorized into four topics: \textit{World}, \textit{Sports}, Business, and \textit{Sci/Tech} \cite{zhang-2015-agnews}. Each article includes a title and a description. \data{SNLI} is a dataset for natural language inference (NLI) tasks \cite{bowman-etal-2015-large}. It consists of premise-hypothesis pairs labeled as \textit{entailment}, \textit{contradiction}, or \textit{neutral}. \looseness=-1

\paragraph{Models.} We employ three widely used open-source LLMs from different model families and with increasing parameter sizes: \lm{OLMo2-7B} \cite{olmo20252olmo2furious}, \lm{Qwen3-32B} \cite{yang2025qwen3technicalreport}, \lm{\lm{LLaMA3.3-70B}} \cite{grattafiori2024llama3herdmodels}.\footnote{Information about inference time and generation parameters is provided in Appendix~\ref{app:setting}.}

\subsection{Baselines}
\label{subsec:baseline}
We compare \textit{i}\texttt{Flip} against three state-of-the-art LLM-based counterfactual generation methods. 
\ding{192} CGG uses feature attributions to select the top 25\% most important words 
and incorporates them into a predefined few-shot prompt, explicitly directing the LLM which words to modify for the target label flip~\citep{nguyen2025guidingllmsgeneratehighfidelity}. \ding{193}~\sys{FIZLE}$_{\text{naive}}$ directly prompts an LLM to minimally edit the input to achieve a label flip, without prior feature identification or guidance \cite{bhattacharjee-etal-2024-zero}.
\ding{194} Causal What-Ifs is an agentic approach that iteratively improves counterfactual generation through self-reflection, enabling the identification of causally consistent edits \textit{without external feedback} \cite{10.1007/978-981-95-4367-0_13}. 

\section{Evaluation}
\subsection{Automatic Evaluation}
\label{subsec:metrics}
To assess the quality of generated counterfactuals, we adopt three commonly used automatic evaluation metrics following prior work \cite{ross-etal-2021-explaining, wang-etal-2024-survey, nguyen-etal-2024-ceval-benchmark, bhattacharjee-etal-2024-zero}: \textit{validity}, \textit{minimality}, and \textit{fluency}.

\paragraph{Label Flipping Rate (LFR)} evaluates the \emph{validity} of counterfactuals by measuring the proportion of instances whose predicted label changes after modification. Given the explained model $\mathcal{M}$\footnote{Model $\mathcal{M}$ is \lm{BERT} in Table~\ref{tab:main_results} and \lm{RoBERTa} in Appendix~\ref{subsec::roberta_results}.}
, the original inputs $x$, and counterfactuals $\tilde{x}$, LFR is calculated by:
{
\setlength{\abovedisplayskip}{5pt}
\setlength{\belowdisplayskip}{5pt}
\setlength{\abovedisplayshortskip}{5pt}
\setlength{\belowdisplayshortskip}{5pt}
\[
{\small
\mathrm{LFR}=\frac{1}{N}\sum_{i=1}^{N}\mathbb{I}\!\left[\mathcal{M}(x_i)\neq \mathcal{M}(\tilde{x}_i)\right]
}
\]
}
where $\mathbb{I}[\cdot]$ denotes the indicator function.


\paragraph{Semantic Similarity (SS)} \textit{Minimality} can be assessed by semantic similarity between the original input $x$ and its counterfactual $\tilde{x}$ using cosine similarity between their sentence embeddings $\mathbf{e}(\cdot)$, where each embedding is produced by a pretrained encoder\footnote{\href{https://huggingface.co/sentence-transformers/all-MiniLM-L6-v2}{\texttt{sentence-transformers/all-MiniLM-L6-v2}}}:
{
\setlength{\abovedisplayskip}{5pt}
\setlength{\belowdisplayskip}{5pt}
\setlength{\abovedisplayshortskip}{5pt}
\setlength{\belowdisplayshortskip}{5pt}
\[
{\small
\mathrm{SS}(x_i, \tilde{x}_i)=
\frac{\mathbf{e}(x_i)^\top \mathbf{e}(\tilde{x}_i)}
{\|\mathbf{e}(x_i)\|_2\,\|\mathbf{e}(\tilde{x}_i)\|_2}
}
\]
}

\paragraph{Perplexity (PPL)}
To assess \emph{fluency}, we compute the perplexity of generated counterfactuals using \lm{GPT-2}.  
Given a counterfactual $\tilde{x} = (t_1, \ldots, t_n)$ consisting of $n$ tokens and a model parameterized by $\theta$, the perplexity of $\tilde{x}$ is defined as:
{
\setlength{\abovedisplayskip}{5pt}
\setlength{\belowdisplayskip}{5pt}
\setlength{\abovedisplayshortskip}{5pt}
\setlength{\belowdisplayshortskip}{5pt}
\[
{\small
\mathrm{PPL}(\tilde{x}) = \exp\!\left(-\frac{1}{n}\sum_{i=1}^{n} \log p_{\theta}(t_i \mid t_{<i})\right)
}
\]
}

\subsection{User Study \& LLM-as-a-Judge}
We further assess the quality of generated counterfactuals by conducting a \textit{user study} and a complementary \textit{LLM-as-a-Judge evaluation}. Participants and judge models, respectively, evaluate the counterfactuals along three dimensions (\S\ref{subsubsec:subjective_rating}) on a 6-point Likert scale.

\begin{table*}[t!]
    \centering
    \renewcommand*{\arraystretch}{1}

    \footnotesize
    \resizebox{\textwidth}{!}{%
        \begin{tabular}{ll|ccc|ccc|ccc|ccc}
            \toprule
            \textbf{} & \textbf{Method} 
            & \multicolumn{3}{c|}{\textbf{\data{IMDb}}}
            & \multicolumn{3}{c|}{\textbf{\data{AG News}}}
            & \multicolumn{3}{c|}{\makecell{\textbf{\data{SNLI}} \textbf{(Premise)}}}
            & \multicolumn{3}{c}{\makecell{\textbf{\data{SNLI}} \textbf{(Hypothesis)}}}\\
            \cmidrule(lr){3-5}\cmidrule(lr){6-8}\cmidrule(lr){9-11}\cmidrule(lr){12-14}
             & & LFR$\uparrow$ & SS$\uparrow$ & PPL$\downarrow$
               & LFR$\uparrow$ & SS$\uparrow$ & PPL$\downarrow$
               & LFR$\uparrow$ & SS$\uparrow$ & PPL$\downarrow$
               & LFR$\uparrow$ & SS$\uparrow$ & PPL$\downarrow$
            \\
            \midrule
            \multirow{9}{*}[-0.4cm]{\rotatebox[origin=c]{90}{\lm{OLMo2-7B}}}
            & \cellcolor[HTML]{d8d8d8}{CGG}               
            & \cellcolor[HTML]{d8d8d8}0.885 & \cellcolor[HTML]{d8d8d8}\textbf{0.837} & \cellcolor[HTML]{d8d8d8}44.67
            & \cellcolor[HTML]{d8d8d8}0.510 & \cellcolor[HTML]{d8d8d8}0.380 & \cellcolor[HTML]{d8d8d8}208.56
            & \cellcolor[HTML]{d8d8d8}0.169 & \cellcolor[HTML]{d8d8d8}\textbf{0.882} & \cellcolor[HTML]{d8d8d8}57.03
            & \cellcolor[HTML]{d8d8d8}0.253 & \cellcolor[HTML]{d8d8d8}\textbf{0.948} & \cellcolor[HTML]{d8d8d8}53.92 \\
            & \cellcolor[HTML]{d8d8d8}{$\text{FIZLE}$}      
            & \cellcolor[HTML]{d8d8d8}0.585 & \cellcolor[HTML]{d8d8d8}0.584 & \cellcolor[HTML]{d8d8d8}38.71
            & \cellcolor[HTML]{d8d8d8}0.308 & \cellcolor[HTML]{d8d8d8}0.584 & \cellcolor[HTML]{d8d8d8}77.38
            & \cellcolor[HTML]{d8d8d8}0.305 & \cellcolor[HTML]{d8d8d8}0.730 & \cellcolor[HTML]{d8d8d8}173.94
            & \cellcolor[HTML]{d8d8d8}0.404 & \cellcolor[HTML]{d8d8d8}0.882 & \cellcolor[HTML]{d8d8d8}74.15 \\
            & \cellcolor[HTML]{d8d8d8}{Causal What-Ifs} 
            & \cellcolor[HTML]{d8d8d8}0.890 & \cellcolor[HTML]{d8d8d8}0.801 & \cellcolor[HTML]{d8d8d8}37.97
            & \cellcolor[HTML]{d8d8d8}0.400 & \cellcolor[HTML]{d8d8d8}\textbf{0.607} & \cellcolor[HTML]{d8d8d8}\textbf{37.54}
            & \cellcolor[HTML]{d8d8d8}0.413 & \cellcolor[HTML]{d8d8d8}0.854 & \cellcolor[HTML]{d8d8d8}\textbf{47.27}
            & \cellcolor[HTML]{d8d8d8}0.487 & \cellcolor[HTML]{d8d8d8}0.911 & \cellcolor[HTML]{d8d8d8}\textbf{35.56} \\
            \cmidrule(lr){2-14}
            & \cellcolor[HTML]{D9EAFD}\textit{i}\texttt{Flip}-Conf              
            & 0.964 & 0.809 & 41.19
            & 0.784 & 0.503 & \uwave{41.41}
            & \textbf{\uwave{0.660}} & 0.854 & \uwave{53.72}
            & \textbf{\uwave{0.826}} & 0.870 & 40.65 \\
            & \cellcolor[HTML]{E1F5E1}\textit{i}\texttt{Flip}-SHAP              
            & 0.968 & 0.805 & 41.87
            & \textbf{\uwave{0.802}} & 0.488 & 42.95
            & 0.587 & 0.847 & 56.81
            & 0.760 & 0.867 & 42.39 \\
            & \cellcolor[HTML]{E1F5E1}\textit{i}\texttt{Flip}-AttnLRP           
            & 0.978 & 0.803 & 42.61
            & 0.768 & 0.500 & 46.61
            & 0.596 & 0.851 & 55.30
            & 0.810 & 0.869 & \uwave{40.31} \\
            & \cellcolor[HTML]{E1F5E1}\textit{i}\texttt{Flip}-Grad$\times$Input 
            & 0.976 & 0.801 & 43.30
            & 0.773 & 0.498 & 50.32
            & 0.624 & 0.854 & 56.02
            & 0.782 & 0.871 & 41.81 \\
            & \cellcolor[HTML]{E1F5E1}\textit{i}\texttt{Flip}-LIME              
            & 0.968 & 0.805 & 52.82
            & 0.730 & 0.504 & 44.32
            & 0.558 & 0.866 & 58.18
            & 0.736 & 0.875 & 41.02 \\
            & \cellcolor[HTML]{FFEACC}\textit{i}\texttt{Flip}-NL 
            & 0.984 & \uwave{0.822} & 36.47
            & 0.735 & \uwave{0.515} & 44.73
            & 0.638 & \uwave{0.879} & 148.41
            & 0.770 & \uwave{0.901} & 43.81 \\
            & \cellcolor[HTML]{F4E1F9}\textit{i}\texttt{Flip}-Combined
            & \textbf{\uwave{0.985}} & 0.810 & \textbf{\uwave{32.50}}
            & 0.745 & 0.477 & 55.21
            & 0.640 & 0.873 & 66.19
            & 0.775 & 0.886 & 43.59 \\
            \midrule
            \multirow{9}{*}[-0.4cm]{\rotatebox[origin=c]{90}{\lm{Qwen3-32B}}}
            & \cellcolor[HTML]{d8d8d8}{CGG}               
            & \cellcolor[HTML]{d8d8d8}0.863 & \cellcolor[HTML]{d8d8d8}0.877 & \cellcolor[HTML]{d8d8d8}62.47
            & \cellcolor[HTML]{d8d8d8}0.364 & \cellcolor[HTML]{d8d8d8}0.697 & \cellcolor[HTML]{d8d8d8}97.10
            & \cellcolor[HTML]{d8d8d8}0.200 & \cellcolor[HTML]{d8d8d8}0.881 & \cellcolor[HTML]{d8d8d8}50.31
            & \cellcolor[HTML]{d8d8d8}0.283 & \cellcolor[HTML]{d8d8d8}\textbf{0.938} & \cellcolor[HTML]{d8d8d8}47.94 \\
            & \cellcolor[HTML]{d8d8d8}{$\text{FIZLE}$}      
            & \cellcolor[HTML]{d8d8d8}0.684 & \cellcolor[HTML]{d8d8d8}\textbf{0.900} & \cellcolor[HTML]{d8d8d8}36.00
            & \cellcolor[HTML]{d8d8d8}0.164 & \cellcolor[HTML]{d8d8d8}\textbf{0.864} & \cellcolor[HTML]{d8d8d8}61.46
            & \cellcolor[HTML]{d8d8d8}0.454 & \cellcolor[HTML]{d8d8d8}0.901 & \cellcolor[HTML]{d8d8d8}38.66
            & \cellcolor[HTML]{d8d8d8}0.524 & \cellcolor[HTML]{d8d8d8}0.918 & \cellcolor[HTML]{d8d8d8}35.97 \\
            & \cellcolor[HTML]{d8d8d8}{Causal What-Ifs}  
            & \cellcolor[HTML]{d8d8d8}0.848 & \cellcolor[HTML]{d8d8d8}0.821 & \cellcolor[HTML]{d8d8d8}42.03
            & \cellcolor[HTML]{d8d8d8}0.436 & \cellcolor[HTML]{d8d8d8}0.570 & \cellcolor[HTML]{d8d8d8}53.50
            & \cellcolor[HTML]{d8d8d8}0.482 & \cellcolor[HTML]{d8d8d8}0.842 & \cellcolor[HTML]{d8d8d8}\textbf{36.14}
            & \cellcolor[HTML]{d8d8d8}0.518 & \cellcolor[HTML]{d8d8d8}0.920 & \cellcolor[HTML]{d8d8d8}\textbf{34.18}\\
            \cmidrule(lr){2-14}
            & \cellcolor[HTML]{D9EAFD}\textit{i}\texttt{Flip}-Conf              
            & 0.996 & 0.855 & 33.30
            & 0.726 & 0.521 & \textbf{\uwave{41.59}}
            & 0.466 & 0.895 & 62.69
            & 0.562 & 0.871 & 47.03 \\
            & \cellcolor[HTML]{E1F5E1}\textit{i}\texttt{Flip}-SHAP              
            & 0.994 & 0.857 & 33.70
            & 0.736 & 0.519 & 44.02
            & 0.452 & 0.878 & 58.60
            & 0.586 & 0.866 & 45.04 \\
            & \cellcolor[HTML]{E1F5E1}\textit{i}\texttt{Flip}-AttnLRP           
            & \textbf{\uwave{1.000}} & 0.858 & 33.19
            & 0.746 & 0.513 & 44.15
            & 0.448 & 0.882 & \uwave{58.41}
            & 0.610 & 0.867 & 44.77 \\
            & \cellcolor[HTML]{E1F5E1}\textit{i}\texttt{Flip}-Grad$\times$Input 
            & 0.996 & 0.865 & 33.65
            & 0.730 & 0.522 & 44.33
            & 0.440 & 0.881 & 58.66
            & 0.542 & 0.865 & 45.60 \\
            & \cellcolor[HTML]{E1F5E1}\textit{i}\texttt{Flip}-LIME              
            & 0.998 & 0.863 & \textbf{\uwave{32.57}}
            & 0.742 & 0.527 & 44.52
            & 0.458 & 0.891 & 62.98
            & 0.570 & 0.870 & \uwave{44.22} \\
            & \cellcolor[HTML]{FFEACC}\textit{i}\texttt{Flip}-NL 
            & 0.980 & \uwave{0.878} & 32.99
            & 0.732 & \uwave{0.536} & 49.68
            & \textbf{\uwave{0.596}} & 0.895 & 68.16
            & \textbf{\uwave{0.754}} & 0.906 & 47.20 \\
            & \cellcolor[HTML]{F4E1F9}\textit{i}\texttt{Flip}-Combined
            & 0.970 & 0.868 & 32.66
            & \textbf{\uwave{0.975}} & 0.465 & 66.36
            & 0.512 & \textbf{\uwave{0.916}} & 58.33
            & 0.607 & \uwave{0.932} & \uwave{42.81} \\
            \midrule
            \multirow{9}{*}[-0.3cm]{\rotatebox[origin=c]{90}{\lm{LLaMA3.3-70B}}}
            & \cellcolor[HTML]{d8d8d8}{CGG}               
            & \cellcolor[HTML]{d8d8d8}0.887 & \cellcolor[HTML]{d8d8d8}0.869 & \cellcolor[HTML]{d8d8d8}52.81
            & \cellcolor[HTML]{d8d8d8}0.526 & \cellcolor[HTML]{d8d8d8}0.694 & \cellcolor[HTML]{d8d8d8}122.40
            & \cellcolor[HTML]{d8d8d8}0.307 & \cellcolor[HTML]{d8d8d8}0.846 & \cellcolor[HTML]{d8d8d8}51.62
            & \cellcolor[HTML]{d8d8d8}0.271 & \cellcolor[HTML]{d8d8d8}\textbf{0.937} & \cellcolor[HTML]{d8d8d8}46.58 \\
            & \cellcolor[HTML]{d8d8d8}{$\text{FIZLE}$} 
            & \cellcolor[HTML]{d8d8d8}0.924 & \cellcolor[HTML]{d8d8d8}0.868 & \cellcolor[HTML]{d8d8d8}34.88
            & \cellcolor[HTML]{d8d8d8}0.416 & \cellcolor[HTML]{d8d8d8}\textbf{0.707} & \cellcolor[HTML]{d8d8d8}56.59
            & \cellcolor[HTML]{d8d8d8}0.402 & \cellcolor[HTML]{d8d8d8}\textbf{0.903} & \cellcolor[HTML]{d8d8d8}41.54
            & \cellcolor[HTML]{d8d8d8}0.534 & \cellcolor[HTML]{d8d8d8}0.929 & \cellcolor[HTML]{d8d8d8}42.39 \\
            & \cellcolor[HTML]{d8d8d8}{Causal What-Ifs} 
            & \cellcolor[HTML]{d8d8d8}0.922 & \cellcolor[HTML]{d8d8d8}0.788 & \cellcolor[HTML]{d8d8d8}\textbf{31.93}
            & \cellcolor[HTML]{d8d8d8}0.636 & \cellcolor[HTML]{d8d8d8}0.461 & \cellcolor[HTML]{d8d8d8}\textbf{43.56}
            & \cellcolor[HTML]{d8d8d8}0.372 & \cellcolor[HTML]{d8d8d8}0.816 & \cellcolor[HTML]{d8d8d8}\textbf{38.97}
            & \cellcolor[HTML]{d8d8d8}0.420 & \cellcolor[HTML]{d8d8d8}0.904 & \cellcolor[HTML]{d8d8d8}41.29 \\
            \cmidrule(lr){2-14}
            &\cellcolor[HTML]{D9EAFD}\textit{i}\texttt{Flip}-Conf              
            & 0.996 & 0.877 & 34.74
            & 0.890 & 0.468 & 44.38
            & 0.580 & 0.763 & 58.65
            & 0.642 & 0.843 & 43.88 \\
            & \cellcolor[HTML]{E1F5E1}\textit{i}\texttt{Flip}-SHAP              
            & 0.996 & 0.878 & \uwave{34.32}
            & 0.860 & 0.481 & 46.07
            & 0.522 & 0.743 & 57.62
            & 0.606 & 0.828 & 41.09 \\
            & \cellcolor[HTML]{E1F5E1}\textit{i}\texttt{Flip}-AttnLRP           
            & 0.992 & 0.879 & 34.58
            & 0.858 & 0.477 & 45.78
            & 0.528 & 0.743 & 56.42
            & 0.602 & 0.822 & 39.92 \\
            & \cellcolor[HTML]{E1F5E1}\textit{i}\texttt{Flip}-Grad$\times$Input 
            & 0.996 & 0.878 & 34.80
            & 0.852 & 0.479 & 46.12
            & 0.580 & 0.751 & \uwave{55.38}
            & 0.603 & 0.825 & 39.59 \\
            & \cellcolor[HTML]{E1F5E1}\textit{i}\texttt{Flip}-LIME              
            & 0.996 & 0.877 & 34.54
            & 0.852 & 0.475 & \uwave{44.20}
            & 0.558 & 0.759 & 57.87
            & 0.608 & 0.825 & 42.91 \\
            & \cellcolor[HTML]{FFEACC}\textit{i}\texttt{Flip}-NL 
            & 0.996 & 0.895 & 36.90
            & 0.900 & \uwave{0.527} & 63.86
            & 0.566 & 0.821 & 69.45
            & \textbf{\uwave{0.680}} & 0.861 & 69.45 \\
            & \cellcolor[HTML]{F4E1F9}\textit{i}\texttt{Flip}-Combined
            & \textbf{\uwave{1.000}} & \textbf{\uwave{0.896}} & 34.48
            & \textbf{\uwave{0.975}} & 0.460 & 66.33
            & \textbf{\uwave{0.592}} & \uwave{0.858} & 59.17
            & 0.602 & \uwave{0.931} & \textbf{\uwave{39.48}} \\
            \bottomrule
        \end{tabular}
    }

    \caption{
        Automatic evaluation results of counterfactuals generated by baselines {\setlength{\fboxsep}{1pt}\colorbox{gray}{(CGG, FIZLE, Causal What-Ifs)}} and our \textit{i}\texttt{Flip} methods on the \lm{BERT} models with different feedback:
        \ding{182} confidence (\mbox{{\setlength{\fboxsep}{1pt}\colorbox[HTML]{D9EAFD}{\strut\textsf{Conf}}}}),
        \ding{183} feature attribution (\mbox{{\setlength{\fboxsep}{1pt}\colorbox[HTML]{E1F5E1}{\strut\textsf{SHAP}}}}, \mbox{{\setlength{\fboxsep}{1pt}\colorbox[HTML]{E1F5E1}{\strut\textsf{AttnLRP}}}}, \mbox{{\setlength{\fboxsep}{1pt}\colorbox[HTML]{E1F5E1}{\strut\textsf{Grad$\times$Input}}}}, \mbox{{\setlength{\fboxsep}{1pt}\colorbox[HTML]{E1F5E1}{\strut\textsf{LIME}}}}),
        \ding{184} natural language (\mbox{{\setlength{\fboxsep}{1pt}\colorbox[HTML]{FFEACC}{\strut\textsf{NL}}}}), and
    \ding{185} combined feedback (\mbox{{\setlength{\fboxsep}{1pt}\colorbox[HTML]{F4E1F9}{\strut\textsf{Combined}}}}), which integrates confidence, SHAP, and natural language feedback.
        Results are reported on \data{IMDb}, \data{AG News}, and \data{SNLI}
        using Label Flipping Rate (LFR), Semantic Similarity (SS), and Perplexity (PPL).
        \uwave{Wavy underline} indicates the best feedback type within \textit{i}\texttt{Flip}. \textbf{Boldface} indicates the best result across methods.
    }
    \label{tab:main_results}
\end{table*}


\subsubsection{Subjective Ratings}
\label{subsubsec:subjective_rating}
Following \citet{nguyen-etal-2024-ceval-benchmark, domnich2024unifyingevaluationcounterfactualexplanations}, we ask human annotators to evaluate counterfactuals across three subjective dimensions:
\begin{itemize}
    \item \textbf{Completeness}: The counterfactual sufficiently explains the model's decision.



    \item \textbf{Overall Satisfaction}: The counterfactual effectively shows how to reach a different outcome.

    \item \textbf{Feasibility}: The counterfactual provides actions that are practical, realistic, and actionable.

\end{itemize} 

\subsubsection{User Study Setup}
\label{subsubsec:user_study_setup}
We conduct a user study on \data{AG News}, from which we randomly sample 10 examples, with ($N=3$) participants, all proficient in English.\footnote{The annotation guideline is detailed in Appendix~\ref{sec:user_study_guidelines}.}
We evaluate two LLM-based baselines (FIZLE and CGG) alongside \textit{i}\texttt{Flip} configured with natural language feedback, which demonstrates the best performance among all feedback types (\S\ref{sec:automatic_evaluation}). In Table \ref{tab:sub_evaluation}, We report inter-annotator and inter-judge-model agreements with Krippendorff’s $\alpha$ for each dimension. \looseness=-1

\subsubsection{LLM-as-a-Judge Setup}
LLM-as-a-Judge (LaaJ) has become a widely adopted approach for conducting evaluations by assigning quality scores that align with human judgment \cite{huang-etal-2024-chatgpt, li-etal-2025-generation}. Following \citet{gu2025surveyllmasajudge}, we employ three commonly used open-source LLMs of varying sizes: \lm{Gemma3-27B} \cite{gemmateam2025gemma3technicalreport}, \lm{GPT-OSS-120B} \cite{openai2025gptoss120bgptoss20bmodel}, and \lm{DeepSeek-R1} \cite{deepseekai2025deepseekr1incentivizingreasoningcapability}. Judge models are instructed to assess the quality of generated counterfactuals based on three subjective dimensions, as in the user study (\S\ref{subsubsec:user_study_setup}).\footnote{The full judge prompt is provided in Appendix~\ref{subsec::prompt_llm:judge}.}

\subsection{Ablation Study Setup}
As outlined in \S\ref{subsec:method_pipeline}, our framework incorporates three essential components: (1) the number of refinement iteration steps, (2) feedback signals, and (3) early stopping. We conduct ablation studies to assess the contribution of each component.

\paragraph{The Number of Refinement Iterations.}
We study whether iterative refinement improves counterfactual generation. Following the intuition of \textit{pass@k}~\citep{brown2024largelanguagemonkeysscaling}, we propose flip@k, which is defined as follows: 
{\setlength{\abovedisplayskip}{4pt}
\setlength{\belowdisplayskip}{3pt}
\[
\text{flip@k}
=
\frac{1}{N}\sum_{i=1}^{N}\mathbb{I}\!\left(\sum_{t=1}^{k} z_i^{(t)} \ge 1\right),
\]
}where \(z_i^{(t)}\) indicates whether the counterfactual at step \(t\) successfully flips the label. Therefore, flip@k captures the fraction of instances that achieve at least one successful label flip within \(k\) iterations.



\paragraph{Feedback Signal.} We analyze the role of different feedback signals in counterfactual generation by considering three feedback settings: (\textit{i}) the baseline \emph{no feedback}, where the model refines without any external signal; (\textit{ii}) a \emph{random feedback}, where the important features are randomly selected;
(\textit{iii}) \emph{least-attributed feedback}, where feedback is derived from the least important words, identified by the corresponding attribution methods. \looseness=-1

\paragraph{Without Early Stopping.}
We investigate the impact of early stopping on the counterfactual refinement. 
Without it, the model may further refine counterfactuals with unnecessary edits even after they become valid, across $\mathcal{K}$ iterations, potentially undermining the minimality objective. 

\subsection{Counterfactual Data Augmentation}
We examine to what extent counterfactuals generated by \textit{i}\texttt{Flip} enhance both model performance and robustness through counterfactual data augmentation \cite{kaushik2020learning, dixit-etal-2022-core, wang2025truthtwistoptimalmodel}. Our baseline is a model $\mathcal{M}_{\text{base}}$ which is trained only on the original dataset $\mathcal{D}_{\text{base}} = \{(x_i,y_i)\}_{i=1}^N$, while the CDA-enhanced model $\mathcal{M}_c$ is trained on both the original data and the corresponding counterfactuals (either generated by \textit{i}\texttt{Flip} or human-annotated) $\mathcal{D}_{c} = \{(x_i,y_i), (\tilde{x}_i, \hat{y_i})\}_{i=1}^N$. The CDA evaluation involves assessment using both the test set data and human-annotated counterfactuals.


\section{Results}

\subsection{Automatic Evaluation}

\label{sec:automatic_evaluation}

\paragraph{Comparison among the baselines.}
Table~\ref{tab:main_results} shows that 
among the three LLM-based baselines, CGG, which leverages feature importance methods to identify the most important words, demonstrates higher LFR on \data{IMDb} and \data{AG News} than \text{FIZLE}, where such word-level guidance facilitates more effective label flips. In contrast, \text{FIZLE}, which generates counterfactuals through direct prompting, performs better on \data{SNLI}, likely because for NLI tasks, modifying important words identified by feature importance methods often fails to fully capture the underlying logical relations between the premise and the hypothesis. Notably, Causal What-Ifs attains the highest validity among baselines on NLI tasks, though it struggles to outperform attribution-guided methods (CGG) on news topic classification (\data{AG News}). However, Causal What-Ifs requires noticeably more edits compared to FIZLE and CGG, suggesting a trade-off between validity gains and input modifications. \looseness=-1

\paragraph{\textit{i}\texttt{Flip} outperforms all baselines in terms of LFR across datasets and models.}
Table~\ref{tab:main_results} illustrates that \textit{i}\texttt{Flip} consistently achieves the highest LFR across all datasets and models while maintaining competitive fluency.\footnote{Appendix~\ref{subsec::transferability} reports the transferability of generated counterfactuals across architectures and the transferability of feedback extracted from \lm{BERT}.}
 Averaged across all experimental settings, \textit{i}\texttt{Flip} improves LFR by 98.4\% and fluency by 10.8\% relative to single-pass LLM-based baselines. These gains are particularly pronounced on \data{AG News}, where \textit{i}\texttt{Flip} achieves a 157.5\% higher LFR than those baselines. Compared with self-reflection-based Causal What-Ifs, \textit{i}\texttt{Flip} further improves validity by 52.8\% on average. This additional gain further validates the effectiveness of the feedback mechanism (\S\ref{subsubsec:feedback}) and the early stopping strategy (\S\ref{subsubsec::early_stopping}).

\paragraph{\textit{i}\texttt{Flip} achieves a more favorable balance between validity and minimality.} 
Table~\ref{tab:main_results} shows an inherent trade-off between validity and minimality, consistent with prior works \citep{mayne-etal-2025-llms}. 
Consequently, while validity improvements inevitably involve additional edits, \textit{i}\texttt{Flip} shows a more favorable balance between validity and minimality than the baselines, rather than simply maximizing validity.\footnote{See Appendix~\ref{app:tradeoff_ts_lfr} for a visualization of SS-LFR trade-off. Appendix~\ref{app:analysis_iter_rounds1} and~\ref{app:analysis_iter_rounds2} report the average number of refinements and the average length of counterfactuals at each iteration.}
On average, \textit{i}\texttt{Flip} achieves a substantial validity gain (+78.8\%) with only a moderate similarity degradation (-4.95\%). Figure~\ref{fig:tradeoff} further shows that \textit{i}\texttt{Flip} more frequently occupies the upper-right region of the SS--LFR trade-off Pareto Front, especially in the \textit{i}\texttt{Flip}-NL configuration.

\paragraph{Comparison of different feedback signals.} 
\label{par:feedback_signal_comparison_6.1}
As shown in Figure \ref{fig:feedback_fr_per_model}, NL feedback generally achieves the most (consistent) enhancement among individual feedback signals in terms of LFR, while the combined feedback achieves the best average LFR overall. As further analyzed in Appendix~\ref{subsec:ablation_nl_feedback}, the effectiveness of NL feedback stems from both its informativeness and its ability to provide actionable refinement guidance. However, these gains come at the cost of substantially higher inference time due to the LLM-generated NL feedback \cite{madaan2023selfrefine} (Table~\ref{tab:inference_time}). Among individual signals, confidence-based feedback ranks second, achieving a balance between effectiveness and computational efficiency. In contrast, attribution-based feedback generally performs poorly, likely because LLMs neither strictly adhere to the identified important words nor limit themselves to targeted modifications, instead often rephrasing entire sentences.\footnote{Representative examples are presented in Appendix~\ref{app:additional_examples}.} 

\paragraph{Case Study.} We examine \data{SNLI} more closely, where generating valid counterfactuals is particularly challenging due to the inherent logical relationships between premises and hypotheses, making it an outlier relative to \data{IMDb} and \data{AG News}. 
Following prior work on counterfactual generation for NLI \citep{nguyen-etal-2024-llms,nguyen2025guidingllmsgeneratehighfidelity}, 
we generate counterfactuals by modifying either the premise or hypothesis. For \data{SNLI}, editing hypotheses is substantially more effective, as their shorter length allows \textit{i}\texttt{Flip} to induce label flips more easily.
Empirically, however, we observe that modifying important words is less effective in improving LFR compared to other ways.
Among \textit{i}\texttt{Flip} variants, attribution-based feedback yields lower LFR than other feedback (Figure~\ref{fig:feedback_fr_per_task}, Appendix~\ref{app:feedback_analysis}), suggesting that targeting only salient words is insufficient to reverse entailment relations in NLI tasks.
Moreover, \data{SNLI} exhibits both substantially lower feature attribution faithfulness than \data{IMDb} and \data{AG News}, and weaker alignment between faithfulness and counterfactual quality.\footnote{Appendix~\ref{sec:faithfulness} reports the faithfulness of the employed attribution methods, while Appendix~\ref{subsec:corr_faith_subsec} details the correlation between feature attribution faithfulness and counterfactual quality.}

{%

\begin{table}[t!]
    \centering
    \footnotesize
    \renewcommand{\arraystretch}{1}
    
    \begin{subtable}{\linewidth}
          \resizebox{\columnwidth}{!}{%
        \centering
    
    \begin{tabular}{l|ccc}
        \toprule
        \textbf{Method} 
        & \textbf{Complet.} 
        & \textbf{Overall Sat.}
        & \textbf{Feasib.} \\
        
        \midrule
        \cellcolor[HTML]{d8d8d8}CGG 
        & \cellcolor[HTML]{d8d8d8} 2.83 $\pm$ 1.69 
        & \cellcolor[HTML]{d8d8d8} 2.67 $\pm$ 1.49
        & \cellcolor[HTML]{d8d8d8} 2.80 $\pm$ 1.60
        \\
        \cellcolor[HTML]{d8d8d8}$\text{FIZLE}_{\text{naive}}$
        & \cellcolor[HTML]{d8d8d8} 3.27 $\pm$ 2.00
        & \cellcolor[HTML]{d8d8d8} 2.93 $\pm$ 1.71
        & \cellcolor[HTML]{d8d8d8} 3.40 $\pm$ 1.80
        
        \\
        \midrule
        \cellcolor[HTML]{FFEACC}\textit{i}\texttt{Flip}-NL 
        & \textbf{5.10 $\pm$ 1.19}
        & \textbf{5.13 $\pm$ 1.12}
        & \textbf{4.83 $\pm$ 1.27}\\
        \midrule
        
        \textbf{Krippendorff's $\alpha$} 
        & 0.6467
        & 0.8690
        & 0.8694
        \\
        
        \bottomrule
    \end{tabular}
    }
    
    \caption{Human evaluation.}
    \label{tab:user_evaluation}
    \end{subtable}
    
    \vspace{0.5em}
    
    \begin{subtable}{\linewidth}
      \resizebox{\columnwidth}{!}{%
        \centering
    \begin{tabular}{l|ccc}
        \toprule
        \textbf{Method} 
        & \textbf{Complet.} 
        
        & \textbf{Overall Sat.}
        & \textbf{Feasib.} \\
        
        \midrule
        \cellcolor[HTML]{d8d8d8}CGG 
        & \cellcolor[HTML]{d8d8d8} 3.20 $\pm$ 1.94 
        
        & \cellcolor[HTML]{d8d8d8} 3.10 $\pm$ 2.06
        & \cellcolor[HTML]{d8d8d8} 3.93 $\pm$ 2.03
        \\
        \cellcolor[HTML]{d8d8d8}$\text{FIZLE}_{\text{naive}}$
        & \cellcolor[HTML]{d8d8d8} 3.10 $\pm$ 2.12
        
        & \cellcolor[HTML]{d8d8d8}3.07 $\pm$ 2.13
        & \cellcolor[HTML]{d8d8d8} 3.80 $\pm$ 2.17
        
        \\
        \midrule
        \cellcolor[HTML]{FFEACC}\textit{i}\texttt{Flip}-NL 
        & \textbf{4.53 $\pm$ 1.78}
        
        & \textbf{4.47 $\pm$ 1.86}
        & \textbf{4.40 $\pm$ 1.80}\\
        \midrule
        
        \textbf{Krippendorff's $\alpha$} 
        & 0.7615
        & 0.8215
        & 0.4005
        \\

        \bottomrule
    \end{tabular}
    }
    \caption{LLM-as-a-judge evaluation.}
    \label{tab:laaj_evaluation}
    \end{subtable}
    \caption{Evaluation results (\textbf{mean $\pm$ std}) on three subjective dimensions: \textit{Completeness}, \textit{Overall Satisfaction} and \textit{Feasibility}. Best results are \textbf{bolded}.}
    \label{tab:sub_evaluation}

\end{table}

\begin{figure}[t]
    \centering
    \includegraphics[width=\columnwidth]{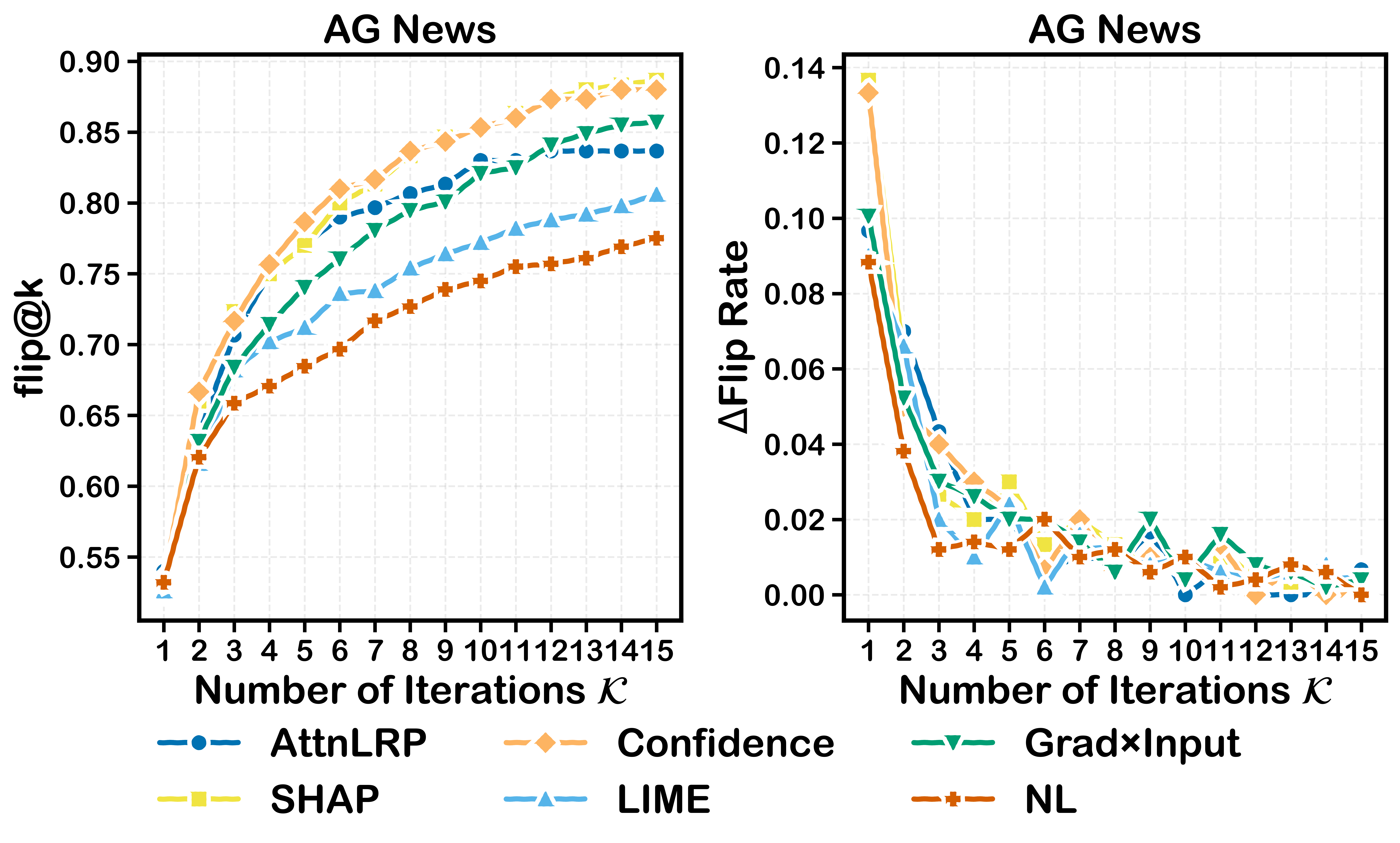}
    \caption{Iterative refinement effectiveness on \lm{OLMo2-7B} (\data{AG News}) across feedback signals. Left: flip@k curves up to $\mathcal{K}=15$. Right: per-iteration improvement in flip rate ($\Delta$Flip Rate). 
    }
    \label{fig:passk_delta}
    
\end{figure}
}%

\subsection{User Study \& LLM-as-a-Judge}
\label{subsec:user_study_and_laaj}

\paragraph{User Study.} 
As a complementary evidence, Table~\ref{tab:user_evaluation} \textbf{quantitatively} confirms that \textit{i}\texttt{Flip}-NL consistently outperforms CGG and \text{FIZLE} across all three subjective evaluation dimensions (\S\ref{subsubsec:subjective_rating}), achieving an average relative gain of 68.27\%. 
Notably, the most substantial improvement occurs in \textit{Overall Satisfaction}, with an 83.21\% relative gain.
To understand the source of these improvements, we \textbf{qualitatively} analyze representative failure cases in Appendix~\ref{sec:error_analysis}.
We observe two recurring issues in baseline methods: (i) they frequently edit only partial entity mentions, resulting in \textit{incomplete} changes that fail to flip the model prediction; and (ii) 
 they sometimes produce contextually inappropriate edits, yielding \textit{unnatural} counterfactuals.
In contrast, \textit{i}\texttt{Flip}-NL mitigates these issues through iterative refinement, improving both \textit{Completeness} and \textit{Feasibility} of counterfactuals.\footnote{See Appendix~\ref{subsec::example_user_study} and ~\ref{subsec:good_iter_refine} for per-annotator human/LLM ratings and examples showing how iterative refinement fixes \textit{incomplete} and \textit{unnatural} edits}

\paragraph{LLM-as-a-Judge.} 
Table~\ref{tab:laaj_evaluation} reveals that LaaJ results exhibit a trend consistent with human evaluation: \textit{i}\texttt{Flip}-NL invariably outperforms selected baselines across all three dimensions.\footnote{Evaluation across judge models are reported in Appendix~\ref{subsec::eval_llms}.}
Notably, the LLM judgments demonstrate larger standard deviations than human ratings, indicating that LLMs tend to assign more extreme scores.
Moreover, the human-LLM alignment, as measured by Krippendorff’s $\alpha$ 0.72, indicates relatively strong agreement between human and LaaJ evaluation. \looseness=-1

\subsection{Ablation Study}
\label{subsec:ablation}
\subsubsection{Iterative Refinement}

\begin{figure}[!t]
\centering

        \includegraphics[width=\columnwidth]{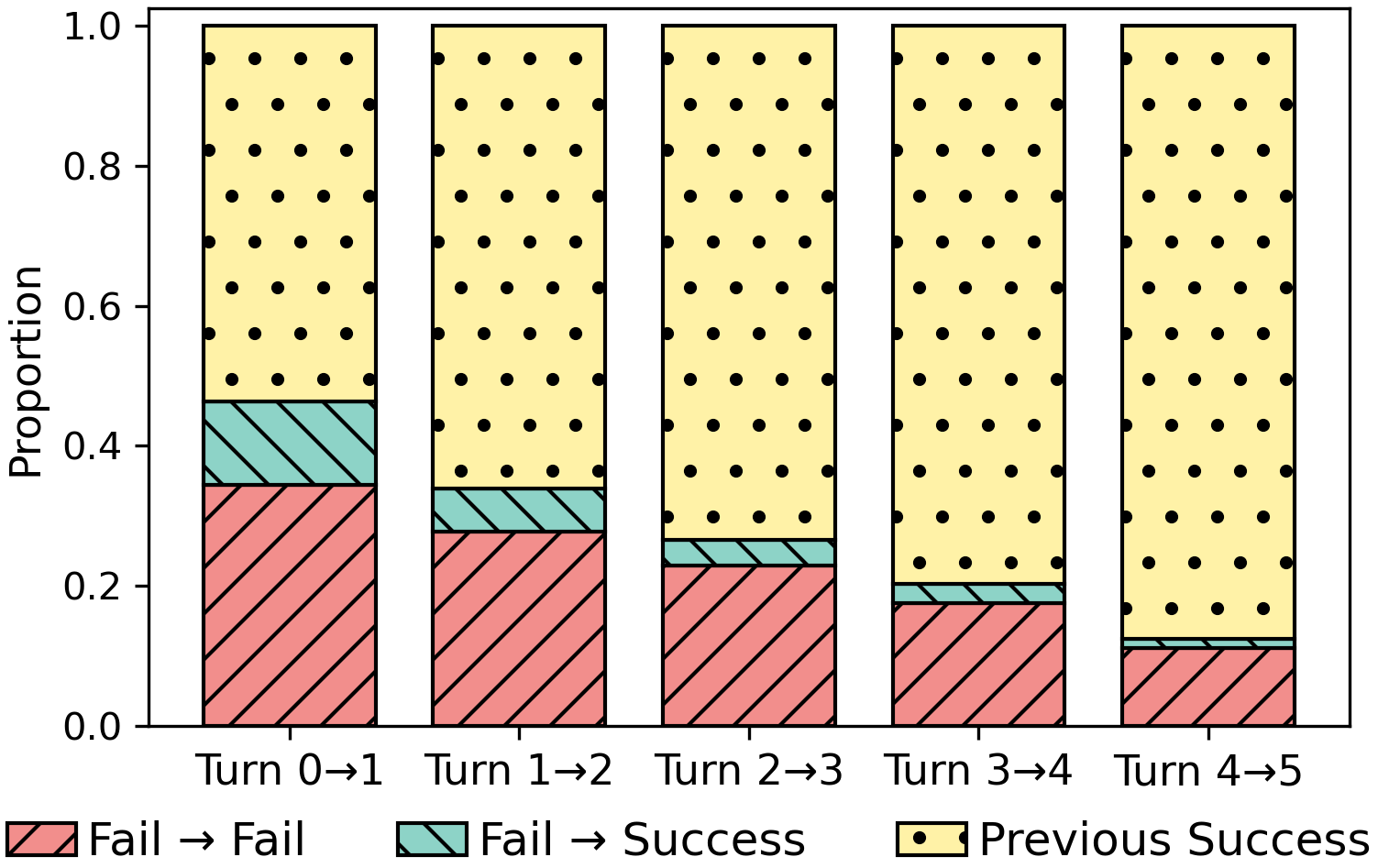}
    \caption{Iterative refinement with \lm{OLMo2-7B} under early stopping, averaged across all datasets and feedback signals. \textit{Fail $\rightarrow$ Fail} indicates no label change, \textit{Fail $\rightarrow$ Success} denotes a successful label flip in the current turn, and \textit{Previous Success} marks instances flipped in earlier turns.}
    \label{fig:overall_earlystop}
\end{figure}

Figure~\ref{fig:passk_delta} illustrates that \textbf{flip@k increases steadily with the number of refinement iterations}, though the improvement per round gradually decreases as refinement progresses.\footnote{Full flip@k curves and flip-rate improvements with \lm{OLMo2-7B} across all datasets are provided in Appendix~\ref{app:iter_effect_full}.} 
This trend demonstrates LLMs' feedback-driven self-correction capacity while revealing single-pass generation's substantial performance gap relative to iterative refinement, especially when initial counterfactuals exhibit low validity (e.g., \data{SNLI} premise editing achieves less than 30\% LFR in the first iteration).
Even with a small number of iterations ($\mathcal{K}=5$), iterative refinement yields markedly higher-validity counterfactuals than single-pass generation. Figure~\ref{fig:overall_earlystop} further provides these dynamics under early stopping, where \textit{Fail~$\rightarrow$~Fail} transitions diminish and \textit{Previous~Success} cases dominate over time.\footnote{\textit{Success~$\rightarrow$~Fail} are not observed due to early stopping.}

\begin{table}[t]
\centering
\footnotesize
\renewcommand{\arraystretch}{1}
\renewcommand\theadfont{\normalsize\bfseries}
\begin{adjustbox}{max width=\columnwidth}
\begin{tabular}{l|cccc}
\toprule
\textbf{\makecell{Feedback of\\ \textit{i}\texttt{Flip}}}
& \textbf{\data{IMDb}$\uparrow$}
& \textbf{\data{AG News}$\uparrow$}
& \makecell{\textbf{\data{SNLI}}\\\textbf{(Premise)}$\uparrow$}
& \makecell{\textbf{\data{SNLI}}\\\textbf{(Hypothesis)}$\uparrow$} \\
\midrule
\cellcolor[HTML]{C3F2F2}Without Feedback
 & \cellcolor[HTML]{C3F2F2}0.964 & \cellcolor[HTML]{C3F2F2}0.742 & \cellcolor[HTML]{C3F2F2}0.508 & \cellcolor[HTML]{C3F2F2}0.778 \\
\midrule
\cellcolor[HTML]{FFF8B8}Random Feedback
 & 0.970(\textcolor{morandiRed}{$\uparrow$0.006})
 & 0.794(\textcolor{morandiRed}{$\uparrow$0.052})
 & 0.672(\textcolor{morandiRed}{$\uparrow$0.164})
 & 0.764(\textcolor{morandiGreen}{$\downarrow$0.014}) \\
\cellcolor[HTML]{E8D8FF}Least SHAP
 & 0.970(\textcolor{morandiRed}{$\uparrow$0.006})
 & 0.770(\textcolor{morandiRed}{$\uparrow$0.028})
 & 0.612(\textcolor{morandiRed}{$\uparrow$0.104})
 & 0.780(\textcolor{morandiRed}{$\uparrow$0.002}) \\
\cellcolor[HTML]{E8D8FF}Least AttnLRP
 & 0.980(\textcolor{morandiRed}{$\uparrow$0.016})
 & 0.776(\textcolor{morandiRed}{$\uparrow$0.034})
 & 0.562(\textcolor{morandiRed}{$\uparrow$0.054})
 & 0.764(\textcolor{morandiGreen}{$\downarrow$0.014}) \\
\cellcolor[HTML]{E8D8FF}Least Grad$\times$Input
 & 0.968(\textcolor{morandiRed}{$\uparrow$0.004})
 & 0.758(\textcolor{morandiRed}{$\uparrow$0.016})
 & 0.593(\textcolor{morandiRed}{$\uparrow$0.085})
 & 0.760(\textcolor{morandiGreen}{$\downarrow$0.018}) \\
\cellcolor[HTML]{E8D8FF}Least LIME
 & 0.968(\textcolor{morandiRed}{$\uparrow$0.004})
 & 0.725(\textcolor{morandiGreen}{$\downarrow$0.017})
 & 0.556(\textcolor{morandiRed}{$\uparrow$0.048})
 & 0.754(\textcolor{morandiGreen}{$\downarrow$0.024}) \\
\cmidrule(lr){1-5}
\cellcolor[HTML]{D9EAFD}Conf
 & 0.964(=)
 & 0.784(\textcolor{morandiRed}{$\uparrow$0.042})
 & 0.660(\textcolor{morandiRed}{$\uparrow$0.152})
 & 0.826(\textcolor{morandiRed}{$\uparrow$0.048}) \\
 \cellcolor[HTML]{E1F5E1}SHAP
 & 0.968(\textcolor{morandiRed}{$\uparrow$0.004})
 & 0.802(\textcolor{morandiRed}{$\uparrow$0.060})
 & 0.587(\textcolor{morandiRed}{$\uparrow$0.079})
 & 0.760(\textcolor{morandiGreen}{$\downarrow$0.018}) \\
 \cellcolor[HTML]{E1F5E1}AttnLRP
 & 0.978(\textcolor{morandiRed}{$\uparrow$0.014})
 & 0.768(\textcolor{morandiRed}{$\uparrow$0.026})
 & 0.596(\textcolor{morandiRed}{$\uparrow$0.088})
 & 0.810(\textcolor{morandiRed}{$\uparrow$0.032}) \\
 \cellcolor[HTML]{E1F5E1}Grad$\times$Input
 & 0.976(\textcolor{morandiRed}{$\uparrow$0.012})
 & 0.773(\textcolor{morandiRed}{$\uparrow$0.031})
 & 0.624(\textcolor{morandiRed}{$\uparrow$0.116})
 & 0.782(\textcolor{morandiRed}{$\uparrow$0.004}) \\
 \cellcolor[HTML]{E1F5E1}LIME
 & 0.968(\textcolor{morandiRed}{$\uparrow$0.004})
 & 0.730(\textcolor{morandiGreen}{$\downarrow$0.012})
 & 0.558(\textcolor{morandiRed}{$\uparrow$0.050})
 & 0.736(\textcolor{morandiGreen}{$\downarrow$0.042}) \\
\cellcolor[HTML]{FFEACC}NL
 & 0.984(\textcolor{morandiRed}{$\uparrow$0.020})
 & 0.735(\textcolor{morandiGreen}{$\downarrow$0.007})
 & 0.638(\textcolor{morandiRed}{$\uparrow$0.130})
 & 0.770(\textcolor{morandiGreen}{$\downarrow$0.008}) \\
\bottomrule
\end{tabular}
\end{adjustbox}

\caption{
Automatic evaluation results in terms of LFR of \lm{OLMo2-7B}. 
The tables present an ablation study of different feedback settings, including 
\mbox{{\setlength{\fboxsep}{1pt}\colorbox[HTML]{C3F2F2}{\textit{Without Feedback}}}}, 
\mbox{{\setlength{\fboxsep}{1pt}\colorbox[HTML]{FFF8B8}{\textit{Random Feedback}}}}, 
\mbox{{\setlength{\fboxsep}{1pt}\colorbox[HTML]{E8D8FF}{\textit{Least-Attributed Feedback}}}}, and the full variants of \textit{i}\texttt{Flip}.
Colored values indicate relative changes compared to the \mbox{{\setlength{\fboxsep}{1pt}\colorbox[HTML]{C3F2F2}{\textit{Without Feedback} }}} baseline 
(\textcolor{morandiRed}{$\uparrow$} = increase, \textcolor{morandiGreen}{$\downarrow$} = decrease).
}
\label{tab:ablation_feedback_all_lfr}

\end{table}

\subsubsection{Feedback Signal}
\label{subsubsec:feedback}


We examine whether incorporating feedback is intrinsically beneficial for iterative counterfactual refinement. Table~\ref{tab:ablation_feedback_all_lfr} illustrates that \textbf{incorporating feedback generally improves LFR} compared to the \emph{without feedback} baseline (Appendix~\ref{app:extra_ablation_feedback_signal}).\footnote{We want to highlight that \emph{without feedback} baseline is not equivalent to Causal What-Ifs: Causal What-Ifs does not leverage early stopping; instead, it continues refining counterfactuals after identifying a valid one to obtain a more minimal edit. However, our results in \S\ref{subsubsec::early_stopping} show that this additional refinement is not beneficial in practice.} However, the optimal choice of feedback is task- and signal-dependent, and certain signals may exhibit variable effectiveness, potentially introducing misguidance into the refinement process (\S\ref{par:feedback_signal_comparison_6.1}). 
We further assess the effectiveness of \textit{i}\texttt{Flip} with \emph{random feedback}, which provides only modest LFR gains over the \emph{without feedback} setting. Among attribution-based signals, directing edits towards the most important words consistently outperforms targeting the least important words, yielding a mean LFR gain of 1.09\% alongside an 11.02\% reduction in edits, indicating more effective yet less disruptive edits. 



\subsubsection{Early Stopping}
\label{subsubsec::early_stopping}

Tables~\ref{tab:noearlystop1} and~\ref{tab:noearlystop2} show that \textbf{the removal of early stopping induces a decline in LFR}, which is attributed to valid counterfactuals being overturned in successive iterations, as illustrated by the proportion of \textit{(Success $\rightarrow$ Fail)} cases (Figure~\ref{fig:without_early_stop}).\footnote{Detailed examples appear in Appendix \ref{app:no_early_stop_example1} and \ref{app:no_early_stop_example2}.} 
This demonstrates the crucial role of early stopping in maintaining reliable counterfactual generation. Conversely, although additional iterations enable LLMs to enhance text fluency, they typically involve more extensive edits.
Our extended analysis in Appendix~\ref{subsec::transferability} further reveals that early stopping is the primary driver of cross-architecture transferability, as it prevents the generator from overfitting to the specific decision boundary of the explained model.



\begin{table}[!t]
    \centering
    \footnotesize
    \renewcommand{\arraystretch}{1}
    \resizebox{\columnwidth}{!}{
    \begin{tabular}{c|cc|c|cc|cc}
        \toprule
        \textbf{Method}
        & \textbf{\data{IMDb}} & \textbf{CFs}
        & \textbf{\data{AG News}}
        & \textbf{\data{SNLI} (P)} & \textbf{CFs}
        & \textbf{\data{SNLI} (H)} & \textbf{CFs} \\
        \midrule
        \cellcolor[HTML]{d8d8d8}Baseline
        & \cellcolor[HTML]{d8d8d8}77.93 & \cellcolor[HTML]{d8d8d8}50.09
        & \cellcolor[HTML]{d8d8d8}66.87
        & \cellcolor[HTML]{d8d8d8}40.87 & \cellcolor[HTML]{d8d8d8}33.43
        & \cellcolor[HTML]{d8d8d8}40.87 & \cellcolor[HTML]{d8d8d8}33.85 \\
        \cellcolor[HTML]{d8d8d8}Human
        & \cellcolor[HTML]{d8d8d8}\textbf{92.73} & \cellcolor[HTML]{d8d8d8}\textbf{91.26}
        & \cellcolor[HTML]{d8d8d8}--
        & \cellcolor[HTML]{d8d8d8}\textbf{42.27} & \cellcolor[HTML]{d8d8d8}\textbf{50.90}
        & \cellcolor[HTML]{d8d8d8}\textbf{58.40} & \cellcolor[HTML]{d8d8d8}\textbf{56.95} \\

        \cellcolor[HTML]{d8d8d8}CGG
        & \cellcolor[HTML]{d8d8d8}61.40 & \cellcolor[HTML]{d8d8d8}70.83
        & \cellcolor[HTML]{d8d8d8}46.27
        & \cellcolor[HTML]{d8d8d8}32.80 & \cellcolor[HTML]{d8d8d8}33.37
        & \cellcolor[HTML]{d8d8d8}32.80 & \cellcolor[HTML]{d8d8d8}33.37 \\
        \cellcolor[HTML]{d8d8d8}FIZLE
        & \cellcolor[HTML]{d8d8d8}79.93 & \cellcolor[HTML]{d8d8d8}71.73
        & \cellcolor[HTML]{d8d8d8}\textbf{72.47}
        & \cellcolor[HTML]{d8d8d8}36.93 & \cellcolor[HTML]{d8d8d8}35.35
        & \cellcolor[HTML]{d8d8d8}40.27 & \cellcolor[HTML]{d8d8d8}37.64 \\
        \cellcolor[HTML]{d8d8d8}Causal What-Ifs
        & \cellcolor[HTML]{d8d8d8}91.87 & \cellcolor[HTML]{d8d8d8}84.45
        & \cellcolor[HTML]{d8d8d8}73.53
        & \cellcolor[HTML]{d8d8d8}35.20 & \cellcolor[HTML]{d8d8d8}25.23
        & \cellcolor[HTML]{d8d8d8}36.40 & \cellcolor[HTML]{d8d8d8}33.22 \\
        \midrule
        
        \cellcolor[HTML]{D9EAFD}Conf
        & 87.93 & 71.09
        & 78.60
        & 38.47 & \textbf{32.98}
        & \textbf{44.13} & \textbf{36.88} \\
        \cellcolor[HTML]{E1F5E1}SHAP
        & 91.33 & 90.30
        & 80.27
        & 37.40 & 27.82
        & 40.33 & 35.83 \\
        \cellcolor[HTML]{E1F5E1}AttenLRP
        & 89.27 & 90.28
        & 77.87
        & 38.13 & 27.97
        & 40.13 & 35.53 \\
        \cellcolor[HTML]{E1F5E1}Grad$\times$Input
        & \textbf{93.07} & 90.07
        & 79.07
        & 39.00 & 26.98
        & 39.40 & 35.98 \\
        \cellcolor[HTML]{E1F5E1}LIME
        & 91.40 & \textbf{90.91}
        & 78.80
        & 38.33 & 28.93
        & 38.93 & 35.74 \\
        \cellcolor[HTML]{FFEACC}NL
        & 90.73 & 71.97
        & \textbf{81.80}
        & \textbf{40.40} & 30.91
        & 37.47 & 33.79 \\
        \bottomrule
    \end{tabular}
    }
    \caption{Accuracy (\%) on test sets and human-annotated counterfactuals (CFs). Counterfactuals used for CDA are generated by \lm{OLMo2-7B}. Best results are \textbf{bolded}.}
    \label{tab:augmentation_all}
    
\end{table}

\subsection{Counterfactual Data Augmentation}



Although human-annotated counterfactuals provide marginally stronger performance gains than \textit{i}\texttt{Flip}-generated counterfactuals (Table~\ref{tab:augmentation_all}), their prohibitive cost and time requirements motivate automated approaches. 
Notably, CDA with \textit{i}\texttt{Flip}-generated counterfactuals consistently outperforms CDA using counterfactuals from other baselines, while also improving performance and robustness over the baseline on most datasets.\footnote{Additional OOD augmentation results are provided in Appendix~\ref{app:augmentation_detail}. For the News Classification task, human-annotated counterfactuals are not available.
} We further observe a moderate correlation between CDA performance and counterfactual quality (Appendix~\ref{subsec:corr_aug_subsec}).



\section{Conclusion}
In this work, we propose \textit{i}\texttt{Flip}, a feedback-driven framework that iteratively refines counterfactuals by incorporating multiple feedback signals to enhance their validity. 
Empirical results demonstrate that \textit{i}\texttt{Flip} consistently surpasses selected baselines and achieves a more favorable validity-minimality balance.
In addition, a complementary user study shows that \textit{i}\texttt{Flip}-NL receives higher ratings than the baselines across all subjective dimensions and that its iterative refinement mechanism substantially improves the completeness and realism of generated counterfactuals.
Moreover, our ablation studies reveal three key components of \textit{i}\texttt{Flip}: iterative refinement, feedback signals, and early stopping, with early stopping  emerging as the simplest and most cost-effective.
Finally, we extend the study to counterfactual data augmentation and show that counterfactuals generated by \textit{i}\texttt{Flip} effectively enhance model performance and robustness.




\section*{Limitations}
In this work, we propose a feedback-driven framework \textit{i}\texttt{Flip} for iterative counterfactual refinement. While \textit{i}\texttt{Flip} noticeably improves the validity of generated counterfactuals compared to five selected baselines, we acknowledge several limitations.

\paragraph{Usage of limited Feature Attribution Methods.} We do not exhaustively explore all feature attribution-based feedback signals; investigating additional feature attribution methods remains future work.

\paragraph{Computational Costs.} Despite the effectiveness of \textit{i}\texttt{Flip} in producing high-quality counterfactuals, it incurs relatively high computational costs, particularly when using natural language feedback, as the iterative refinement process requires multiple rounds of generation and validation compared to single-pass methods.

\paragraph{English-Centric Evaluation.} In addition, all experiments in this work were carried out exclusively on English datasets, and it remains unclear how well the approach would generalize to other languages. We plan to extend our evaluation to a multilingual setting in future work to assess the generalizability of our findings across languages.

\paragraph{Limited Scale of the User Study.} Following prior work \cite{nguyen-etal-2024-ceval-benchmark, mcaleese2024comparativeanalysiscounterfactualexplanation}, we predominantly assess the quality of generated counterfactuals by three widely used automatic metrics regarding \textit{validity}, \textit{fluency}, and \textit{minimality}, given that user studies for counterfactuals are uncommon in the literature. Nevertheless, to further assess textual quality beyond automatic metrics, we conduct a small-scale user study with $(N=3)$ participants evaluating \textit{completeness}, \textit{overall satisfaction}, and \textit{feasibility}, following \citet{domnich2024unifyingevaluationcounterfactualexplanations}. As future work, we consider to include a comprehensive and large-scale user study to assess textual quality, e.g., coherence, understandability, or complexity \cite{domnich2024unifyingevaluationcounterfactualexplanations, wang2026largelanguagemodelsexplain}, and the effectiveness of counterfactuals for model predictions.

\section*{Ethics Statement}
The participants in our user studies were compensated at or above the minimum wage
in accordance with the standards of our host institutions’
regions. The annotation took each annotator 30
minutes on average.

\section*{Acknowledgments}
We are indebted to the anonymous reviewers of ACL 2026 and ARR March for their helpful and rigorous feedback.
We sincerely thank \textbf{Fedor Splitt}, \textbf{Jiaao Li}, and \textbf{Alexandra Tsiakalou} for their valuable participation in the user study.

\bibliography{custom}

\appendix

\section{Dataset}
\label{app:dataset}

\subsection{Dataset Example}

\begin{figure}[htbp]
    \centering
    \begin{adjustbox}{max width=\columnwidth}
        \includegraphics{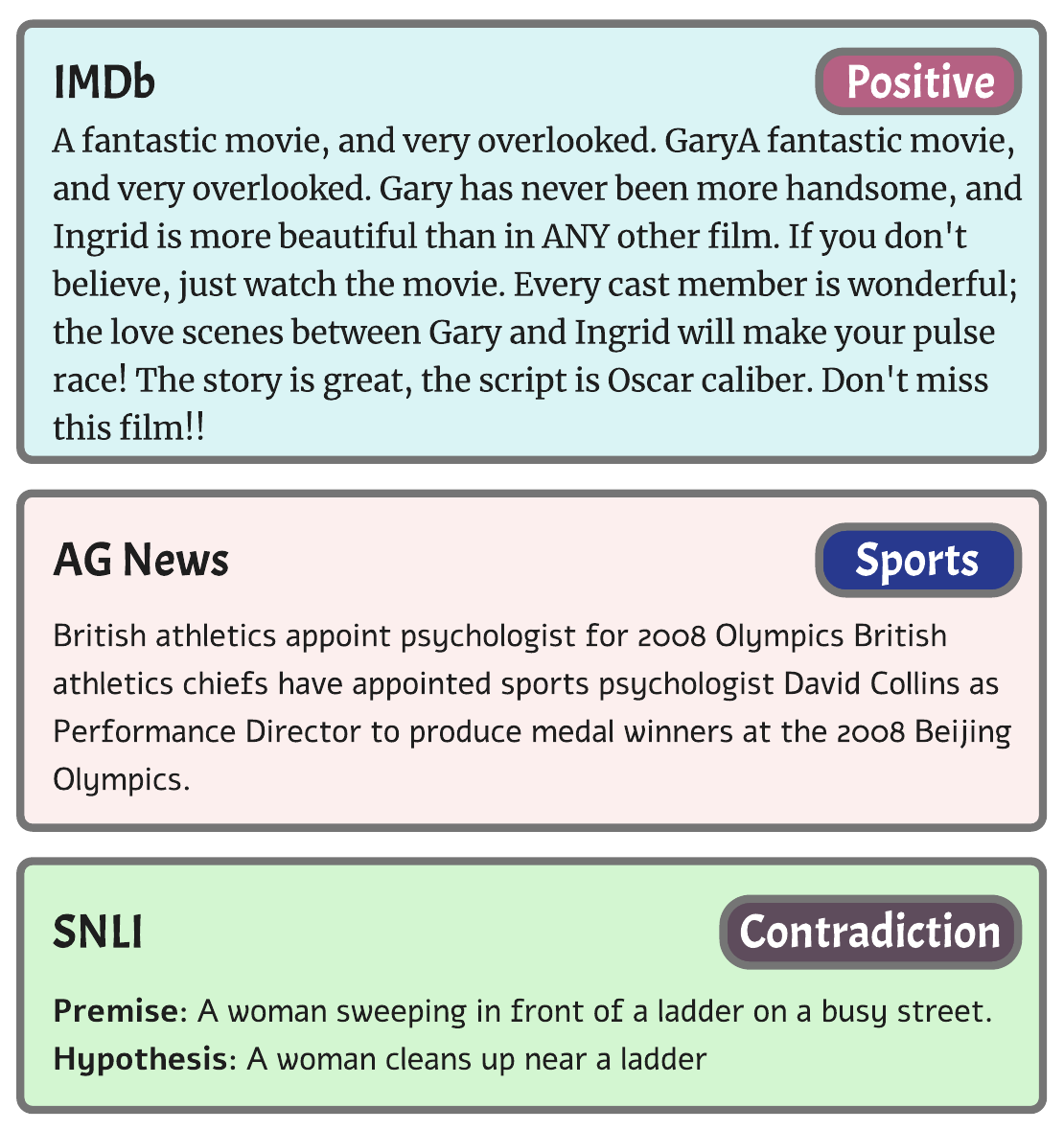}

    \end{adjustbox}
    \caption{Examples from \data{IMDb}, \data{AG News}, and \data{SNLI} datasets.}
    \label{fig:dataset_examples}
\end{figure}

Figure~\ref{fig:dataset_examples} presents a representative example and its corresponding gold label for each dataset.

\subsection{Label Distributions}
\begin{figure}[h]
    \centering
    \includegraphics[width=\columnwidth]{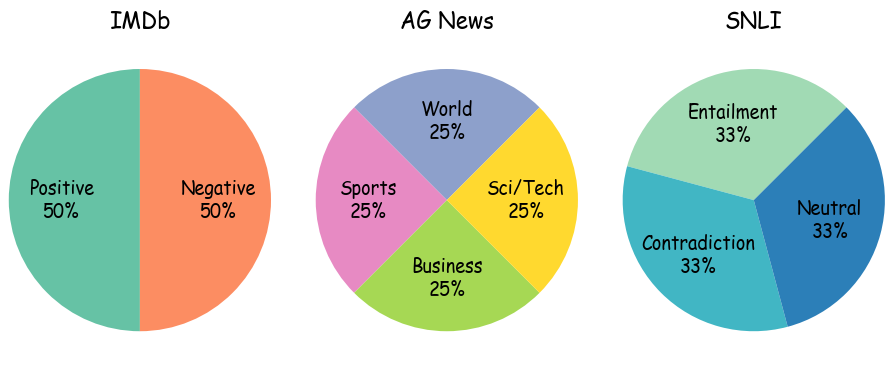}
    \caption{Label distributions across datasets.}
    \label{fig:label_distributions}
\end{figure}

Figure~\ref{fig:label_distributions} summarizes the label distributions across three datasets.

\subsection{Dataset Sources}
Table~\ref{tab:dataset_sources} illustrates the sources of the datasets used in our experiments.

\section{Experimental Settings}
\label{app:setting}

\subsection{Experimental Parameters}
\label{app:exp_settings}
Table~\ref{tab:exp_settings} summarizes the key parameters used in our experiments. These values were kept consistent across datasets unless stated otherwise.

\begin{table*}[htbp]
\centering
\renewcommand\arraystretch{1.2}
\resizebox{\textwidth}{!}{  
\begin{tabular}{lll}
\toprule
\textbf{Dataset} & \textbf{Citation} & \textbf{Source} \\
\midrule
\data{IMDb} & \citet{maas-etal-2011-imdb} & \url{https://huggingface.co/datasets/stanfordnlp/imdb} \\
\data{AG News} & \citet{zhang-2015-agnews} & \url{https://huggingface.co/datasets/sentence-transformers/agnews} \\
\data{SNLI} & \citet{bowman-etal-2015-large} & \url{https://huggingface.co/datasets/stanfordnlp/snli} \\
\bottomrule
\end{tabular}
}
\caption{Datasets employed in our experiments with their citations and sources.}
\label{tab:dataset_sources}
\end{table*}

\begin{table}[htbp]
\centering
\small
\renewcommand\arraystretch{1.15}
\setlength{\tabcolsep}{6pt}
\begin{tabularx}{\linewidth}{@{}lY@{}}
\toprule
\textbf{Parameter} & \textbf{Value} \\
\midrule
\multicolumn{2}{@{}l}{\textit{Generator Parameters}} \\
Temperature & 0.9 \\
Top-p & 0.95 \\
Top-k & 50 \\
Max new tokens & 4096 \\
Number of iterations & 5 \\
Top-$k$ important words & \(\max(10, \lfloor 0.10 \times |\text{orig\_words}| \rfloor)\) \\
\midrule
\multicolumn{2}{@{}l}{\textit{Explained Model Settings}} \\
Voting Strategy & Majority voting \\
Window Size & 512 \\
Stride & 256 \\
\midrule
\multicolumn{2}{@{}l}{\textit{Hardware}} \\
\lm{OLMo2-7B} & 1× V100 (32\,GB) \\
\lm{Qwen3-32B} & 1× H100 (80\,GB) \\
\lm{LLaMA3.3-70B} & 2× H100 (80\,GB) \\
\bottomrule
\end{tabularx}
\caption{Experimental settings.}
\label{tab:exp_settings}
\end{table}

\subsection{Inference Time}
\label{app:inference_time}
Table~\ref{tab:inference_time} reports the inference time  required for the \lm{OLMo2-7B} generator model across all tasks.

\begin{table}[htbp]
\centering
\small
\renewcommand\arraystretch{1.15}
\setlength{\tabcolsep}{6pt}
\begin{tabularx}{\linewidth}{@{}p{.33\linewidth}Y r@{}}
\toprule
\textbf{Task} & \textbf{Method} & \textbf{Duration (hours)} \\
\midrule
\data{IMDb}             & Conf            & 3.40 \\
                 & AttnLRP         & 3.22 \\
                 & SHAP            & 3.75 \\
                 & Grad$\times$Input & 3.45 \\
                 & LIME            & 3.72 \\
                 & NL              & 4.55 \\
\midrule
\data{AG News}           & Conf            & 7.30 \\
                 & AttnLRP         & 7.22 \\
                 & SHAP            & 9.73 \\
                 & Grad$\times$Input & 7.37 \\
                 & LIME            & 9.86 \\
                 & NL              & 17.90 \\
\midrule
\data{SNLI}-Premise     & Conf            & 1.39 \\
                 & AttnLRP         & 1.64 \\
                 & SHAP            & 2.10 \\
                 & Grad$\times$Input & 1.55 \\
                 & LIME            & 4.80 \\
                 & NL              & 7.57 \\
\midrule
\data{SNLI}-Hypothesis  & Conf            & 0.93 \\
                 & AttnLRP         & 1.13 \\
                 & SHAP            & 1.37 \\
                 & Grad$\times$Input & 1.14 \\
                 & LIME            & 3.45 \\
                 & NL              & 5.26 \\
\bottomrule
\end{tabularx}
\caption{Inference time (in hours) for \lm{OLMo2-7B} across \data{SNLI}, \data{IMDb}, and \data{AG News}.}
\label{tab:inference_time}
\end{table}

\subsection{Generator Models}
\label{app:generators}
We employ three LLMs as generator models to produce counterfactuals, representing diverse scales, parameter sizes, and model families. Table~\ref{tab:generator_models} summarizes their details.

\begin{table*}[htbp]
\centering
\renewcommand\arraystretch{1.2}
\resizebox{\textwidth}{!}{  
\begin{tabular}{llll}
\toprule
\textbf{Name} & \textbf{Citation} & \textbf{Size} & \textbf{Source} \\
\midrule
\texttt{OLMo2-7B} & \citet{olmo20252olmo2furious} & 7B & \url{https://huggingface.co/allenai/OLMo-2-1124-7B-Instruct} \\
\texttt{Qwen3-32B} & \citet{yang2025qwen3technicalreport} & 32B & \url{https://huggingface.co/Qwen/Qwen3-32B} \\
\texttt{LlaMa3.3-70B} & \citet{grattafiori2024llama3herdmodels} & 70B & \url{https://huggingface.co/meta-llama/Llama-3.3-70B-Instruct} \\
\bottomrule
\end{tabular}
}
\caption{Generator models employed in our experiments.}
\label{tab:generator_models}
\end{table*}

\subsection{Explained Models}
\label{app:verifier}

Table~\ref{tab:explained_models} summarizes the \lm{BERT} and \lm{RoBERTa} models fine-tuned on the selected datasets. These models serve as the classifier that determines whether the counterfactuals truly change the previous model prediction.

\begin{table*}[htbp]
\centering
\renewcommand\arraystretch{1.2}
\resizebox{\textwidth}{!}{  
\begin{tabular}{lll}
\toprule
\textbf{Task} & \textbf{Model} & \textbf{Source} \\
\midrule
\multirow{2}{*}{\data{IMDb}}   & \texttt{textattack/bert-base-uncased-imdb} & \url{https://huggingface.co/textattack/bert-base-uncased-imdb} \\
& \texttt{textattack/roberta-base-imdb} & \url{https://huggingface.co/textattack/roberta-base-imdb} \\
\midrule
\multirow{2}{*}{\data{AG News}} & \texttt{textattack/bert-base-uncased-ag-news} & \url{https://huggingface.co/textattack/bert-base-uncased-ag-news} \\
& \texttt{textattack/roberta-base-ag-news} & \url{https://huggingface.co/textattack/roberta-base-ag-news} \\
\midrule
\multirow{2}{*}{\data{SNLI}}   & \texttt{textattack/bert-base-uncased-snli} & \url{https://huggingface.co/textattack/bert-base-uncased-snli} \\
& \texttt{utahnlp/snli\_roberta-base\_seed-2} & \url{https://huggingface.co/utahnlp/snli_roberta-base_seed-2} \\
\bottomrule
\end{tabular}
}
\caption{Explained models (BERT and RoBERTa) used for different datasets.}
\label{tab:explained_models}
\end{table*}

\section{Extended Evaluation Results}

\subsection{Evaluation Results on \lm{RoBERTa-base}}
\label{subsec::roberta_results}

Table \ref{tab:roberta_results} presents the automatic evaluation results for counterfactuals generated by \textit{i}\texttt{Flip} on the \texttt{RoBERTa-base} model, using \lm{LLaMA3.3-70B} as the generator. 
While agentic baselines like Causal What-Ifs remain competitive on NLI tasks, \textit{i}\texttt{Flip} demonstrates superior consistency, particularly on sentiment analysis and news topic classification where it frequently outperforms baselines.

\begin{table*}[t!]
    \centering
    \renewcommand*{\arraystretch}{0.5}
    
    \footnotesize
    \resizebox{\textwidth}{!}{%
        \begin{tabular}{ll|ccc|ccc|ccc|ccc}
            \toprule
            \textbf{} & \textbf{Method} 
            & \multicolumn{3}{c|}{\textbf{\data{IMDb}}}
            & \multicolumn{3}{c|}{\textbf{\data{AG News}}}
            & \multicolumn{3}{c|}{\makecell{\textbf{\data{SNLI}} \textbf{(Premise)}}}
            & \multicolumn{3}{c}{\makecell{\textbf{\data{SNLI}} \textbf{(Hypothesis)}}}\\
            \cmidrule(lr){3-5}\cmidrule(lr){6-8}\cmidrule(lr){9-11}\cmidrule(lr){12-14}
             & & LFR$\uparrow$ & SS$\uparrow$ & PPL$\downarrow$
               & LFR$\uparrow$ & SS$\uparrow$ & PPL$\downarrow$
               & LFR$\uparrow$ & SS$\uparrow$ & PPL$\downarrow$
               & LFR$\uparrow$ & SS$\uparrow$ & PPL$\downarrow$
            \\
            \midrule
            \midrule
            
            \multicolumn{14}{c}{\texttt{RoBERTa-base}}\\
            \midrule

            
            
            \multirow{6}{*}[-0.7cm]{%
              \rotatebox[origin=c]{90}{\makecell{\lm{LLaMA}\\\lm{3.3-70B}}}%
            }
            
            & \cellcolor[HTML]{d8d8d8}{CGG} 
            & \cellcolor[HTML]{d8d8d8}0.930 & \cellcolor[HTML]{d8d8d8}0.892 & \cellcolor[HTML]{d8d8d8}54.57
            & \cellcolor[HTML]{d8d8d8}0.748 & \cellcolor[HTML]{d8d8d8}0.542  & \cellcolor[HTML]{d8d8d8}130.31
            & \cellcolor[HTML]{d8d8d8}0.635 & \cellcolor[HTML]{d8d8d8}0.893 & \cellcolor[HTML]{d8d8d8}59.23
            & \cellcolor[HTML]{d8d8d8}0.680 & \cellcolor[HTML]{d8d8d8}\textbf{0.944} & \cellcolor[HTML]{d8d8d8}54.23  \rule{0pt}{10pt}\\
            
            & \cellcolor[HTML]{d8d8d8}{$\text{FIZLE}$} 
            & \cellcolor[HTML]{d8d8d8}0.956 & \cellcolor[HTML]{d8d8d8}0.868 & \cellcolor[HTML]{d8d8d8}34.88
            & \cellcolor[HTML]{d8d8d8}0.345 & \cellcolor[HTML]{d8d8d8}\textbf{0.707} & \cellcolor[HTML]{d8d8d8}56.59
            & \cellcolor[HTML]{d8d8d8}\textbf{0.759} & \cellcolor[HTML]{d8d8d8}\textbf{0.908} & \cellcolor[HTML]{d8d8d8}41.54
            & \cellcolor[HTML]{d8d8d8}0.821 & \cellcolor[HTML]{d8d8d8}0.929 & \cellcolor[HTML]{d8d8d8}\textbf{42.39}  \rule{0pt}{10pt}\\
            
            & \cellcolor[HTML]{d8d8d8}{Causal What-Ifs} 
            & \cellcolor[HTML]{d8d8d8}0.908 & \cellcolor[HTML]{d8d8d8}0.788 & \cellcolor[HTML]{d8d8d8}\textbf{32.78}
            & \cellcolor[HTML]{d8d8d8}0.656 & \cellcolor[HTML]{d8d8d8}0.442 & \cellcolor[HTML]{d8d8d8}\textbf{46.18}
            & \cellcolor[HTML]{d8d8d8}0.750 & \cellcolor[HTML]{d8d8d8}0.864 & \cellcolor[HTML]{d8d8d8}\textbf{41.07}
            & \cellcolor[HTML]{d8d8d8}0.740 & \cellcolor[HTML]{d8d8d8}0.892 & \cellcolor[HTML]{d8d8d8}43.63  \rule{0pt}{10pt}\\
            
            &\cellcolor[HTML]{D9EAFD}\textit{i}\texttt{Flip}-Conf              
            & \uwave{\textbf{1.000}} & \uwave{\textbf{0.895}} & 37.48
            & 0.880 & 0.575 & \uwave{55.55}
            & 0.665 & 0.688 & 62.94
            & \uwave{\textbf{0.830}} & 0.799 & 47.43 \rule{0pt}{10pt}\\
            
            & \cellcolor[HTML]{E1F5E1}\textit{i}\texttt{Flip}-SHAP              
            & \uwave{\textbf{1.000}} & \uwave{\textbf{0.895}} & \uwave{36.19}
            & 0.885 & 0.542 & 57.69
            & 0.680 & 0.700 & \uwave{57.99}
            & 0.800 & 0.795 & \uwave{44.69}  \rule{0pt}{10pt}\\
            
            & \cellcolor[HTML]{FFEACC}\textit{i}\texttt{Flip}-NL 
            & \uwave{\textbf{1.000}} & 0.893 & 36.48
            & \uwave{\textbf{0.915}} & \uwave{0.583} & 57.66
            & \uwave{0.730} & \uwave{0.789} & 77.00
            & 0.815 & \uwave{0.815} & 52.91 \rule{0pt}{10pt}\\

            \bottomrule
        \end{tabular}
    }
    \caption{Automatic evaluation results of counterfactuals on the \texttt{RoBERTa-base} model using \textit{i}\texttt{Flip} methods with different feedback types: \ding{182} confidence (\mbox{{\setlength{\fboxsep}{1pt}\colorbox[HTML]{D9EAFD}{\strut\textsf{Conf}}}}), \ding{183} feature attribution (\mbox{{\setlength{\fboxsep}{1pt}\colorbox[HTML]{E1F5E1}{\strut\textsf{SHAP}}}}), and \ding{184} natural language (\mbox{{\setlength{\fboxsep}{1pt}\colorbox[HTML]{FFEACC}{\strut\textsf{NL}}}}). \textbf{Boldface} indicates the best result across methods. 
    \uwave{Wavy underline} indicates the best feedback type within \textit{i}\texttt{Flip}.
    }
    \label{tab:roberta_results}
\end{table*}

\subsection{Transferability}
\label{subsec::transferability}

\subsubsection{Transferability of the counterfactuals across architectures}
Table~\ref{tab:transfer_results1} reports the \emph{transferability} of the generated counterfactuals across classifier architectures. 
Specifically, all counterfactuals are generated for \lm{BERT} classifiers, while the automatic evaluation is conducted by replacing the original \lm{BERT} classifier with a \lm{RoBERTa} classifier.
\begin{table*}[t!]
    \centering
    \renewcommand*{\arraystretch}{0.5}
    
    \footnotesize
    \resizebox{\textwidth}{!}{%
\begin{tabular}{ll|ccc|ccc|ccc|ccc}
\toprule
\textbf{} & \textbf{Method} 
& \multicolumn{3}{c|}{\textbf{\data{IMDb}}}
& \multicolumn{3}{c|}{\textbf{\data{AG News}}}
& \multicolumn{3}{c|}{\makecell{\textbf{\data{SNLI}} \textbf{(Premise)}}}
& \multicolumn{3}{c}{\makecell{\textbf{\data{SNLI}} \textbf{(Hypothesis)}}}\\
\cmidrule(lr){3-5}\cmidrule(lr){6-8}\cmidrule(lr){9-11}\cmidrule(lr){12-14}
 & & LFR$\uparrow$ & SS$\uparrow$ & PPL$\downarrow$
   & LFR$\uparrow$ & SS$\uparrow$ & PPL$\downarrow$
   & LFR$\uparrow$ & SS$\uparrow$ & PPL$\downarrow$
   & LFR$\uparrow$ & SS$\uparrow$ & PPL$\downarrow$
\\


\midrule

\multirow{9}{*}[-0.5cm]{\rotatebox[origin=c]{90}{\lm{OLMo2-7B}}}
& \cellcolor[HTML]{d8d8d8}{CGG}               
& \cellcolor[HTML]{d8d8d8}0.891 & \cellcolor[HTML]{d8d8d8}\textbf{0.837} & \cellcolor[HTML]{d8d8d8}44.67
& \cellcolor[HTML]{d8d8d8}\textbf{0.560} & \cellcolor[HTML]{d8d8d8}0.380 & \cellcolor[HTML]{d8d8d8}208.56
& \cellcolor[HTML]{d8d8d8}0.257 & \cellcolor[HTML]{d8d8d8}\textbf{0.882} & \cellcolor[HTML]{d8d8d8}57.03
& \cellcolor[HTML]{d8d8d8}0.323 & \cellcolor[HTML]{d8d8d8}\textbf{0.948} & \cellcolor[HTML]{d8d8d8}53.92 \\

& \cellcolor[HTML]{d8d8d8}{$\text{FIZLE}$}      
& \cellcolor[HTML]{d8d8d8}0.585 & \cellcolor[HTML]{d8d8d8}0.584 & \cellcolor[HTML]{d8d8d8}38.71
& \cellcolor[HTML]{d8d8d8}0.236 & \cellcolor[HTML]{d8d8d8}0.584 & \cellcolor[HTML]{d8d8d8}77.38
& \cellcolor[HTML]{d8d8d8}\textbf{0.586} & \cellcolor[HTML]{d8d8d8}0.730 & \cellcolor[HTML]{d8d8d8}173.94
& \cellcolor[HTML]{d8d8d8}0.634 & \cellcolor[HTML]{d8d8d8}0.882 & \cellcolor[HTML]{d8d8d8}74.15 \\

& \cellcolor[HTML]{d8d8d8}{Causal What-Ifs} 
& \cellcolor[HTML]{d8d8d8}0.850 & \cellcolor[HTML]{d8d8d8}0.801 & \cellcolor[HTML]{d8d8d8}37.97
& \cellcolor[HTML]{d8d8d8}0.273 & \cellcolor[HTML]{d8d8d8}\textbf{0.607} & \cellcolor[HTML]{d8d8d8}\textbf{37.54}
& \cellcolor[HTML]{d8d8d8}0.443 & \cellcolor[HTML]{d8d8d8}0.854 & \cellcolor[HTML]{d8d8d8}\textbf{47.27}
& \cellcolor[HTML]{d8d8d8}0.627 & \cellcolor[HTML]{d8d8d8}0.911 & \cellcolor[HTML]{d8d8d8}\textbf{35.56} \\

\cmidrule(lr){2-14}
& \cellcolor[HTML]{D9EAFD}\textit{i}\texttt{Flip}-Conf
& 0.895 & 0.809 & 41.19
& \uwave{0.552} & 0.504 & \uwave{41.41}
& 0.440 & 0.854 & 53.52
& 0.620 & 0.870 & 40.65 \\

& \cellcolor[HTML]{E1F5E1}\textit{i}\texttt{Flip}-SHAP
& 0.906 & 0.805 & 41.87
& \uwave{0.552} & 0.489 & 42.95
& 0.441 & 0.847 & 52.76
& 0.594 & 0.867 & 42.38 \\

& \cellcolor[HTML]{E1F5E1}\textit{i}\texttt{Flip}-AttnLRP
& 0.923 & 0.803 & 42.61
& 0.536 & 0.501 & 46.61
& \uwave{0.458} & 0.851 & \uwave{51.89}
& 0.624 & 0.869 & \uwave{40.31} \\

& \cellcolor[HTML]{E1F5E1}\textit{i}\texttt{Flip}-Grad$\times$Input
& 0.912 & 0.801 & 43.30
& 0.516 & 0.499 & 50.32
& 0.432 & 0.854 & 52.93
& 0.634 & 0.871 & 41.81 \\

& \cellcolor[HTML]{E1F5E1}\textit{i}\texttt{Flip}-LIME
& 0.911 & 0.805 & 52.82
& 0.522 & 0.506 & 44.32
& 0.420 & 0.866 & 52.33
& 0.606 & 0.875 & 41.02 \\

& \cellcolor[HTML]{FFEACC}\textit{i}\texttt{Flip}-NL
& \uwave{\textbf{0.931}} & \uwave{0.822} & \uwave{\textbf{36.47}}
& 0.532 & \uwave{0.516} & 44.73
& 0.444 & \uwave{0.879} & 148.41
& \uwave{\textbf{0.636}} & \uwave{0.901} & 43.81 \\

\midrule
\multirow{9}{*}[-0.55cm]{\rotatebox[origin=c]{90}{\lm{Qwen3-32B}}}
& \cellcolor[HTML]{d8d8d8}{CGG}               
& \cellcolor[HTML]{d8d8d8}0.890 & \cellcolor[HTML]{d8d8d8}0.877 & \cellcolor[HTML]{d8d8d8}62.47
& \cellcolor[HTML]{d8d8d8}0.364 & \cellcolor[HTML]{d8d8d8}0.697 & \cellcolor[HTML]{d8d8d8}97.10
& \cellcolor[HTML]{d8d8d8}0.273 & \cellcolor[HTML]{d8d8d8}0.881 & \cellcolor[HTML]{d8d8d8}50.31
& \cellcolor[HTML]{d8d8d8}0.343 & \cellcolor[HTML]{d8d8d8}\textbf{0.938} & \cellcolor[HTML]{d8d8d8}47.94 \\

& \cellcolor[HTML]{d8d8d8}{$\text{FIZLE}$}      
& \cellcolor[HTML]{d8d8d8}0.686 & \cellcolor[HTML]{d8d8d8}\textbf{0.900} & \cellcolor[HTML]{d8d8d8}36.00
& \cellcolor[HTML]{d8d8d8}0.140 & \cellcolor[HTML]{d8d8d8}\textbf{0.864} & \cellcolor[HTML]{d8d8d8}61.46
& \cellcolor[HTML]{d8d8d8}\textbf{0.689} & \cellcolor[HTML]{d8d8d8}\textbf{0.901} & \cellcolor[HTML]{d8d8d8}38.66
& \cellcolor[HTML]{d8d8d8}\textbf{0.810} & \cellcolor[HTML]{d8d8d8}0.918 & \cellcolor[HTML]{d8d8d8}35.97 \\

& \cellcolor[HTML]{d8d8d8}{Causal What-Ifs}  
& \cellcolor[HTML]{d8d8d8}0.798 & \cellcolor[HTML]{d8d8d8}0.821 & \cellcolor[HTML]{d8d8d8}42.03
& \cellcolor[HTML]{d8d8d8}0.436 & \cellcolor[HTML]{d8d8d8}0.570 & \cellcolor[HTML]{d8d8d8}53.50
& \cellcolor[HTML]{d8d8d8}0.580 & \cellcolor[HTML]{d8d8d8}0.842 & \cellcolor[HTML]{d8d8d8}\textbf{36.14}
& \cellcolor[HTML]{d8d8d8}0.680 & \cellcolor[HTML]{d8d8d8}0.920 & \cellcolor[HTML]{d8d8d8}\textbf{34.18}\\

\cmidrule(lr){2-14}
& \cellcolor[HTML]{D9EAFD}\textit{i}\texttt{Flip}-Conf
& \uwave{\textbf{0.940}} & 0.855 & 33.30
& 0.622 & 0.521 & \uwave{\textbf{41.59}}
& 0.490 & \uwave{0.895} & 50.44
& 0.640 & 0.871 & 47.03 \\

& \cellcolor[HTML]{E1F5E1}\textit{i}\texttt{Flip}-SHAP
& 0.936 & 0.857 & 33.70
& 0.604 & 0.520 & 44.02
& 0.542 & 0.878 & 47.16
& 0.634 & 0.866 & 45.04 \\

& \cellcolor[HTML]{E1F5E1}\textit{i}\texttt{Flip}-AttnLRP
& 0.934 & 0.858 & 33.19
& \uwave{\textbf{0.626}} & 0.513 & 44.15
& 0.532 & 0.882 & \uwave{46.93}
& 0.648 & 0.867 & 44.77 \\

& \cellcolor[HTML]{E1F5E1}\textit{i}\texttt{Flip}-Grad$\times$Input
& 0.934 & 0.865 & 33.65
& 0.592 & 0.522 & 44.33
& 0.528 & 0.881 & 47.40
& 0.658 & 0.865 & 45.60 \\

& \cellcolor[HTML]{E1F5E1}\textit{i}\texttt{Flip}-LIME
& 0.938 & 0.863 & \uwave{\textbf{32.57}}
& 0.624 & 0.527 & 44.52
& 0.498 & 0.891 & 50.60
& 0.654 & 0.870 & \uwave{44.22} \\

& \cellcolor[HTML]{FFEACC}\textit{i}\texttt{Flip}-NL
& 0.922 & \uwave{0.878} & 32.99
& 0.618 & \uwave{0.537} & 49.68
& \uwave{0.546} & \uwave{0.895} & 68.16
& \uwave{0.668} & \uwave{0.906} & 47.20 \\

\midrule
\multirow{9}{*}[-0.5cm]{\rotatebox[origin=c]{90}{\lm{LLaMA3.3-70B}}}
& \cellcolor[HTML]{d8d8d8}{CGG}               
& \cellcolor[HTML]{d8d8d8}0.851 & \cellcolor[HTML]{d8d8d8}0.869 & \cellcolor[HTML]{d8d8d8}52.81
& \cellcolor[HTML]{d8d8d8}0.516 & \cellcolor[HTML]{d8d8d8}0.694 & \cellcolor[HTML]{d8d8d8}122.40
& \cellcolor[HTML]{d8d8d8}0.357 & \cellcolor[HTML]{d8d8d8}0.846 & \cellcolor[HTML]{d8d8d8}51.62
& \cellcolor[HTML]{d8d8d8}0.412 & \cellcolor[HTML]{d8d8d8}\textbf{0.937} & \cellcolor[HTML]{d8d8d8}46.58 \\

& \cellcolor[HTML]{d8d8d8}{$\text{FIZLE}$} 
& \cellcolor[HTML]{d8d8d8}\textbf{0.956} & \cellcolor[HTML]{d8d8d8}0.868 & \cellcolor[HTML]{d8d8d8}34.88
& \cellcolor[HTML]{d8d8d8}0.345 & \cellcolor[HTML]{d8d8d8}\textbf{0.707} & \cellcolor[HTML]{d8d8d8}56.59
& \cellcolor[HTML]{d8d8d8}\textbf{0.759} & \cellcolor[HTML]{d8d8d8}\textbf{0.903} & \cellcolor[HTML]{d8d8d8}41.54
& \cellcolor[HTML]{d8d8d8}\textbf{0.821} & \cellcolor[HTML]{d8d8d8}0.929 & \cellcolor[HTML]{d8d8d8}42.39 \\

& \cellcolor[HTML]{d8d8d8}{Causal What-Ifs} 
& \cellcolor[HTML]{d8d8d8}0.852 & \cellcolor[HTML]{d8d8d8}0.788 & \cellcolor[HTML]{d8d8d8}\textbf{31.93}
& \cellcolor[HTML]{d8d8d8}0.628 & \cellcolor[HTML]{d8d8d8}0.461 & \cellcolor[HTML]{d8d8d8}\textbf{43.56}
& \cellcolor[HTML]{d8d8d8}0.528 & \cellcolor[HTML]{d8d8d8}0.816 & \cellcolor[HTML]{d8d8d8}\textbf{38.97}
& \cellcolor[HTML]{d8d8d8}0.560 & \cellcolor[HTML]{d8d8d8}0.904 & \cellcolor[HTML]{d8d8d8}41.29 \\

\cmidrule(lr){2-14}
& \cellcolor[HTML]{D9EAFD}\textit{i}\texttt{Flip}-Conf
& 0.944 & 0.877 & 34.74
& 0.804 & 0.468 & 44.38
& 0.654 & 0.763 & 45.99
& 0.678 & 0.843 & 43.88 \\

& \cellcolor[HTML]{E1F5E1}\textit{i}\texttt{Flip}-SHAP
& 0.944 & 0.878 & \uwave{34.32}
& 0.794 & 0.481 & 46.07
& 0.652 & 0.743 & 45.15
& 0.678 & 0.828 & 41.09 \\

& \cellcolor[HTML]{E1F5E1}\textit{i}\texttt{Flip}-AttnLRP
& \uwave{0.946} & 0.879 & 34.58
& \uwave{\textbf{0.816}} & 0.477 & 45.78
& 0.642 & 0.743 & 44.66
& 0.684 & 0.822 & 39.92 \\

& \cellcolor[HTML]{E1F5E1}\textit{i}\texttt{Flip}-Grad$\times$Input
& 0.942 & 0.878 & 34.80
& 0.814 & 0.479 & 46.12
& \uwave{0.666} & 0.751 & \uwave{44.14}
& 0.696 & 0.825 & \uwave{\textbf{39.55}} \\

& \cellcolor[HTML]{E1F5E1}\textit{i}\texttt{Flip}-LIME
& 0.944 & 0.877 & 34.54
& 0.796 & 0.476 & \uwave{44.20}
& 0.654 & 0.759 & 45.51
& 0.680 & 0.825 & 42.91 \\

& \cellcolor[HTML]{FFEACC}\textit{i}\texttt{Flip}-NL
& 0.942 & \uwave{\textbf{0.895}} & 36.90
& 0.782 & \uwave{0.527} & 63.86
& 0.624 & \uwave{0.821} & 69.45
& \uwave{0.726} & \uwave{0.861} & 49.74 \\

\bottomrule
\end{tabular}
}

\caption{Automatic evaluation results of counterfactual generated for \lm{BERT} models, and evaluated on \lm{RoBERTa} classifiers.
Baselines {\setlength{\fboxsep}{1pt}\colorbox{gray}{(CGG, FIZLE and Causal What-Ifs)}} are compared with our \textit{i}\texttt{Flip} variants under different feedback types:
\ding{182} confidence (\mbox{{\setlength{\fboxsep}{1pt}\colorbox[HTML]{D9EAFD}{\strut\textsf{Conf}}}}),
\ding{183} feature attribution (\mbox{{\setlength{\fboxsep}{1pt}\colorbox[HTML]{E1F5E1}{\strut\textsf{SHAP}}}}, \mbox{{\setlength{\fboxsep}{1pt}\colorbox[HTML]{E1F5E1}{\strut\textsf{AttnLRP}}}}, \mbox{{\setlength{\fboxsep}{1pt}\colorbox[HTML]{E1F5E1}{\strut\textsf{Grad$\times$Input}}}}, \mbox{{\setlength{\fboxsep}{1pt}\colorbox[HTML]{E1F5E1}{\strut\textsf{LIME}}}}),
and \ding{184} natural language (\mbox{{\setlength{\fboxsep}{1pt}\colorbox[HTML]{FFEACC}{\strut\textsf{NL}}}}).
Results are reported on \data{IMDb}, \data{AG News}, and \data{SNLI} using Label Flipping Rate (LFR), Semantic Similarity (SS), and Perplexity (PPL).
\textbf{Boldface} indicates the best result across methods. \uwave{Wavy underline} indicates the best feedback type within \textit{i}\texttt{Flip}.}
\label{tab:transfer_results1}
\end{table*}

\subsubsection{Transferability of the feedback extracted from \lm{BERT}}

Table~\ref{tab:transfer_results2} evaluates the \textit{transferability} of feedback signals extracted from \lm{BERT}. We use these signals to guide \textit{i}\texttt{Flip} in generating counterfactuals, and then evaluate the resulting counterfactuals with \lm{RoBERTa}.
\begin{table*}[t!]
    \centering
    \renewcommand*{\arraystretch}{0.5}
    \footnotesize
    \resizebox{\textwidth}{!}{%
\begin{tabular}{ll|ccc|ccc|ccc|ccc}
\toprule
\textbf{} & \textbf{Method}
& \multicolumn{3}{c|}{\textbf{\data{IMDb}}}
& \multicolumn{3}{c|}{\textbf{\data{AG News}}}
& \multicolumn{3}{c|}{\makecell{\textbf{\data{SNLI}} \textbf{(Premise)}}}
& \multicolumn{3}{c}{\makecell{\textbf{\data{SNLI}} \textbf{(Hypothesis)}}}\\
\cmidrule(lr){3-5}\cmidrule(lr){6-8}\cmidrule(lr){9-11}\cmidrule(lr){12-14}
 & & LFR$\uparrow$ & SS$\uparrow$ & PPL$\downarrow$
   & LFR$\uparrow$ & SS$\uparrow$ & PPL$\downarrow$
   & LFR$\uparrow$ & SS$\uparrow$ & PPL$\downarrow$
   & LFR$\uparrow$ & SS$\uparrow$ & PPL$\downarrow$\\
\midrule

\multirow{5}{*}[0cm]{%
  \rotatebox[origin=c]{90}{\makecell{\lm{LLaMA}\\\lm{3.3-70B}}}%
}
& \cellcolor[HTML]{D9EAFD}\textit{i}\texttt{Flip}-Conf
& 0.995 & 0.897 & 37.17
& 0.896 & \textbf{0.539} & 58.65
& 0.672 & 0.702 & 62.81
& 0.816 & 0.792 & 46.84 \\

& \cellcolor[HTML]{E1F5E1}\textit{i}\texttt{Flip}-SHAP
& \textbf{1.000} & \textbf{0.897} & 37.12
& 0.905 & 0.516 & 57.88
& \textbf{0.677} & 0.697 & 62.30
& 0.831 & 0.790 & \textbf{44.08} \\

& \cellcolor[HTML]{E1F5E1}\textit{i}\texttt{Flip}-AttnLRP
& \textbf{1.000} & 0.895 & \textbf{36.66}
& 0.876 & 0.524 & 59.95
& \textbf{0.677} & \textbf{0.706} & 56.80
& 0.841 & \textbf{0.796} & 45.66 \\

& \cellcolor[HTML]{E1F5E1}\textit{i}\texttt{Flip}-Grad$\times$Input
& \textbf{1.000} & \textbf{0.897} & 36.78
& \textbf{0.910} & 0.518 & 59.34
& 0.672 & 0.699 & \textbf{55.40}
& 0.811 & 0.792 & 44.98 \\

& \cellcolor[HTML]{E1F5E1}\textit{i}\texttt{Flip}-LIME
& \textbf{1.000} & \textbf{0.898} & 36.92
& 0.905 & 0.509 & \textbf{57.58}
& 0.662 & 0.704 & 55.88
& \textbf{0.846} & 0.788 & 45.36 \\

\bottomrule
\end{tabular}
}
\caption{Automatic evaluation results of counterfactual generated by \textit{i}\texttt{Flip} using feedback extracted from \lm{BERT} models, and evaluated on \lm{RoBERTa} models.
We report results on \data{IMDb}, \data{AG News}, and \data{SNLI} using Label Flipping Rate (LFR), Semantic Similarity (SS), and Perplexity (PPL).
Feedback types include \ding{182} confidence (\mbox{{\setlength{\fboxsep}{1pt}\colorbox[HTML]{D9EAFD}{\strut\textsf{Conf}}}}),
\ding{183} feature attribution (\mbox{{\setlength{\fboxsep}{1pt}\colorbox[HTML]{E1F5E1}{\strut\textsf{SHAP}}}}, \mbox{{\setlength{\fboxsep}{1pt}\colorbox[HTML]{E1F5E1}{\strut\textsf{AttnLRP}}}}, \mbox{{\setlength{\fboxsep}{1pt}\colorbox[HTML]{E1F5E1}{\strut\textsf{Grad$\times$Input}}}} and \mbox{{\setlength{\fboxsep}{1pt}\colorbox[HTML]{E1F5E1}{\strut\textsf{LIME}}}}). \textbf{Boldface} indicates  the best feedback type within \textit{i}\texttt{Flip}.}
\label{tab:transfer_results2}
\end{table*}

\subsubsection{Discussion}
\paragraph{Early stopping is the key factor for transferability.}
Regarding transferability, early stopping emerges as the primary driver of robustness across architectures. In Table~\ref{tab:transfer_results1}, when counterfactuals generated for \lm{BERT} are evaluated on \lm{RoBERTa}, methods employing early stopping (e.g., \texttt{iFlip} and Causal What-Ifs) exhibit substantial validity reductions compared to counterfactuals generated directly for \lm{RoBERTa} (Table~\ref{tab:roberta_results}). This pattern indicates that early stopping enhances transferability by providing the LLM with direct validity feedback on whether candidate counterfactuals successfully flip classifier predictions.

\paragraph{The source of the feedback has limited impact on transferability.}
Table~\ref{tab:transfer_results2} suggests that the \emph{source} of the feedback (extracted from \lm{BERT} or \lm{RoBERTa}) is comparatively less influential on transferability: using feedback from different models yields only small performance differences, and the resulting counterfactuals achieve similar quality to those guided by \lm{RoBERTa} feedback (Table~\ref{tab:roberta_results}). Nevertheless, feedback is still beneficial for counterfactual generation, where our ablation results showing that removing feedback leads to worse performance than using feedback (see \S\ref{subsubsec:feedback}).

\subsubsection{Summary of experimental settings}
\label{subsubsec:transfer_settings}
Table~\ref{tab:settings_overview} summarizes the experimental settings used in Tables~\ref{tab:roberta_results}, \ref{tab:transfer_results1}, and \ref{tab:transfer_results2}, including (i) the \emph{feedback source} model, (ii) the model used for \emph{early stopping} (validity check during generation), and (iii) the \emph{evaluation} classifier.

\begin{table*}[t]
\centering
\small
\setlength{\tabcolsep}{8pt}
\begin{tabular}{lccc}
\hline
Table & Feedback source & Early stop under & Evaluation model \\
\hline
Table~\ref{tab:roberta_results} & \lm{RoBERTa-base} & \lm{RoBERTa-base} & \lm{RoBERTa-base} \\
Table~\ref{tab:transfer_results1} & \lm{BERT} & \lm{BERT} & \lm{RoBERTa-base} \\
Table~\ref{tab:transfer_results2} & \lm{BERT} & \lm{RoBERTa-base} & \lm{RoBERTa-base} \\
\hline
\end{tabular}
\caption{Overview of experimental settings of \textit{i}\texttt{Flip} across tables: feedback source, early-stopping model, and evaluation classifier.}
\label{tab:settings_overview}
\end{table*}

\section{Feedback Signal Comparison}
\label{app:feedback_analysis}

\begin{figure}[t]
    \centering
    \begin{adjustbox}{max width=\columnwidth}
        \includegraphics{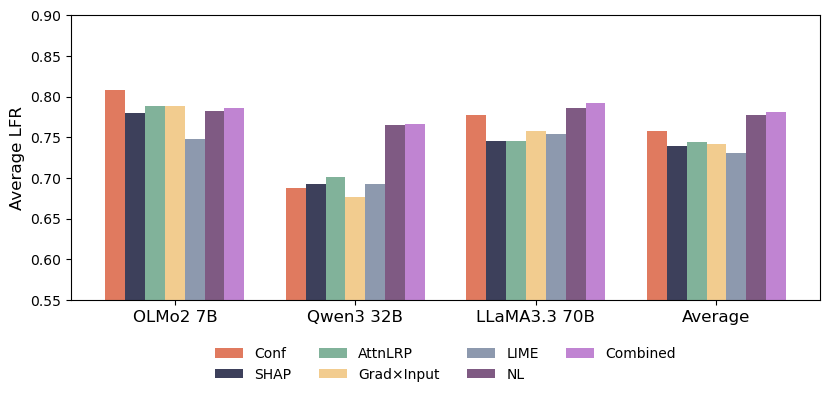}
    \end{adjustbox}
    \caption{Averaged label flipping rate (LFR) across feedback signals and models on the \data{IMDb}, \data{AG News}, and \data{SNLI} datasets.}
    \label{fig:feedback_fr_per_model}
\end{figure}

\begin{figure}[t]
    \centering
    \begin{adjustbox}{max width=\columnwidth}
        \includegraphics{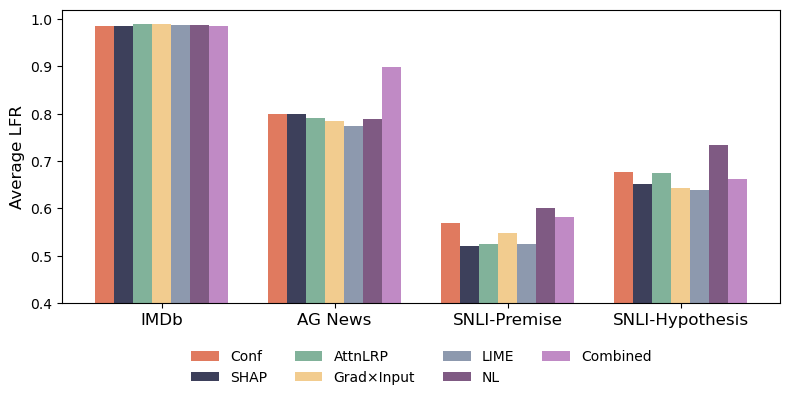}
    \end{adjustbox}
    \caption{Average label flipping rate (LFR) across feedback signals and tasks on the \data{IMDb}, \data{AG News}, and \data{SNLI} datasets.}
    \label{fig:feedback_fr_per_task}
\end{figure}

We compare different feedback signals in terms of their label flipping rate (LFR) across models and tasks.  
Figure~\ref{fig:feedback_fr_per_model} shows the results across models, and Figure~\ref{fig:feedback_fr_per_task} shows the results across tasks.

\section{Extended Ablation Results}

\subsection{Iterative Refinement}
\label{app:analysis_iter_rounds}

\subsubsection{Average Early Stop Rounds}
\label{app:analysis_iter_rounds1}

Table~\ref{tab:early_stop_rounds} reports the average number of refinement
rounds before early stopping across datasets and models.

\begin{table}[h]
\centering
\renewcommand\theadfont{\normalsize\bfseries}
\begin{adjustbox}{max width=\linewidth}
\begin{tabular}{lccc}
\toprule
\textbf{Dataset} & \textbf{\lm{OLMo2-7B}} & \textbf{\lm{Qwen3-32B}} & \textbf{\lm{LLaMA3.3-70B}} \\
\midrule
\textbf{IMDb}            & 0.13 & 0.05 & 0.01 \\
\textbf{AG News}         & 1.48 & 1.53 & 0.80 \\
\textbf{SNLI (Premise)}  & 2.20 & 3.20 & 2.75 \\
\textbf{SNLI (Hypothesis)} & 1.83 & 2.62 & 2.41 \\
\bottomrule
\end{tabular}
\end{adjustbox}
\caption{Average early stop rounds per dataset and model.}
\label{tab:early_stop_rounds}
\end{table}

\subsubsection{Counterfactual Length across Iteration Rounds}
\label{app:analysis_iter_rounds2}

\begin{figure}[h]
\centering
\includegraphics[width=\columnwidth]{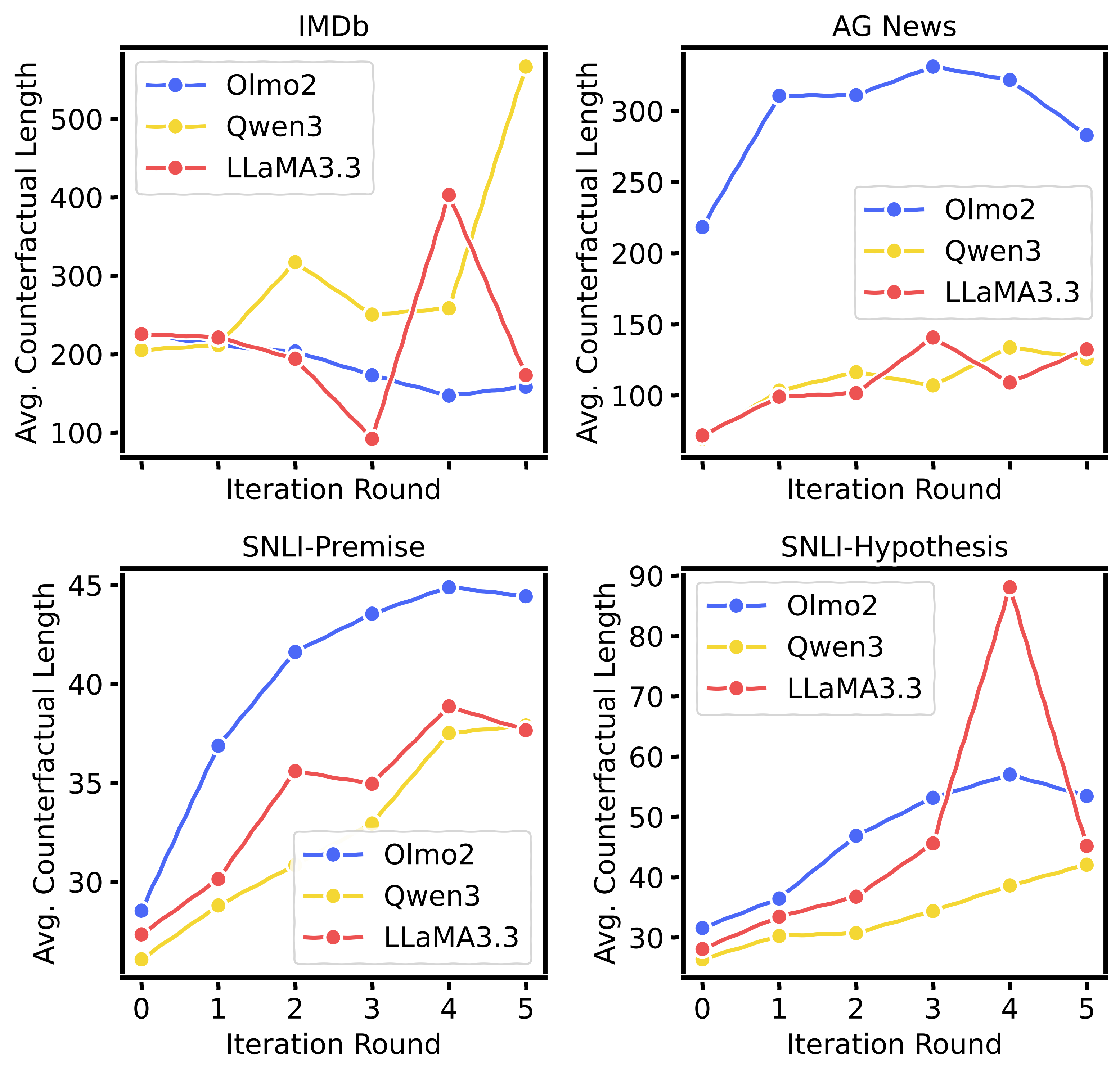}
\caption{Average counterfactual length vs. iteration round across tasks and models.}
\label{fig:cfe_length_rounds}
\end{figure}

Figure~\ref{fig:cfe_length_rounds} illustrates how the average length of
generated counterfactuals evolves over refinement iterations.

\subsubsection{Trade-off between Validity and Similarity}
\label{app:tradeoff_ts_lfr}
\begin{figure}[h]
\centering
\includegraphics[width=\columnwidth]{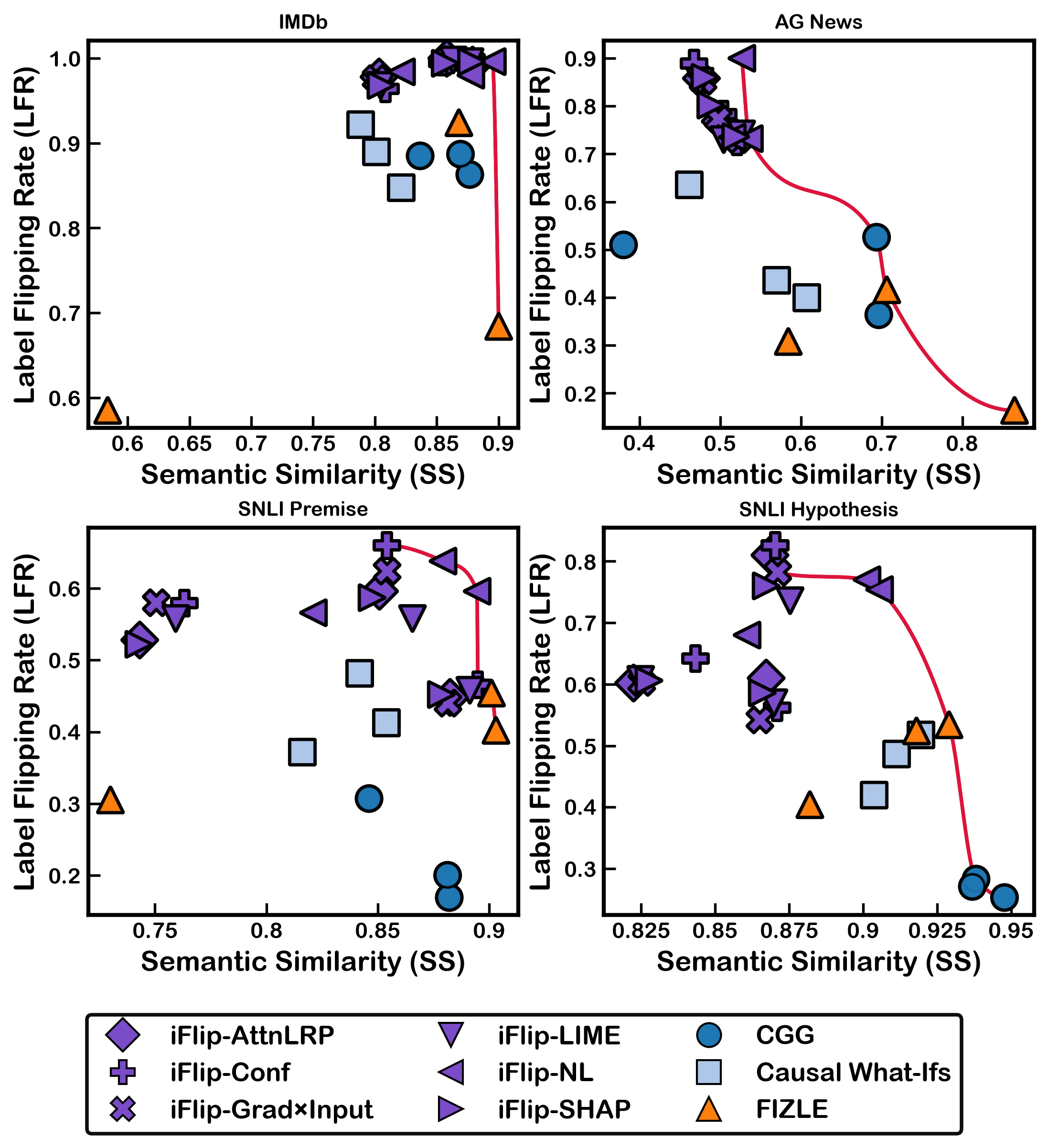}
\caption{Trade-off between semantic similarity and label flipping rate.  The upper-right region indicates the most favorable trade-off. The red curve denotes the Pareto frontier.}
\label{fig:tradeoff}
\end{figure}

\begin{figure}[h]
\centering

\includegraphics[width=\columnwidth]{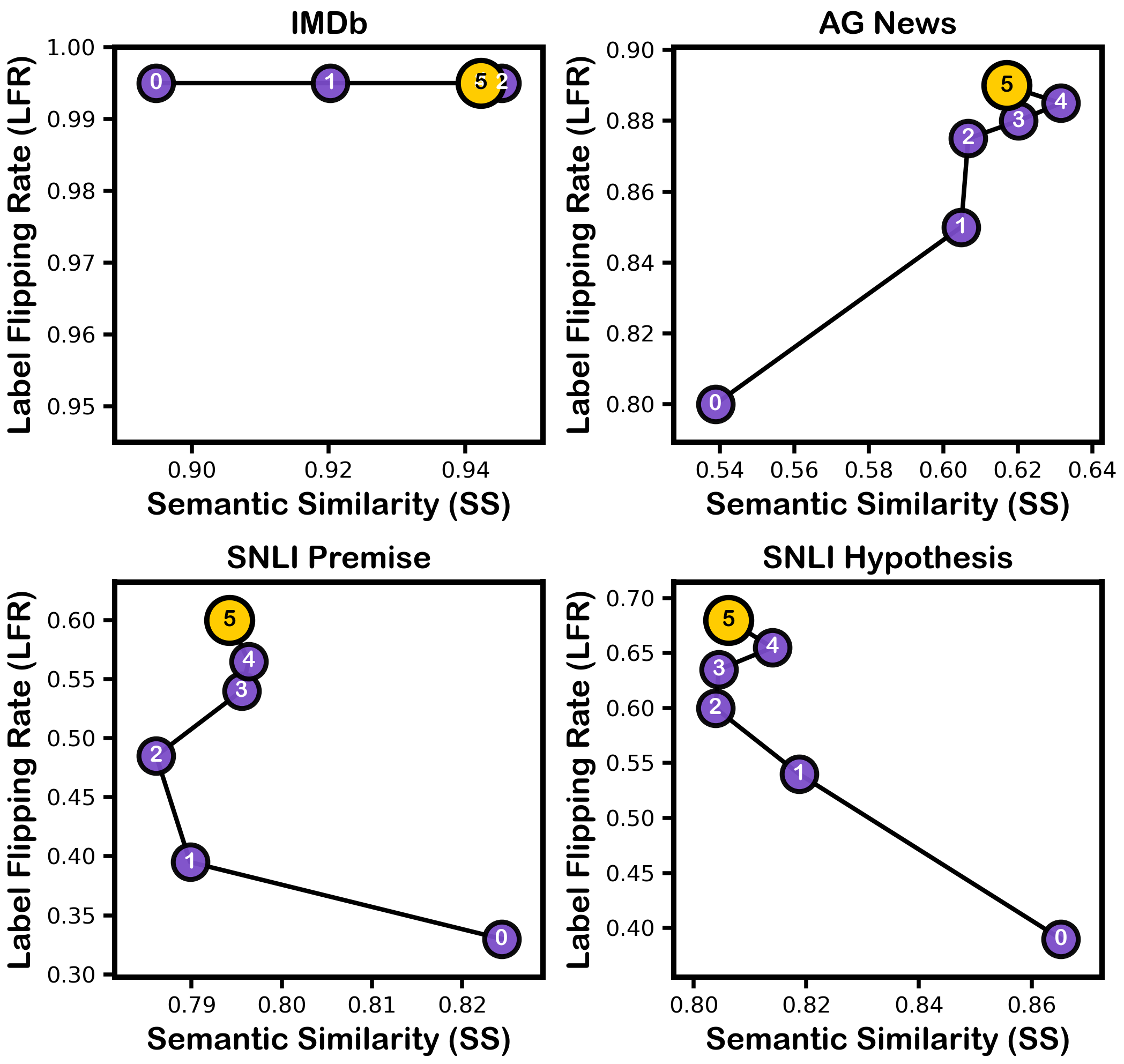}
\caption{Trade-off between semantic similarity and label flipping rate. Markers show iteration $\mathcal{K}$. Results are from \textit{i}\texttt{Flip}-NL using \lm{LLaMA3.3-70B}.}
\label{fig:tradeoff_k}
\end{figure}

Figure~\ref{fig:tradeoff} plots the trade-off between semantic similarity (SS) and label flipping rate (LFR). Figure~\ref{fig:tradeoff_k} further breaks down this trade-off by showing how the SS-LFR trajectory evolves across \textit{i}\texttt{Flip}-NL iteration steps $k$ on each task. 
After $\mathcal{K}=4$, the degradation in semantic similarity accelerates, so for practical applications requiring high fidelity, $\mathcal{K}=3$ or $\mathcal{K}=4$ may offer a better balance.

\subsubsection{Iterative Refinement Effectiveness}
\label{app:iter_effect_full}

\begin{figure*}[!t]
    \centering
    \begin{adjustbox}{max width=\textwidth}
    \begin{tabular}{cccc}
        \includegraphics[width=\textwidth]{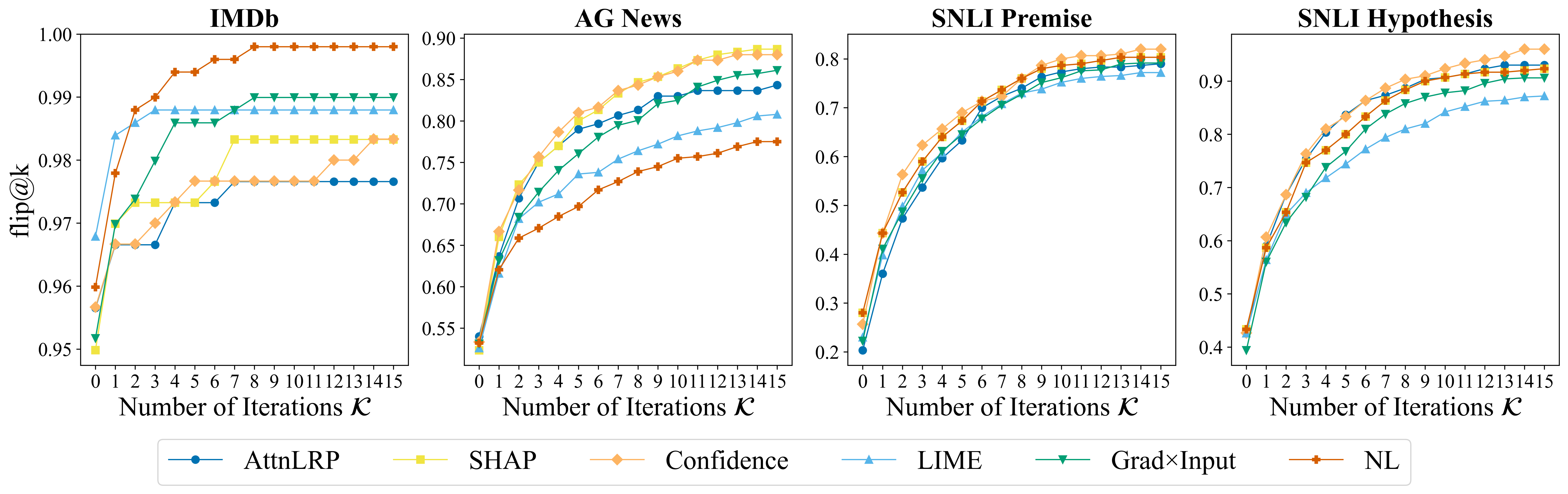}
    \end{tabular}
    \end{adjustbox}
    \caption{Iterative refinement effectiveness on \lm{OLMo2-7B}. \textit{flip@k} curves for each dataset (\data{IMDb}, \data{AG News}, \data{SNLI}), 
evaluated up to $\mathcal{K}=15$ refinement steps.}
    \label{fig:passk}
\end{figure*}

\begin{figure*}[!t]
    \centering
    \includegraphics[width=1\linewidth]{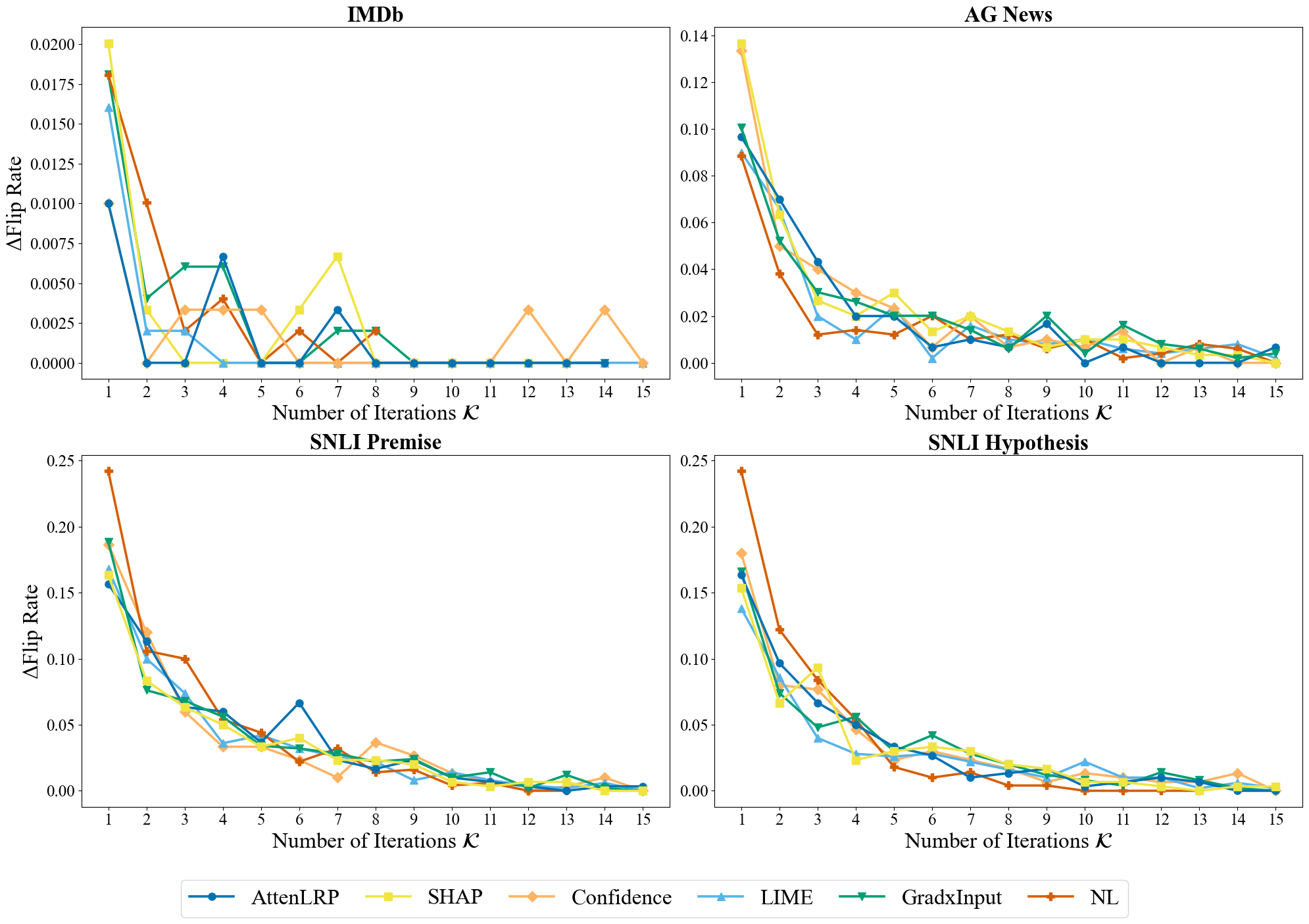}
    \caption{Change in flip rate ($\Delta$Flip Rate) across iterative refinement rounds with \lm{OLMo2-7B} for different feedback signals. The results show how much additional gain each refinement step contributes beyond the previous round.}

    \label{fig:delta_flip}
\end{figure*}

Figures~\ref{fig:passk} and \ref{fig:delta_flip}  show that iterative refinement with \lm{OLMo2-7B} yields steady \textit{flip@k} gains across \data{IMDb}, \data{AG News}, and \data{SNLI} up to \(k=15\), while the largest marginal improvements in flip rate occur in the early rounds and taper off thereafter depending on the feedback signal.

\subsection{Comparison with Baselines}

Table~\ref{tab:appendix_compare_baselines} summarizes the main differences between \textit{i}\texttt{Flip} and the baselines.

\begin{table*}[t]
\centering
\small
\setlength{\tabcolsep}{5pt}
\begin{tabular}{lccccc}
\toprule
Method & Iterative & Feedback & Max iters. & Early stopping & Cost \\
\midrule
FIZLE & $\times$ & $\times$ & 1 & -- & Low \\
CGG & $\times$ & $\checkmark$ & 1 & -- & Low \\
Causal What-Ifs & $\checkmark$ & $\times$ & 5 & $\times$ & High \\
\textit{i}\texttt{Flip} & $\checkmark$ & $\checkmark$ & 1--5 & $\checkmark$ & Moderate \\
\bottomrule
\end{tabular}
\caption{Comparison of \textit{i}\texttt{Flip} and the baselines.}
\label{tab:appendix_compare_baselines}
\end{table*}

\subsection{Feedback Signals}
\label{app:extra_ablation_feedback_signal}

Table~\ref{tab:ablation_feedback1} reports the evaluation results of our feedback signal ablations on \data{IMDb} and \data{AG News}, while Table~\ref{tab:ablation_feedback2} presents the corresponding results on \data{SNLI}.

\begin{table*}[!t]
\centering
\renewcommand\theadfont{\normalsize\bfseries}
\begin{adjustbox}{max width=\textwidth}
\begin{tabular}{ll|ccc|ccc}
\toprule
\textbf{} & \textbf{\makecell{Feedback of\\ \textit{i}\texttt{Flip}}}
& \multicolumn{3}{c|}{\textbf{\data{IMDb}}}
& \multicolumn{3}{c}{\textbf{\data{AG News}}} \\
\cmidrule(lr){3-5}\cmidrule(lr){6-8}
 & & LFR$\uparrow$ & SS$\uparrow$ & PPL$\downarrow$ & LFR$\uparrow$ & SS$\uparrow$ & PPL$\downarrow$ \\
\midrule
\multirow{13}{*}{\rotatebox{90}{\lm{OLMo2-7B}}}
 & \cellcolor[HTML]{C3F2F2}Without Feedback   & \cellcolor[HTML]{C3F2F2}0.964 & \cellcolor[HTML]{C3F2F2}0.803 & \cellcolor[HTML]{C3F2F2}39.12 & \cellcolor[HTML]{C3F2F2}0.742 & \cellcolor[HTML]{C3F2F2}0.494 & \cellcolor[HTML]{C3F2F2}44.71 \\
 \midrule
 & \cellcolor[HTML]{FFF8B8}Random Feedback    & 0.970(\textcolor{morandiRed}{$\uparrow$0.006}) & 0.803(---) & 44.58(\textcolor{morandiRed}{$\uparrow$5.46})
                      & 0.794(\textcolor{morandiRed}{$\uparrow$0.052}) & 0.498(\textcolor{morandiRed}{$\uparrow$0.004}) & 43.59(\textcolor{morandiGreen}{$\downarrow$1.12}) \\

 & \cellcolor[HTML]{E8D8FF}Least SHAP         & 0.970(\textcolor{morandiRed}{$\uparrow$0.006}) & 0.802(\textcolor{morandiGreen}{$\downarrow$0.001}) & 41.10(\textcolor{morandiRed}{$\uparrow$1.98})
                      & 0.770(\textcolor{morandiRed}{$\uparrow$0.028}) & 0.490(\textcolor{morandiGreen}{$\downarrow$0.004}) & 41.78(\textcolor{morandiGreen}{$\downarrow$2.93}) \\
 & \cellcolor[HTML]{E8D8FF}Least AttnLRP      & 0.980(\textcolor{morandiRed}{$\uparrow$0.016}) & 0.805(\textcolor{morandiRed}{$\uparrow$0.002}) & 41.34(\textcolor{morandiRed}{$\uparrow$2.22})
                      & 0.776(\textcolor{morandiRed}{$\uparrow$0.034}) & 0.495(\textcolor{morandiRed}{$\uparrow$0.001}) & 40.74(\textcolor{morandiGreen}{$\downarrow$3.97}) \\
 & \cellcolor[HTML]{E8D8FF}Least Grad$\times$Input & 0.968(\textcolor{morandiRed}{$\uparrow$0.004}) & 0.803(---) & 43.02(\textcolor{morandiRed}{$\uparrow$3.90})
                      & 0.758(\textcolor{morandiRed}{$\uparrow$0.016}) & 0.495(\textcolor{morandiRed}{$\uparrow$0.001}) & 49.09(\textcolor{morandiRed}{$\uparrow$4.38}) \\
 & \cellcolor[HTML]{E8D8FF}Least LIME         & 0.968(\textcolor{morandiRed}{$\uparrow$0.004}) & 0.800(\textcolor{morandiGreen}{$\downarrow$0.003}) & 44.99(\textcolor{morandiRed}{$\uparrow$5.87})
                      & 0.725(\textcolor{morandiGreen}{$\downarrow$0.017}) & 0.508(\textcolor{morandiRed}{$\uparrow$0.014}) & 43.53(\textcolor{morandiGreen}{$\downarrow$1.18}) \\

\cmidrule(lr){2-8}
 & \cellcolor[HTML]{D9EAFD}Conf          & 0.964(---) & 0.809(\textcolor{morandiRed}{$\uparrow$0.006}) & 41.19(\textcolor{morandiRed}{$\uparrow$2.07})
                      & 0.784(\textcolor{morandiRed}{$\uparrow$0.042}) & 0.503(\textcolor{morandiRed}{$\uparrow$0.009}) & 41.41(\textcolor{morandiGreen}{$\downarrow$3.30}) \\
 & \cellcolor[HTML]{E1F5E1}SHAP          & 0.968(\textcolor{morandiRed}{$\uparrow$0.004}) & 0.805(\textcolor{morandiRed}{$\uparrow$0.002}) & 41.87(\textcolor{morandiRed}{$\uparrow$2.75})
                      & 0.802(\textcolor{morandiRed}{$\uparrow$0.060}) & 0.488(\textcolor{morandiGreen}{$\downarrow$0.006}) & 42.95(\textcolor{morandiGreen}{$\downarrow$1.76}) \\
 & \cellcolor[HTML]{E1F5E1}AttnLRP       & 0.978(\textcolor{morandiRed}{$\uparrow$0.014}) & 0.803(---) & 42.61(\textcolor{morandiRed}{$\uparrow$3.49})
                      & 0.768(\textcolor{morandiRed}{$\uparrow$0.026}) & 0.500(\textcolor{morandiRed}{$\uparrow$0.006}) & 46.61(\textcolor{morandiRed}{$\uparrow$1.90}) \\
 & \cellcolor[HTML]{E1F5E1}Grad$\times$Input & 0.976(\textcolor{morandiRed}{$\uparrow$0.012}) & 0.801(\textcolor{morandiGreen}{$\downarrow$0.002}) & 43.30(\textcolor{morandiRed}{$\uparrow$4.18})
                      & 0.773(\textcolor{morandiRed}{$\uparrow$0.031}) & 0.498(\textcolor{morandiRed}{$\uparrow$0.004}) & 50.32(\textcolor{morandiRed}{$\uparrow$5.61}) \\
 & \cellcolor[HTML]{E1F5E1}LIME          & 0.968(\textcolor{morandiRed}{$\uparrow$0.004}) & 0.805(\textcolor{morandiRed}{$\uparrow$0.002}) & 52.82(\textcolor{morandiRed}{$\uparrow$13.70})
                      & 0.730(\textcolor{morandiGreen}{$\downarrow$0.012}) & 0.504(\textcolor{morandiRed}{$\uparrow$0.010}) & 44.32(\textcolor{morandiGreen}{$\downarrow$0.39}) \\
 & \cellcolor[HTML]{FFEACC}NL & 0.984(\textcolor{morandiRed}{$\uparrow$0.020}) & 0.822(\textcolor{morandiRed}{$\uparrow$0.019}) & 36.47(\textcolor{morandiGreen}{$\downarrow$2.65})
             & 0.735(\textcolor{morandiGreen}{$\downarrow$0.007}) & 0.515(\textcolor{morandiRed}{$\uparrow$0.021}) & 44.73(\textcolor{morandiRed}{$\uparrow$0.02}) \\

\bottomrule
\end{tabular}
\end{adjustbox}

\caption{
Automatic evaluation results of \lm{OLMo2-7B} on \data{IMDb} and \data{AG News}.
The tables present an ablation study of different feedback settings, including
\mbox{{\setlength{\fboxsep}{1pt}\colorbox[HTML]{C3F2F2}{\textit{Without Feedback}}}},
\mbox{{\setlength{\fboxsep}{1pt}\colorbox[HTML]{FFF8B8}{\textit{Random Feedback}}}},
\mbox{{\setlength{\fboxsep}{1pt}\colorbox[HTML]{E8D8FF}{\textit{Least-Attributed Feedback}}}}, and the full variants of \textit{i}\texttt{Flip}.
Colored values indicate relative changes compared to the \mbox{{\setlength{\fboxsep}{1pt}\colorbox[HTML]{C3F2F2}{\textit{Without Feedback} }}} baseline
(\textcolor{morandiRed}{$\uparrow$} = increase, \textcolor{morandiGreen}{$\downarrow$} = decrease).
}
\label{tab:ablation_feedback1}
\end{table*}

\begin{table*}[!t]
\centering
\renewcommand\theadfont{\normalsize\bfseries}
\begin{adjustbox}{max width=\textwidth}
\begin{tabular}{ll|ccc|ccc}
\toprule
\textbf{} & \textbf{\makecell{Feedback of\\ \textit{i}\texttt{Flip}}}
& \multicolumn{3}{c|}{\makecell{\textbf{\data{SNLI}}\\\textbf{(Premise)}}}
& \multicolumn{3}{c}{\makecell{\textbf{\data{SNLI}}\\\textbf{(Hypothesis)}}}\\
\cmidrule(lr){3-5}\cmidrule(lr){6-8}
 & & LFR$\uparrow$ & SS$\uparrow$ & PPL$\downarrow$ & LFR$\uparrow$ & SS$\uparrow$ & PPL$\downarrow$\\
\midrule
\multirow{13}{*}{\rotatebox{90}{\lm{OLMo2-7B}}}
 & \cellcolor[HTML]{C3F2F2}Without Feedback
 & \cellcolor[HTML]{C3F2F2}0.508 & \cellcolor[HTML]{C3F2F2}0.872 & \cellcolor[HTML]{C3F2F2}66.42
 & \cellcolor[HTML]{C3F2F2}0.778 & \cellcolor[HTML]{C3F2F2}0.879 & \cellcolor[HTML]{C3F2F2}41.61 \\

\midrule

 & \cellcolor[HTML]{FFF8B8}Random Feedback
 & 0.672(\textcolor{morandiRed}{$\uparrow$0.164}) & 0.865(\textcolor{morandiGreen}{$\downarrow$0.007}) & 53.64(\textcolor{morandiGreen}{$\downarrow$12.78})
 & 0.764(\textcolor{morandiGreen}{$\downarrow$0.014}) & 0.870(\textcolor{morandiGreen}{$\downarrow$0.009}) & 40.25(\textcolor{morandiGreen}{$\downarrow$1.36}) \\

 & \cellcolor[HTML]{E8D8FF}Least SHAP
 & 0.612(\textcolor{morandiRed}{$\uparrow$0.104}) & 0.854(\textcolor{morandiGreen}{$\downarrow$0.018}) & 54.51(\textcolor{morandiGreen}{$\downarrow$11.91})
 & 0.780(\textcolor{morandiRed}{$\uparrow$0.002}) & 0.874(\textcolor{morandiGreen}{$\downarrow$0.005}) & 41.30(\textcolor{morandiGreen}{$\downarrow$0.31}) \\

 & \cellcolor[HTML]{E8D8FF}Least AttnLRP
 & 0.562(\textcolor{morandiRed}{$\uparrow$0.054}) & 0.860(\textcolor{morandiGreen}{$\downarrow$0.012}) & 55.31(\textcolor{morandiGreen}{$\downarrow$11.11})
 & 0.764(\textcolor{morandiGreen}{$\downarrow$0.014}) & 0.869(\textcolor{morandiGreen}{$\downarrow$0.010}) & 40.58(\textcolor{morandiGreen}{$\downarrow$1.03}) \\

 & \cellcolor[HTML]{E8D8FF}Least Grad$\times$Input
 & 0.593(\textcolor{morandiRed}{$\uparrow$0.085}) & 0.868(\textcolor{morandiGreen}{$\downarrow$0.004}) & 56.99(\textcolor{morandiGreen}{$\downarrow$9.43})
 & 0.760(\textcolor{morandiGreen}{$\downarrow$0.018}) & 0.870(\textcolor{morandiGreen}{$\downarrow$0.009}) & 41.25(\textcolor{morandiGreen}{$\downarrow$0.36}) \\

 & \cellcolor[HTML]{E8D8FF}Least LIME
 & 0.556(\textcolor{morandiRed}{$\uparrow$0.048}) & 0.868(\textcolor{morandiGreen}{$\downarrow$0.004}) & 57.02(\textcolor{morandiGreen}{$\downarrow$9.40})
 & 0.754(\textcolor{morandiGreen}{$\downarrow$0.024}) & 0.870(\textcolor{morandiGreen}{$\downarrow$0.009}) & 41.27(\textcolor{morandiGreen}{$\downarrow$0.34}) \\

\cmidrule(lr){2-8}

 & \cellcolor[HTML]{D9EAFD}Conf
 & 0.660(\textcolor{morandiRed}{$\uparrow$0.152}) & 0.854(\textcolor{morandiGreen}{$\downarrow$0.018}) & 53.72(\textcolor{morandiGreen}{$\downarrow$12.70})
 & 0.826(\textcolor{morandiRed}{$\uparrow$0.048}) & 0.870(\textcolor{morandiGreen}{$\downarrow$0.009}) & 40.65(\textcolor{morandiGreen}{$\downarrow$0.96}) \\

 & \cellcolor[HTML]{E1F5E1}SHAP
 & 0.587(\textcolor{morandiRed}{$\uparrow$0.079}) & 0.847(\textcolor{morandiGreen}{$\downarrow$0.025}) & 56.81(\textcolor{morandiGreen}{$\downarrow$9.61})
 & 0.760(\textcolor{morandiGreen}{$\downarrow$0.018}) & 0.867(\textcolor{morandiGreen}{$\downarrow$0.012}) & 42.39(\textcolor{morandiRed}{$\uparrow$0.78}) \\

 & \cellcolor[HTML]{E1F5E1}AttnLRP
 & 0.596(\textcolor{morandiRed}{$\uparrow$0.088}) & 0.851(\textcolor{morandiGreen}{$\downarrow$0.021}) & 55.30(\textcolor{morandiGreen}{$\downarrow$11.12})
 & 0.810(\textcolor{morandiRed}{$\uparrow$0.032}) & 0.869(\textcolor{morandiGreen}{$\downarrow$0.010}) & 40.31(\textcolor{morandiGreen}{$\downarrow$1.30}) \\

 & \cellcolor[HTML]{E1F5E1}Grad$\times$Input
 & 0.624(\textcolor{morandiRed}{$\uparrow$0.116}) & 0.854(\textcolor{morandiGreen}{$\downarrow$0.018}) & 56.02(\textcolor{morandiGreen}{$\downarrow$10.40})
 & 0.782(\textcolor{morandiRed}{$\uparrow$0.004}) & 0.871(\textcolor{morandiGreen}{$\downarrow$0.008}) & 41.81(\textcolor{morandiRed}{$\uparrow$0.20}) \\

 & \cellcolor[HTML]{E1F5E1}LIME
 & 0.558(\textcolor{morandiRed}{$\uparrow$0.050}) & 0.866(\textcolor{morandiGreen}{$\downarrow$0.006}) & 58.18(\textcolor{morandiGreen}{$\downarrow$8.24})
 & 0.736(\textcolor{morandiGreen}{$\downarrow$0.042}) & 0.875(\textcolor{morandiGreen}{$\downarrow$0.004}) & 41.02(\textcolor{morandiGreen}{$\downarrow$0.59}) \\

 & \cellcolor[HTML]{FFEACC}NL
 & 0.638(\textcolor{morandiRed}{$\uparrow$0.130}) & 0.879(\textcolor{morandiRed}{$\uparrow$0.007}) & 148.41(\textcolor{morandiRed}{$\uparrow$81.99})
 & 0.770(\textcolor{morandiGreen}{$\downarrow$0.008}) & 0.901(\textcolor{morandiRed}{$\uparrow$0.022}) & 43.81(\textcolor{morandiRed}{$\uparrow$2.20}) \\

\bottomrule
\end{tabular}
\end{adjustbox}

\caption{
Automatic evaluation results of \lm{OLMo2-7B} on \data{SNLI}.
The tables present an ablation study of different feedback settings, including
\mbox{{\setlength{\fboxsep}{1pt}\colorbox[HTML]{C3F2F2}{\textit{Without Feedback}}}},
\mbox{{\setlength{\fboxsep}{1pt}\colorbox[HTML]{FFF8B8}{\textit{Random Feedback}}}},
\mbox{{\setlength{\fboxsep}{1pt}\colorbox[HTML]{E8D8FF}{\textit{Least-Attributed Feedback}}}}, and the full variants of \textit{i}\texttt{Flip}.
Colored values indicate relative changes compared to the \mbox{{\setlength{\fboxsep}{1pt}\colorbox[HTML]{C3F2F2}{\textit{Without Feedback} }}} baseline
(\textcolor{morandiRed}{$\uparrow$} = increase, \textcolor{morandiGreen}{$\downarrow$} = decrease).
}
\label{tab:ablation_feedback2}
\end{table*}

\subsection{Without Early Stopping}
\begin{figure*}[t]
    \centering
    \includegraphics[width=1\linewidth]{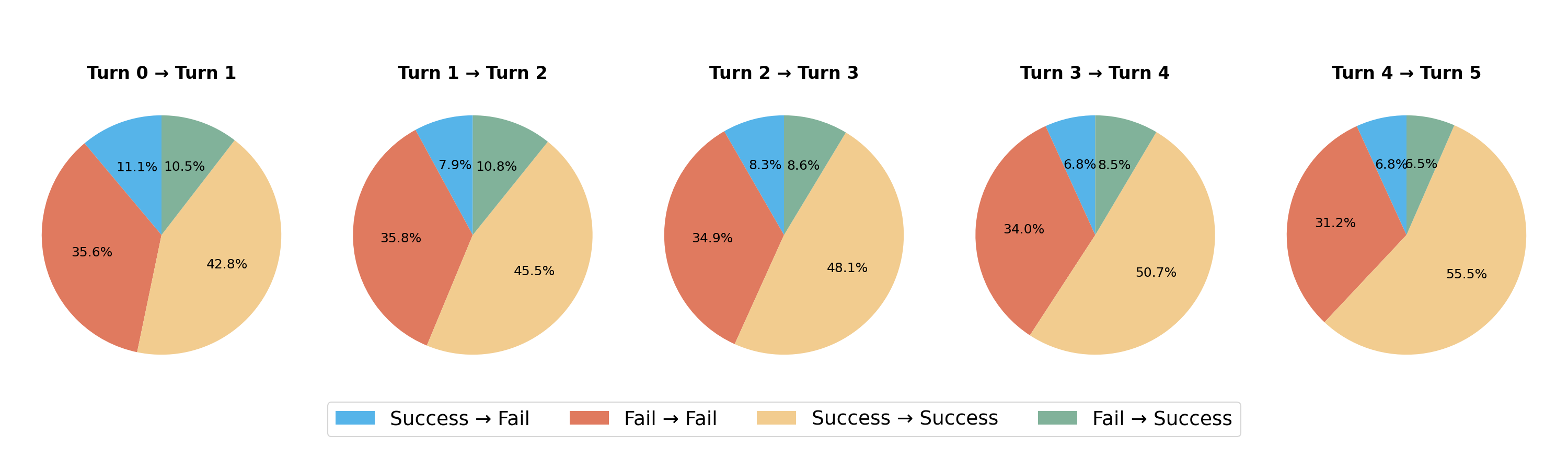}
    \caption{Iterative refinement with \lm{OLMo2-7B} \textbf{without early stopping}, averaged across all datasets and all feedback signals. A substantial share of \textit{Success $\rightarrow$ Fail} transitions indicates reduced stability compared to early-stopping.}
    \label{fig:without_early_stop}
\end{figure*}

Table \ref{tab:noearlystop1} reports the evaluation results of our without early stopping ablations on \data{IMDb} and \data{AG News}, while Table \ref{tab:noearlystop2} presents the corresponding results on \data{SNLI}. Figure \ref{fig:without_early_stop} presents the distribution of state transitions during iterative refinement when the early-stopping mechanism is disabled.

\begin{table*}[t]
\centering
\renewcommand\theadfont{\normalsize\bfseries}
\begin{adjustbox}{max width=\textwidth}
\begin{tabular}{ll|ccc|ccc}
\toprule
\textbf{} & \textbf{\makecell{Feedback of \textit{i}\texttt{Flip}}} & \multicolumn{3}{c|}{\textbf{\data{IMDb}}} & \multicolumn{3}{c}{\textbf{\data{AG News}}} \\
\cmidrule(lr){3-5} \cmidrule(lr){6-8}
& & LFR$\uparrow$ & SS$\uparrow$ & PPL$\downarrow$ & LFR$\uparrow$ & SS$\uparrow$ & PPL$\downarrow$ \\
\midrule
\multirow{6}{*}{\rotatebox{90}{\lm{OLMo2-7B}}}

& \cellcolor[HTML]{D9EAFD}Conf
& 0.677 (\textcolor{morandiGreen}{$\downarrow$0.287})
& 0.774 (\textcolor{morandiGreen}{$\downarrow$0.035})
& 41.32 (\textcolor{morandiRed}{$\uparrow$0.13})
& 0.507 (\textcolor{morandiGreen}{$\downarrow$0.277})
& 0.539 (\textcolor{morandiRed}{$\uparrow$0.036})
& 40.15 (\textcolor{morandiGreen}{$\downarrow$1.26}) \\

& \cellcolor[HTML]{E1F5E1}SHAP
& 0.859 (\textcolor{morandiGreen}{$\downarrow$0.109})
& 0.762 (\textcolor{morandiGreen}{$\downarrow$0.043})
& 41.34 (\textcolor{morandiGreen}{$\downarrow$0.53})
& 0.600 (\textcolor{morandiGreen}{$\downarrow$0.202})
& 0.485 (\textcolor{morandiGreen}{$\downarrow$0.003})
& 40.03 (\textcolor{morandiGreen}{$\downarrow$2.92}) \\

& \cellcolor[HTML]{E1F5E1}AttnLRP
& 0.922 (\textcolor{morandiGreen}{$\downarrow$0.056})
& 0.782 (\textcolor{morandiGreen}{$\downarrow$0.021})
& 40.75 (\textcolor{morandiGreen}{$\downarrow$1.86})
& 0.610 (\textcolor{morandiGreen}{$\downarrow$0.158})
& 0.502 (\textcolor{morandiRed}{$\uparrow$0.002})
& 38.75 (\textcolor{morandiGreen}{$\downarrow$7.86}) \\

& \cellcolor[HTML]{E1F5E1}Grad$\times$Input
& 0.916 (\textcolor{morandiGreen}{$\downarrow$0.060})
& 0.778 (\textcolor{morandiGreen}{$\downarrow$0.023})
& 40.11 (\textcolor{morandiGreen}{$\downarrow$3.19})
& 0.559 (\textcolor{morandiGreen}{$\downarrow$0.214})
& 0.499 (\textcolor{morandiRed}{$\uparrow$0.001})
& 38.61 (\textcolor{morandiGreen}{$\downarrow$11.71}) \\

& \cellcolor[HTML]{E1F5E1}LIME
& 0.890 (\textcolor{morandiGreen}{$\downarrow$0.078})
& 0.784 (\textcolor{morandiGreen}{$\downarrow$0.021})
& 38.47 (\textcolor{morandiGreen}{$\downarrow$14.35})
& 0.600 (\textcolor{morandiGreen}{$\downarrow$0.130})
& 0.503 (\textcolor{morandiGreen}{$\downarrow$0.001})
& 37.56 (\textcolor{morandiGreen}{$\downarrow$6.76}) \\

& \cellcolor[HTML]{FFEACC}NL
& 0.980 (\textcolor{morandiGreen}{$\downarrow$0.004})
& 0.747 (\textcolor{morandiGreen}{$\downarrow$0.075})
& 39.87 (\textcolor{morandiRed}{$\uparrow$3.40})
& 0.548 (\textcolor{morandiGreen}{$\downarrow$0.187})
& 0.499 (\textcolor{morandiGreen}{$\downarrow$0.016})
& 38.52 (\textcolor{morandiGreen}{$\downarrow$6.21}) \\

\bottomrule
\end{tabular}
\end{adjustbox}
\caption{Evaluation of \lm{OLMo2-7B} \emph{Without Early Stopping} on \data{IMDb} and \data{AG News}. Colored values indicate relative changes compared to \emph{Early Stopping} (\textcolor{morandiRed}{$\uparrow$} = increase, \textcolor{morandiGreen}{$\downarrow$} = decrease). }
\label{tab:noearlystop1}
\end{table*}

\begin{table*}[t]
\centering
\renewcommand\theadfont{\normalsize\bfseries}
\begin{adjustbox}{max width=\textwidth}
\begin{tabular}{ll|ccc|ccc}
\toprule
\textbf{} & \textbf{\makecell{Feedback of \textit{i}\texttt{Flip}}} & \multicolumn{3}{c|}{\textbf{\data{SNLI} (Premise)}} & \multicolumn{3}{c}{\textbf{\data{SNLI} (Hypothesis)}}\\
\cmidrule(lr){3-5} \cmidrule(lr){6-8}
& & LFR$\uparrow$ & SS$\uparrow$ & PPL$\downarrow$ & LFR$\uparrow$ & SS$\uparrow$ & PPL$\downarrow$ \\
\midrule
\multirow{6}{*}{\rotatebox{90}{\lm{OLMo2-7B}}}

& \cellcolor[HTML]{D9EAFD}Conf
& 0.340 (\textcolor{morandiGreen}{$\downarrow$0.320})
& 0.860 (\textcolor{morandiRed}{$\uparrow$0.006})
& 52.47 (\textcolor{morandiGreen}{$\downarrow$1.25})
& 0.560 (\textcolor{morandiGreen}{$\downarrow$0.266})
& 0.873 (\textcolor{morandiRed}{$\uparrow$0.003})
& 39.46 (\textcolor{morandiGreen}{$\downarrow$1.19}) \\

& \cellcolor[HTML]{E1F5E1}SHAP
& 0.340 (\textcolor{morandiGreen}{$\downarrow$0.247})
& 0.848 (\textcolor{morandiRed}{$\uparrow$0.001})
& 57.20 (\textcolor{morandiRed}{$\uparrow$0.39})
& 0.543 (\textcolor{morandiGreen}{$\downarrow$0.217})
& 0.871 (\textcolor{morandiRed}{$\uparrow$0.004})
& 41.47 (\textcolor{morandiGreen}{$\downarrow$0.92}) \\

& \cellcolor[HTML]{E1F5E1}AttnLRP
& 0.343 (\textcolor{morandiGreen}{$\downarrow$0.253})
& 0.858 (\textcolor{morandiRed}{$\uparrow$0.007})
& 58.07 (\textcolor{morandiRed}{$\uparrow$2.77})
& 0.563 (\textcolor{morandiGreen}{$\downarrow$0.247})
& 0.871 (\textcolor{morandiRed}{$\uparrow$0.002})
& 40.60 (\textcolor{morandiRed}{$\uparrow$0.29}) \\

& \cellcolor[HTML]{E1F5E1}Grad$\times$Input
& 0.347 (\textcolor{morandiGreen}{$\downarrow$0.277})
& 0.854 (---)
& 55.76 (\textcolor{morandiGreen}{$\downarrow$0.26})
& 0.520 (\textcolor{morandiGreen}{$\downarrow$0.262})
& 0.871 (---)
& 40.76 (\textcolor{morandiGreen}{$\downarrow$1.05}) \\

& \cellcolor[HTML]{E1F5E1}LIME
& 0.323 (\textcolor{morandiGreen}{$\downarrow$0.235})
& 0.875 (\textcolor{morandiRed}{$\uparrow$0.009})
& 57.04 (\textcolor{morandiGreen}{$\downarrow$1.14})
& 0.497 (\textcolor{morandiGreen}{$\downarrow$0.239})
& 0.872 (\textcolor{morandiGreen}{$\downarrow$0.003})
& 40.38 (\textcolor{morandiGreen}{$\downarrow$0.64}) \\

& \cellcolor[HTML]{FFEACC}NL
& 0.387 (\textcolor{morandiGreen}{$\downarrow$0.251})
& 0.844 (\textcolor{morandiGreen}{$\downarrow$0.035})
& 54.87 (\textcolor{morandiGreen}{$\downarrow$93.54})
& 0.546 (\textcolor{morandiGreen}{$\downarrow$0.224})
& 0.903 (\textcolor{morandiRed}{$\uparrow$0.002})
& 43.81 (---) \\

\bottomrule
\end{tabular}
\end{adjustbox}
\caption{Evaluation of \lm{OLMo2-7B} \emph{Without Early Stopping} on \data{SNLI} (Premise) and \data{SNLI} (Hypothesis). Colored values indicate relative changes compared to \emph{Early Stopping} (\textcolor{morandiRed}{$\uparrow$} = increase, \textcolor{morandiGreen}{$\downarrow$} = decrease). }
\label{tab:noearlystop2}
\end{table*}

\subsection{Prompt-Level Ablations on Natural Language Feedback}
\label{subsec:ablation_nl_feedback}

To better understand why natural language (NL) feedback is effective, we conduct prompt-level ablations on the NL-feedback generation step using \lm{Qwen3-32B}. Starting from the full NL-feedback prompt, we construct four variants that request: (i) less specific lexical suggestions, (ii) less explanation about class semantics, (iii) less aggressive edits, and (iv) shorter outputs. Tables~\ref{tab:nl_feedback_ablation_all} report the results.

The results show that \textbf{the effectiveness of NL feedback stems from both its informativeness and its ability to provide actionable refinement guidance.} On \data{IMDb} and \data{AG News}, reducing edit aggressiveness leads to the most noticeable validity drops, suggesting that stronger edits are often necessary for longer texts. On \data{SNLI} (Premise) and \data{SNLI} (Hypothesis), shorter feedback is more harmful, indicating that richer guidance is especially important for NLI tasks. Together with the combined-feedback results in Table~\ref{tab:main_results}, these findings suggest that NL feedback is effective because it provides detailed and actionable refinement instructions.

\begin{table*}[t]
    \centering
    \footnotesize
    \renewcommand{\arraystretch}{1.0}
    \resizebox{\textwidth}{!}{
    \begin{tabular}{l|ccc|ccc|ccc|ccc}
        \toprule
        & \multicolumn{3}{c|}{\textbf{\data{IMDb}}}
        & \multicolumn{3}{c|}{\textbf{\data{AG News}}}
        & \multicolumn{3}{c|}{\textbf{\data{SNLI} (Premise)}}
        & \multicolumn{3}{c}{\textbf{\data{SNLI} (Hypothesis)}} \\
        \cmidrule(lr){2-4} \cmidrule(lr){5-7} \cmidrule(lr){8-10} \cmidrule(lr){11-13}
        & \textbf{LFR}$\uparrow$ & \textbf{SS}$\uparrow$ & \textbf{PPL}$\downarrow$
        & \textbf{LFR}$\uparrow$ & \textbf{SS}$\uparrow$ & \textbf{PPL}$\downarrow$
        & \textbf{LFR}$\uparrow$ & \textbf{SS}$\uparrow$ & \textbf{PPL}$\downarrow$
        & \textbf{LFR}$\uparrow$ & \textbf{SS}$\uparrow$ & \textbf{PPL}$\downarrow$ \\
        \midrule
        Full NL feedback
        & 0.980 & 0.878 & 32.99
        & 0.732 & 0.536 & 49.68
        & 0.596 & 0.895 & 68.16
        & 0.754 & 0.906 & 47.20 \\
        - Less specific lexical suggestions
        & 0.980 & 0.870 & 29.85
        & 0.799 & 0.507 & 44.09
        & 0.588 & 0.895 & 70.38
        & 0.769 & 0.897 & 50.38 \\
        - Less class-semantic explanation
        & 0.985 & 0.869 & 30.92
        & 0.774 & 0.507 & 47.58
        & 0.573 & 0.903 & 69.31
        & 0.688 & 0.905 & 50.85 \\
        - Less edit aggressiveness
        & 0.945 & 0.882 & 31.12
        & 0.719 & 0.552 & 50.76
        & 0.518 & 0.905 & 70.96
        & 0.774 & 0.912 & 51.99 \\
        - Shorter output
        & 0.980 & 0.873 & 32.14
        & 0.749 & 0.543 & 44.54
        & 0.422 & 0.922 & 65.06
        & 0.422 & 0.918 & 50.14 \\
        \bottomrule
    \end{tabular}
    }
    \caption{
    Prompt-level ablations on natural language feedback using \lm{Qwen3-32B}.
    }
    \label{tab:nl_feedback_ablation_all}
\end{table*}

\section{Counterfactual Data Augmentation}
\label{app:augmentation_detail}

Tables~\ref{tab:augmentation_olmo} and~\ref{tab:augmentation_llama} report the results of data augmentation on in-domain (ID), out-of-domain (OOD), and human-annotated counterfactual (Human CFs) test data across four tasks: Sentiment Analysis, News Classification, NLI Premise, and NLI Hypothesis. The CFEs for Table~\ref{tab:augmentation_olmo} are produced with \textit{i}\texttt{Flip} using the \lm{OLMo2-7B} model, while those for Table~\ref{tab:augmentation_llama} are generated using the \lm{LLaMA3-70B} model.

\begin{table*}[!t]
\centering
\small
\setlength{\tabcolsep}{3pt}  
\renewcommand{\arraystretch}{0.9} 
\resizebox{\textwidth}{!}{
\begin{tabular}{l|c|ccc||c|ccc||c|ccc||c|ccc}
\toprule
& \multicolumn{4}{c||}{\textbf{Sentiment Analysis}} 
& \multicolumn{4}{c||}{\textbf{News Classification}} 
& \multicolumn{4}{c||}{\textbf{NLI - Premise}} 
& \multicolumn{4}{c}{\textbf{NLI - Hypothesis}} \\
\cmidrule(lr){2-5}\cmidrule(lr){6-9}\cmidrule(lr){10-13}\cmidrule(lr){14-17}
\multirow{2}{*}{Methods}
& \multicolumn{4}{c||}{\textit{Test Data}}
& \multicolumn{4}{c||}{\textit{Test Data}}
& \multicolumn{4}{c||}{\textit{Test Data}}
& \multicolumn{4}{c}{\textit{Test Data}} \\
\cmidrule(lr){2-5}\cmidrule(lr){6-9}\cmidrule(lr){10-13}\cmidrule(lr){14-17}
& \textbf{ID} & \multicolumn{3}{c||}{\textbf{OOD}}
& \textbf{ID} & \multicolumn{3}{c||}{\textbf{OOD}}
& \textbf{ID} & \multicolumn{3}{c||}{\textbf{OOD}}
& \textbf{ID} & \multicolumn{3}{c}{\textbf{OOD}} \\
& IMDb & Amazon & SST-2 & CFs
& \data{AG News} & BBC & 20NG & CFs
& \data{SNLI} & MNLI & ANLI & CFs
& \data{SNLI} & MNLI & ANLI & CFs \\
\midrule
\cellcolor[HTML]{d8d8d8}Baseline
& \cellcolor[HTML]{d8d8d8}77.93 & \cellcolor[HTML]{d8d8d8}75.00 & \cellcolor[HTML]{d8d8d8}67.43 & \cellcolor[HTML]{d8d8d8}50.09
& \cellcolor[HTML]{d8d8d8}66.87 & \cellcolor[HTML]{d8d8d8}62.80 & \cellcolor[HTML]{d8d8d8}41.07 & \cellcolor[HTML]{d8d8d8}--
& \cellcolor[HTML]{d8d8d8}40.87 & \cellcolor[HTML]{d8d8d8}34.87 & \cellcolor[HTML]{d8d8d8}\textbf{35.00} & \cellcolor[HTML]{d8d8d8}33.43
& \cellcolor[HTML]{d8d8d8}40.87 & \cellcolor[HTML]{d8d8d8}34.87 & \cellcolor[HTML]{d8d8d8}\textbf{35.00} & \cellcolor[HTML]{d8d8d8}33.85 \\
\cellcolor[HTML]{d8d8d8}Human
& \cellcolor[HTML]{d8d8d8}\textbf{92.73} & \cellcolor[HTML]{d8d8d8}\textbf{89.53} & \cellcolor[HTML]{d8d8d8}\textbf{85.55} & \cellcolor[HTML]{d8d8d8}\textbf{91.26}
& \cellcolor[HTML]{d8d8d8}-- & \cellcolor[HTML]{d8d8d8}-- & \cellcolor[HTML]{d8d8d8}-- & \cellcolor[HTML]{d8d8d8}--
& \cellcolor[HTML]{d8d8d8}\textbf{42.27} & \cellcolor[HTML]{d8d8d8}\textbf{37.47} & \cellcolor[HTML]{d8d8d8}33.93 & \cellcolor[HTML]{d8d8d8}\textbf{50.90}
& \cellcolor[HTML]{d8d8d8}\textbf{58.40} & \cellcolor[HTML]{d8d8d8}\textbf{52.07} & \cellcolor[HTML]{d8d8d8}30.13 & \cellcolor[HTML]{d8d8d8}\textbf{56.95} \\
\midrule
\cellcolor[HTML]{D9EAFD}Conf
& 87.93 & 88.33 & 83.83 & 71.09
& 78.60 & 45.30 & 55.80 & --
& 38.47 & \textbf{36.33} & 32.80 & \textbf{32.98}
& \textbf{44.13} & 33.13 & 32.87 & \textbf{36.88} \\
\cellcolor[HTML]{E1F5E1}SHAP
& 91.33 & \textbf{89.33} & 83.94 & 90.30
& 80.27 & \textbf{48.60} & 56.33 & --
& 37.40 & 34.93 & \textbf{32.93} & 27.82
& 40.33 & \textbf{36.87} & 33.13 & 35.83 \\
\cellcolor[HTML]{E1F5E1}AttnLRP
& 89.27 & 88.00 & \textbf{84.75} & 90.28
& 77.87 & 42.50 & 53.60 & --
& 38.13 & 35.67 & 32.87 & 27.97
& 40.13 & 36.33 & 33.13 & 35.53 \\
\cellcolor[HTML]{E1F5E1}Grad$\times$Input
& \textbf{93.07} & 88.00 & 82.45 & 90.07
& 79.07 & 43.70 & 55.80 & --
& 39.00 & 31.93 & \textbf{32.93} & 26.98
& 39.40 & 34.13 & \textbf{33.20} & 35.98 \\
\cellcolor[HTML]{E1F5E1}LIME
& 91.40 & 88.73 & 83.37 & \textbf{90.91}
& 78.80 & 42.50 & 54.53 & --
& 38.33 & 32.60 & \textbf{32.93} & 28.93
& 38.93 & 35.80 & 32.93 & 35.74 \\
\cellcolor[HTML]{FFEACC}NL
& 90.73 & 88.27 & 84.40 & 71.97
&\textbf{ 81.80} & 46.10 & \textbf{58.47} & --
& \textbf{40.40} & 33.60 & 32.73 & 30.91
& 37.47 & 33.27 & \textbf{33.20} & 33.79 \\

\bottomrule
\end{tabular}
}
\caption{Accuracy (\%) on in-domain (ID), out-of-domain (OOD), and human-annotated counterfactual (Human CFs) test data across four tasks: Sentiment Analysis, News Classification, NLI Premise, and NLI Hypothesis. Best results are bolded. The CFEs used for data augmentation, produced with \textit{i}\texttt{Flip}, are generated by an OLMo2-7B model.}
\label{tab:augmentation_olmo}
\end{table*}

\begin{table*}[!t]
\centering
\small
\setlength{\tabcolsep}{3pt}  
\renewcommand{\arraystretch}{0.9} 
\resizebox{\textwidth}{!}{
\begin{tabular}{l|c|ccc||c|ccc||c|ccc||c|ccc}
\toprule
& \multicolumn{4}{c||}{\textbf{Sentiment Analysis}} 
& \multicolumn{4}{c||}{\textbf{News Classification}} 
& \multicolumn{4}{c||}{\textbf{NLI - Premise}} 
& \multicolumn{4}{c}{\textbf{NLI - Hypothesis}} \\
\cmidrule(lr){2-5}\cmidrule(lr){6-9}\cmidrule(lr){10-13}\cmidrule(lr){14-17}
\multirow{2}{*}{Methods}
& \multicolumn{4}{c||}{\textit{Test Data}}
& \multicolumn{4}{c||}{\textit{Test Data}}
& \multicolumn{4}{c||}{\textit{Test Data}}
& \multicolumn{4}{c}{\textit{Test Data}} \\
\cmidrule(lr){2-5}\cmidrule(lr){6-9}\cmidrule(lr){10-13}\cmidrule(lr){14-17}
& \textbf{ID} & \multicolumn{3}{c||}{\textbf{OOD}}
& \textbf{ID} & \multicolumn{3}{c||}{\textbf{OOD}}
& \textbf{ID} & \multicolumn{3}{c||}{\textbf{OOD}}
& \textbf{ID} & \multicolumn{3}{c}{\textbf{OOD}} \\
& IMDb & Amazon & SST-2 & CFs
& \data{AG News} & BBC & 20NG & CFs
& \data{SNLI} & MNLI & ANLI & CFs
& \data{SNLI} & MNLI & ANLI & CFs \\
\midrule
\cellcolor[HTML]{d8d8d8}Baseline
& \cellcolor[HTML]{d8d8d8}77.93 & \cellcolor[HTML]{d8d8d8}75.00 & \cellcolor[HTML]{d8d8d8}67.43 & \cellcolor[HTML]{d8d8d8}50.09
& \cellcolor[HTML]{d8d8d8}66.87 & \cellcolor[HTML]{d8d8d8}62.80 & \cellcolor[HTML]{d8d8d8}41.07 & \cellcolor[HTML]{d8d8d8}--
& \cellcolor[HTML]{d8d8d8}40.87 & \cellcolor[HTML]{d8d8d8}34.87 & \cellcolor[HTML]{d8d8d8}\textbf{35.00} & \cellcolor[HTML]{d8d8d8}33.43
& \cellcolor[HTML]{d8d8d8}40.87 & \cellcolor[HTML]{d8d8d8}34.87 & \cellcolor[HTML]{d8d8d8}\textbf{35.00} & \cellcolor[HTML]{d8d8d8}33.85 \\
\cellcolor[HTML]{d8d8d8}Human
& \cellcolor[HTML]{d8d8d8}\textbf{92.73} & \cellcolor[HTML]{d8d8d8}\textbf{89.53} & \cellcolor[HTML]{d8d8d8}\textbf{85.55} & \cellcolor[HTML]{d8d8d8}\textbf{91.26}
& \cellcolor[HTML]{d8d8d8}-- & \cellcolor[HTML]{d8d8d8}-- & \cellcolor[HTML]{d8d8d8}-- & \cellcolor[HTML]{d8d8d8}--
& \cellcolor[HTML]{d8d8d8}\textbf{42.27} & \cellcolor[HTML]{d8d8d8}\textbf{37.47} & \cellcolor[HTML]{d8d8d8}33.93 & \cellcolor[HTML]{d8d8d8}\textbf{50.90}
& \cellcolor[HTML]{d8d8d8}\textbf{58.40} & \cellcolor[HTML]{d8d8d8}\textbf{52.07} & \cellcolor[HTML]{d8d8d8}30.13 & \cellcolor[HTML]{d8d8d8}\textbf{56.95} \\
\midrule
\cellcolor[HTML]{D9EAFD}Conf
& 91.67 & 88.60 & 84.29 & 90.92
& \textbf{83.07} & 66.80 & 58.33 & --
& 34.53 & \textbf{33.00} & 32.80 & 33.49
& \textbf{48.40} & \textbf{37.40} & 32.00 & \textbf{41.09} \\
\cellcolor[HTML]{E1F5E1}SHAP
& 91.53 & 88.27 & 82.80 & 90.71
& 79.73 & 59.30 & 62.40 & --
& 36.20 & 32.47 & 32.73 & 33.16
& 40.60 & 34.20 & 31.40 & 38.03 \\
\cellcolor[HTML]{E1F5E1}AttnLRP
& 88.80 & 88.80 & \textbf{84,40} & \textbf{91.21}
& 80.33 & 58.50 & 61.33 & --
& 33.27 & 31.87 & 32.80 & 32.32
& 39.80 & 33.40 & 32.73 & 36.58 \\
\cellcolor[HTML]{E1F5E1}Grad$\times$Input
& \textbf{92.27} & \textbf{88.87} & 83.72 & 90.89
& 81.60 & 59.00 & 62.73 & --
& 34.47 & 32.27 & 32.73 & 31.84
& 45.47 & 36.00 & 32.07 & 38.71 \\
\cellcolor[HTML]{E1F5E1}LIME
& 91.00 & 88.80 & 84.06 & 90.83
& 79.80 & 57.30 & 62.47 & --
& 36.13 & 32.87 & 32.60 & 33.28
& 41.47 & 35.47 & \textbf{33.40} & 35.86 \\
\cellcolor[HTML]{FFEACC}NL
& 91.67 & 88.73 & 83.83 & 90.89
& 81.73 & \textbf{69.50} & \textbf{66.27} & --
& \textbf{37.13} & 31.00 & \textbf{32.87} & \textbf{33.94}
& 39.73 & 36.33 & 32.27 & 36.94 \\

\bottomrule
\end{tabular}
}
\caption{Accuracy (\%) on in-domain (ID), out-of-domain (OOD), and human-annotated counterfactual (Human CFs) test data across four tasks: Sentiment Analysis, News Classification, NLI Premise, and NLI Hypothesis. Best results are bolded. The CFEs used for data augmentation, produced with \textit{i}\texttt{Flip}, are generated by an LLaMA3.3-70B model.}
\label{tab:augmentation_llama}
\end{table*}



\section{Faithfulness Evaluation}
\label{sec:faithfulness}
Given that feature attribution-based feedback serve as a central component of our framework, it is essential to assess the faithfulness of different attribution methods. 
Faithfulness measures how accurately explanations reflect the true reasoning of the model \citep{lyu-etal-2024-faith}. 
We evaluate attribution methods using FERRET \citep{Attanasio_2023} with three metrics: comprehensiveness, 
sufficiency \citep{deyoung-etal-2020-eraser}, and Kendall’s $\tau$ correlation with Leave-One-Out \citep{jain-wallace-2019-attention}. 
These metrics jointly assess whether highlighted words faithfully capture the model’s decision process.

\begin{table*}[!t]
\centering
\resizebox{\textwidth}{!}{
\begin{tabular}{llccc|ccc|ccc|ccc}
\toprule
\multirow{2}{*}{Model} & \multirow{2}{*}{Feedback} 
& \multicolumn{3}{c|}{\data{IMDb}} 
& \multicolumn{3}{c|}{\data{AG News}} 
& \multicolumn{3}{c|}{\data{SNLI}-Premise} 
& \multicolumn{3}{c}{\data{SNLI}-Hypothesis} \\
\cmidrule(lr){3-5}\cmidrule(lr){6-8}\cmidrule(lr){9-11}\cmidrule(lr){12-14}
& & comp. & suff. & $\tau$(loo) 
  & comp. & suff. & $\tau$(loo) 
  & comp. & suff. & $\tau$(loo) 
  & comp. & suff. & $\tau$(loo) \\
\midrule
\multirow{4}{*}{\lm{OLMo2-7B}} 
& SHAP     & 0.28 & \textbf{-0.01} & 0.02 & \textbf{0.56} & \textbf{0.04} & 0.03 & 0.09 & -0.19 & 0.21 & 0.07 & -0.22 & 0.20 \\
& AttnLRP  & \textbf{0.76} & \textbf{-0.00} & \textbf{0.09} & 0.53 & 0.20 & \textbf{0.18} & \textbf{0.12} & -0.23 & 0.18 & \textbf{0.12} & -0.25 & 0.17 \\
& Grad$\times$Input & 0.05 & 0.07 & -0.00 & 0.31 & 0.27 & 0.01 & 0.00 & \textbf{-0.07} & 0.05 & -0.02 & \textbf{-0.11} & 0.03 \\
& LIME     & 0.12 & 0.03 & 0.01 & 0.39 & 0.21 & 0.04 & 0.10 & -0.16 & \textbf{0.23} & 0.08 & -0.20 & \textbf{0.23} \\
\midrule
\multirow{4}{*}{Qwen-32B} 
& SHAP     & 0.34 & \textbf{-0.01} & 0.04 & \textbf{0.66} & \textbf{-0.03} & 0.08 & 0.08 & -0.22 & 0.23 & 0.06 & -0.26 & 0.22 \\
& AttnLRP  & \textbf{0.78} & \textbf{-0.01} & \textbf{0.11} & 0.38 & 0.19 & \textbf{0.22} & \textbf{0.10} & -0.26 & 0.22 & \textbf{0.10} & -0.27 & 0.24 \\
& Grad$\times$Input & 0.06 & 0.08 & -0.00 & 0.24 & 0.20 & 0.05 & -0.01 & \textbf{-0.10} & 0.06 & -0.03 & \textbf{-0.11} & 0.07 \\
& LIME     & 0.15 & 0.05 & 0.02 & 0.43 & 0.06 & 0.08 & 0.08 & -0.19 & \textbf{0.26} & 0.07 & -0.24 & \textbf{0.26} \\
\midrule
\multirow{4}{*}{LlaMa-70B} 
& SHAP     & 0.34 & \textbf{-0.00} & 0.03 & \textbf{0.59} & \textbf{-0.00} & 0.07 & 0.06 & -0.21 & 0.19 & 0.06 & -0.25 & 0.19 \\
& AttnLRP  & \textbf{0.79} & \textbf{-0.00} & \textbf{0.11} & 0.32 & 0.19 & \textbf{0.22} & \textbf{0.07} & -0.24 & 0.20 & \textbf{0.08} & -0.28 & 0.21 \\
& Grad$\times$Input & 0.05 & 0.06 & -0.00 & 0.22 & 0.16 & 0.06 & -0.02 & \textbf{-0.09} & 0.05 & -0.04 & \textbf{-0.12} & 0.04 \\
& LIME     & 0.12 & 0.04 & 0.02 & 0.36 & 0.08 & 0.06 & \textbf{0.07} & -0.19 & \textbf{0.25} & 0.06 & -0.24 & \textbf{0.23} \\
\bottomrule
\end{tabular}}
\caption{Faithfulness evaluation results: \textbf{AOPC Comprehensiveness} (comp., $\uparrow$), \textbf{AOPC Sufficiency} (suff., 0 best), and \textbf{Kendall’s $\tau$ correlation with Leave-One-Out} ($\tau$(loo), $\uparrow$). Higher $\uparrow$ and lower $\downarrow$ are better. Best results per column are \textbf{bold}.}
\label{tab:faithfulness}
\end{table*}

The results in Table~\ref{tab:faithfulness} show clear differences in how attribution methods capture the model’s reasoning process. Among the evaluated approaches, AttnLRP provides the most faithful explanations, showing strong performance across comprehensiveness, sufficiency, and Kendall’s $\tau$. Propagation-based methods exhibit solid alignment with the model’s decision-making process, which aligns with our earlier observations in Section~\ref{sec:automatic_evaluation}. SHAP follows as the next most faithful method across datasets and models, with competitive scores on comprehensiveness and sufficiency, though less consistent on $\tau$. 

We further observe that the results on the \data{SNLI} dataset show weaker faithfullness compared to sentiment and news classification. In particular, attribution methods on \data{SNLI} exhibit lower comprehensiveness, together with less favorable sufficiency values, indicating that the most important words identified by attribution methods capture the model’s reasoning less effectively in NLI tasks. This suggests that attribution methods are more challenging when explanations must account for subtle entailment relations rather than more direct sentiment or news topic.


\section{Correlation Analysis}
\label{app:correlation}

We further analyze the relationship between counterfactual quality and two key aspects: 
their performance in Counterfactual Data Augmentation (CDA) and their alignment with attribution-based faithfulness. 
To assess counterfactual quality, we employ the \textit{label flipping rate} and derive feedback-specific rankings.
CDA performance is measured by the ranking of average OOD accuracy on test sets obtained from models trained with different counterfactual sets. Faithfulness is captured through attribution-based metrics -- \textit{comprehensiveness} \cite{deyoung-etal-2020-eraser}, \textit{sufficiency} \cite{deyoung-etal-2020-eraser}, and Kendall’s $\tau$ correlation with Leave-One-Out token removal \cite{jain-wallace-2019-attention}.

\subsection{Correlation between CDA performance and counterfactual quality}
\label{subsec:corr_aug_subsec}
\begin{figure}[t]
    \centering
    \begin{adjustbox}{max width=\columnwidth}
        \includegraphics{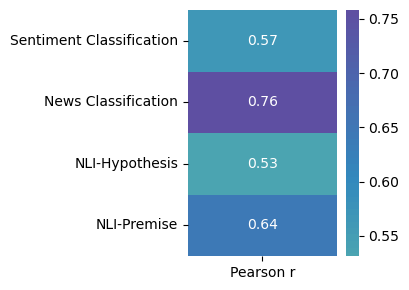}

    \end{adjustbox}
    \caption{Pearson correlation between augmentation performance and counterfactual quality across datasets.}
    \label{fig:corr_aug}
\end{figure}

Figure~\ref{fig:corr_aug} shows the Pearson correlation between CDA performance and counterfactual quality across datasets. Among the datasets examined, \data{AG News} exhibits a strong correlation between the counterfactuals quality and CDA effectiveness, whereas \data{IMDb} and \data{SNLI} demonstrate only moderate correlation. Nonetheless, feedback-based refinement generally yields effective counterfactuals, which in turn lead to enhanced model performance and robustness. \\

\Needspace{5\baselineskip}
\Needspace{5\baselineskip}

\subsection{Correlation between Faithfulness measures and counterfactual quality.}
\label{subsec:corr_faith_subsec}

\Needspace{5\baselineskip}

\begin{figure*}[h]
    \centering
    \begin{adjustbox}{max width=\textwidth}
        \includegraphics{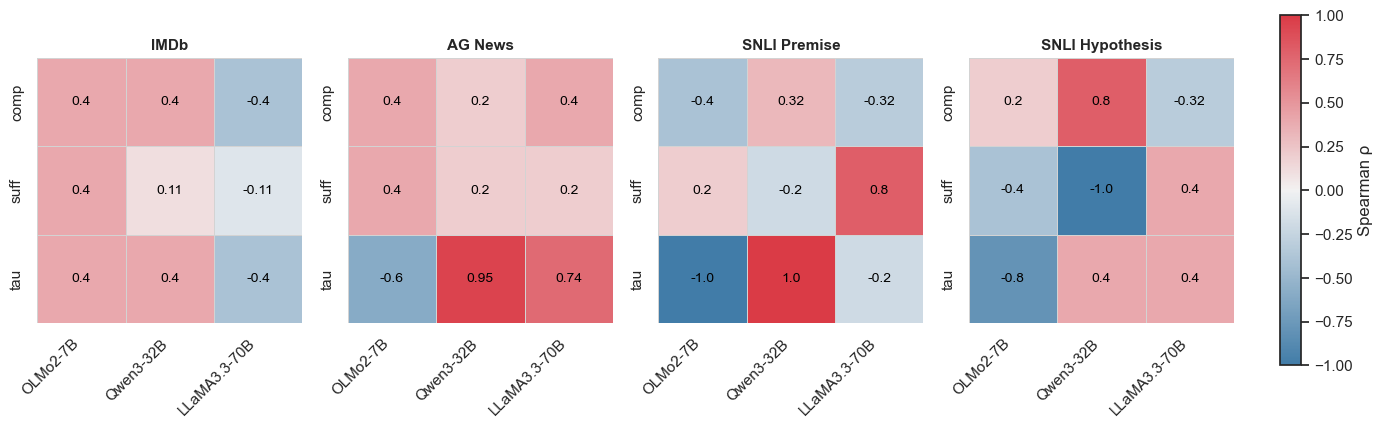}
    \end{adjustbox}
    \caption{Spearman correlation between faithfulness and counterfactual quality across datasets.}
    \label{fig:corr_faith}
\end{figure*}

Figure~\ref{fig:corr_faith} reports the Spearman correlation between faithfulness measures and counterfactual quality. 
The results reveal heterogeneous patterns across tasks: while datasets such as \data{IMDb} and \data{AG News} show positive correlations between flip-rate quality and faithfulness metrics, \data{SNLI} yields weaker or even negative correlations, especially for \data{SNLI} Hypothesis. These findings suggest that higher-quality counterfactuals tend to align better with attribution signals in sentiment and topic classification, whereas this alignment is less stable in NLI due to the complexity of entailment reasoning.

\newcommand{\attrfontsize}{\footnotesize}

\setlength{\tabcolsep}{5pt}
\renewcommand{\arraystretch}{1.18}

\section{Highlighted Words from Attribution Methods}
\label{app:highlight_words}

Table~\ref{tab:smear_highlight} illustrate how different attribution methods 
identify most important words within the same \data{IMDb} review.

\begin{table*}[htbp]
\centering
{\attrfontsize
\begin{tabularx}{\textwidth}{>{\raggedright\arraybackslash}p{0.13\textwidth} >{\raggedright\arraybackslash}X}
\toprule
\textbf{Method} & \textbf{\data{IMDb} Review} \\

\midrule
\textbf{AttenLRP} 
\newline  
&
\hl{This} \hl{movie} is on the level with Welcome Home Roxy Carmichael for \hl{biggest} \hl{pieces} \hl{of} \hl{garbage} that have \hl{ever} hit the silver screen. 
If these guys weren't Adam Sandler's gay friends, \hl{this} \hl{script} would have ended up where it should have: as some big time \hl{movie} exec's toilet paper. 
\hl{I} \hl{hate} \hl{this} \hl{movie}, it makes me want to injure people. 
I will admit that I have high standards, but honestly I'd rather \hl{watch} Step Up 2. 
The ultra sad part was when I logged onto \data{IMDb} and read that you \hl{pieces} of trash actually gave \hl{this} \hl{movie} a 6.9 rating. 
\hl{This} is a testament to all of the retards in our \hl{society} that will go \hl{watch} \hl{terrible} movies that are just hour and a half long dick, fart, and weed jokes with \hl{little} to no originality. 
After seeing this rating, I would like to suggest Tyler Perry's House of Pain to all of you guys who \hl{enjoyed} this film; you'll see some high quality \hl{humor} there on about the same level of \hl{this} abhorrent abomination. \\
\midrule
\textbf{SHAP}
\newline 
&
This movie is on the \hl{level} with Welcome Home Roxy Carmichael \hl{for} biggest pieces of garbage that have ever hit the silver screen. 
If \hl{these} guys \hl{weren't} Adam Sandler's gay friends, this script would have ended up where it should have: as some big time movie exec's toilet paper. 
I hate this movie, it makes me want to injure people. I will admit that I have high standards, but honestly I'd rather watch Step Up 2. 
The ultra sad part was when I logged onto \data{IMDb} and read that you pieces of trash actually gave this movie a 6.9 \hl{rating}. 
This is \hl{a} \hl{testament} to all of the retards in our \hl{society} that will go watch terrible movies that are just \hl{hour} and a half long \hl{dick}, fart, and weed jokes with little to no originality. 
After seeing this \hl{rating}, I would like to suggest Tyler Perry's House of \hl{Pain} to all of you guys who \hl{enjoyed} this film\hl{;} you'll see some high \hl{quality} \hl{humor} there on about the same \hl{level} of this \hl{abomination}. \\
\midrule
\textbf{LIME}
\newline 
&
This movie is on the level with \hl{Welcome} \hl{Home} Roxy Carmichael for biggest pieces of garbage that have ever hit the \hl{silver} screen. 
If these guys weren't Adam \hl{Sandler}'s \hl{gay} friends, this script would have ended up where it should have: as some big time movie exec's toilet paper. 
I hate this movie, \hl{it} makes me want to injure people. 
I will admit that I have high standards, but honestly I'\hl{d} rather watch Step Up 2. 
\hl{The} ultra sad part was when I logged onto \data{IMDb} and read that you pieces of trash \hl{actually} gave this movie a 6.9 rating. 
This is a testament to all of the \hl{retards} in our society that will go watch \hl{terrible} \hl{movies} that are just \hl{hour} and a half long dick, fart, and weed \hl{jokes} with \hl{little} to no originality. 
\hl{After} seeing this rating, I would like to suggest Tyler Perry's House of \hl{Pain} to all of you guys who enjoyed this film; you'll see some high quality humor there on about the same level of this \hl{abhorrent} abomination. \\
\midrule
\textbf{Grad$\times$Input}
\newline 
&
\hl{This} movie is on the level with Welcome Home Roxy Carmichael for \hl{biggest} \hl{pieces} \hl{of} \hl{garbage} that have \hl{ever} hit the silver screen. 
If these guys weren't Adam Sandler's gay friends, \hl{this} \hl{script} would have ended up where it should have: as some big time movie exec's toilet paper. 
\hl{I} \hl{hate} \hl{this} movie, it makes me want to injure people. 
I will admit that I have high standards, but honestly I'd rather \hl{watch} Step Up 2. 
The ultra sad part was when I logged onto \data{IMDb} and read that you \hl{pieces} of trash actually gave this movie a 6.9 rating. 
This is a testament to all of the retards in our \hl{society} that will go watch \hl{terrible} movies that are just hour and a half long dick, fart, and weed jokes with \hl{little} to no originality. 
After seeing this rating, I would like to suggest Tyler Perry's House of Pain to all of you guys who \hl{enjoyed} this film; you'll see some high quality \hl{humor} there on about the same level of this abhorrent abomination. \\
\bottomrule
\end{tabularx}
}
\caption{\data{IMDb} review with Top-k important words highlighted under different attribution methods.}

\label{tab:smear_highlight}
\end{table*}

\section{Additional Examples}
\label{app:additional_examples}

This section presents illustrative case studies of counterfactual generation, showcasing both successful refinements and common failure patterns of the proposed framework.

\subsection{Example 1: Undesired Flip} 

\noindent \textbf{Observation:} 
Table~\ref{tab:example1} shows that the framework can produce label flips, but not always toward the desired direction.
The label was flipped from \textbf{NEUTRAL} to \textbf{ENTAILMENT} instead of the expected \textbf{CONTRADICTION}.

\begin{table*}[h]
\centering
\renewcommand{\arraystretch}{1.25}
\footnotesize
\begin{tabularx}{\textwidth}{>{\raggedright\arraybackslash}p{0.12\textwidth} X >{\raggedright\arraybackslash}p{0.18\textwidth}}
\toprule
\textbf{Round} & \textbf{Counterfactual Example} & \textbf{Classifier Prediction} \\
\midrule
\textbf{Original} 
&
Premise: A small girl with short brown hair is pushing a pink scooter over grass with brown leaves with trees in the background. \newline
Hypothesis: The girl is outside. 
& NEUTRAL \newline (Expected: Contradiction) \\
\midrule
\textbf{Round 1} 
&
Premise: A \hl{small} \hl{girl} with short brown hair is \hl{inside playing} with a \hl{pink scooter}. \newline
Hypothesis: The girl is outside.
& NEUTRAL \newline (Expected: Contradiction) \\
\midrule
\textbf{Round 2} 
&
Counterfactual: \emph{invalid output} 
& NEUTRAL \\
\midrule
\textbf{Round 3} 
&
Premise: A \hl{small} \hl{girl} with short brown hair is \hl{playing} inside with a \hl{pink scooter}. \newline
Hypothesis: The girl is outside.
& NEUTRAL \newline (Expected: Contradiction) \\
\midrule
\textbf{Round 4} 
&
Premise: A \hl{small} \hl{girl} with \hl{short} brown hair is \hl{indoors riding} \hl{a} \hl{pink} scooter. \newline
Hypothesis: The girl is outside.
& \textbf{ENTAILMENT} \newline (Expected: Contradiction) \\
\bottomrule
\end{tabularx}
\caption{Iterative counterfactual refinement on an \data{SNLI} Premise. Top-$k$ important words identified by AttnLRP are highlighted. \textit{i}\texttt{Flip} flips the label, but not always to the desired class.}
\label{tab:example1}
\end{table*}

\subsection{Example 2: Failed Flip} 

\noindent \textbf{Observation:} 
Table \ref{tab:example2} shows a common failure case: the generator tends to produce longer and more elaborate counterfactuals, 
but the label remains unchanged despite multiple refinement rounds.

\begin{table*}[h]
\centering
\renewcommand{\arraystretch}{1.25}
\footnotesize
\begin{tabularx}{\textwidth}{>{\raggedright\arraybackslash}p{0.12\textwidth} X >{\raggedright\arraybackslash}p{0.18\textwidth}}

\toprule
\textbf{Round} & \textbf{Counterfactual Example} & \textbf{Classifier Prediction} \\
\midrule
\textbf{Original} 
&
Premise: A ballerina standing on the toes of one foot with one arm up in the air and one leg lifted behind her. \newline
Hypothesis: The ballerina hates this. 
& CONTRADICTION \newline (Expected: Neutral) \\
\midrule
\textbf{Round 1} 
&
Premise: A ballerina stands \hl{gracefully} on the toes of one foot, her arm reaching towards the sky, while the other leg is lifted behind her, a look of \hl{intense} concentration on her face \hl{as} she \hl{perfectly captures} the \hl{beauty} and intricacy of the \hl{moment} in \hl{dance}, amidst a bustling cityscape outside the studio window, completely \hl{unaware} of the chaos beyond her focus. \newline
Hypothesis: The ballerina hates this.
& CONTRADICTION \newline (Expected: Neutral) \\
\midrule
\textbf{Round 2} 
&
Premise: A ballerina stands \hl{gracefully} on the toes of one foot, her arm reaching towards the sky, while the other leg is lifted behind her, a look of \hl{intense} concentration on her face as she \hl{perfectly captures} the \hl{beauty} and intricacy of the \hl{moment} in dance, happening during her first solo performance in front of a packed theater, completely \hl{unaware} of the standing ovation from the enthralled audience outside the proscenium arch. \newline
Hypothesis: The ballerina hates this.
& CONTRADICTION \newline (Expected: Neutral) \\
\midrule
\textbf{Round 3} 
&
Premise: A \hl{ballerina} stands \hl{gracefully} on the toes of one foot, her arm reaching towards the sky, while the other leg is lifted behind her, \hl{capturing} the \hl{serene} \hl{beauty} of a peaceful morning scene in the park, completely \hl{unaware} of the beautiful sunrise illuminating the clouds overhead. \newline
Hypothesis: The ballerina hates this.
& CONTRADICTION \newline (Expected: Neutral) \\
\midrule
\textbf{Round 4} 
&
Premise: A ballerina stands \hl{gracefully}, her arm elegantly raised, foot poised on the tip, as she prepares for a \hl{grand} performance on a bustling \hl{city} stage, her eyes \hl{filled} with \hl{determination} and focus. The bright lights above illuminate her, casting long shadows on the polished wooden floor. The audience's anticipation \hl{is} palpable in the silent, charged \hl{air} of the theatre. \newline
Hypothesis: The ballerina hates this.
& CONTRADICTION \newline (Expected: Neutral) \\
\midrule
\textbf{Round 5} 
&
Premise: A \hl{ballerina} stands \hl{gracefully} on a beach, one arm stretched upward towards the sky as if greeting the sun, the other hand resting \hl{lightly} on her hip, amidst a sea of beachgoers. The \hl{air} is salty and warm, and the sand is \hl{soft} beneath her feet. Around her, people are lounging, laughing, and enjoying the \hl{freedom} of the day. The \hl{city}'s skyline is visible in the distance, with skyscrapers gleaming against the azure sky. \newline
Hypothesis: The ballerina hates this.
& CONTRADICTION \newline (Expected: Neutral) \\
\bottomrule
\end{tabularx}
\caption{Iterative counterfactual refinement on an \data{SNLI} Premise. 
Top-$k$ important words identified by AttnLRP are highlighted.
Despite multiple refinements, the prediction remains CONTRADICTION instead of flipping to the expected NEUTRAL.}
\label{tab:example2}
\end{table*}

\subsection{Example 3: No Early Stopping (Flip-Flop)}  
\label{app:no_early_stop_example1}

\noindent \textbf{Observation:}  
Table \ref{tab:example3} shows  another failure pattern: even though the generator kept producing essentially the same counterfactual, the classifier’s prediction oscillated between \textbf{NEUTRAL} and \textbf{CONTRADICTION}, suggesting instability near the decision boundary.  

\begin{table*}[h]
\centering
\renewcommand{\arraystretch}{1.25}
\footnotesize
\begin{tabularx}{\textwidth}{>{\raggedright\arraybackslash}p{0.12\textwidth} X >{\raggedright\arraybackslash}p{0.18\textwidth}}
\toprule
\textbf{Round} & \textbf{Counterfactual Example} & \textbf{Classifier Prediction} \\
\midrule
\textbf{Original} 
&
Premise:  A man in a business suit silently meditating in front of the camera. \newline
Hypothesis: A business man is giving a heated interview for the local news.
& NEUTRAL \newline (Expected: Contradiction) \\

\midrule
\textbf{Round 1} 
&
Premise: A \hl{man} \hl{in} a \hl{business} \hl{suit} \hl{silently} \hl{meditating} in \hl{front} of the camera. \newline
Hypothesis: A business man is giving a heated interview for the local news.
& NEUTRAL \newline (Expected: Contradiction) \\
\midrule
\textbf{Round 2} 
&
Premise: A \hl{man} \hl{in} a \hl{business} \hl{suit} \hl{silently} \hl{meditating} in \hl{front} of the camera. \newline
Hypothesis: A business man is giving a heated interview for the local news.
& \textbf{CONTRADICTION} \newline (Expected: Contradiction) \\
\midrule
\textbf{Round 3} 
&
Premise: A \hl{man} \hl{in} a \hl{business} \hl{suit} \hl{silently} \hl{meditating} in \hl{front} of the camera. \newline
Hypothesis: A business man is giving a heated interview for the local news.
& NEUTRAL \newline (Expected: Contradiction) \\
\midrule
\textbf{Round 4} 
&
Premise: A \hl{man} \hl{in} a \hl{business} \hl{suit} \hl{silently} \hl{meditating} in \hl{front} of the camera. \newline
Hypothesis: A business man is giving a heated interview for the local news.
& \textbf{CONTRADICTION} \newline (Expected: Contradiction) \\
\midrule
\textbf{Round 5} 
&
Premise: A \hl{man} \hl{in} a \hl{business} \hl{suit} \hl{silently} \hl{meditating} in \hl{front} of the camera. \newline
Hypothesis: A business man is giving a heated interview for the local news.
& NEUTRAL \newline (Expected: Contradiction) \\
\bottomrule
\end{tabularx}
\caption{Iterative counterfactual refinement on an \data{SNLI} premise. 
Top-$k$ important words identified by AttnLRP are highlighted.
Despite producing nearly identical counterfactuals, the classifier oscillates between CONTRADICTION and NEUTRAL, illustrating instability near the decision boundary.}
\label{tab:example3}
\end{table*}

\subsection{Example 4: No Early Stopping (Flip-Flop)}  
\label{app:no_early_stop_example2}

\noindent \textbf{Observation:}  
Table \ref{tab:example4} illustrates a flip-flop failure pattern: the classifier initially flipped correctly to \textbf{CONTRADICTION}, 
but subsequent refinements pushed the prediction back to \textbf{NEUTRAL}, 
showing instability in the refinement loop.

\begin{table*}[h]
\centering
\renewcommand{\arraystretch}{1.25}
\footnotesize
\begin{tabularx}{\textwidth}{>{\raggedright\arraybackslash}p{0.12\textwidth} X >{\raggedright\arraybackslash}p{0.20\textwidth}}
\toprule
\textbf{Round} & \textbf{Counterfactual Example} & \textbf{Prediction (Expected)} \\
\midrule
\textbf{Original} 
&
Premise: Man in robe making designs on dirt ground. \newline
Hypothesis: A man making designs in the dirt outside. 
& NEUTRAL  \\
\midrule
\textbf{Round 1} 
&
Hypothesis: A man making \hl{designs} in the \hl{dirt} \hl{outdoors}. 
& NEUTRAL (Expected: Contradiction) \\
\midrule
\textbf{Round 2} 
&
Hypothesis: A man is \hl{carefully} \hl{cleaning} a spotless carpet \hl{indoors}. 
& \textbf{CONTRADICTION} (Expected: Contradiction) \\
\midrule
\textbf{Round 3} 
&
Hypothesis: A man is carefully \hl{cleaning} a spotless carpet indoors \hl{while} hl{there's} a man in a robe \hl{making} designs on a \hl{dirt} ground \hl{outside}. 
& NEUTRAL (Expected: Contradiction) \\
\midrule
\textbf{Round 4} 
&
Hypothesis: A man is meticulously \hl{cleaning} a spotless carpet indoors while simultaneously, there's a man in a robe making designs on a dirt ground outside, in \hl{total} \hl{contradiction} to the original \hl{scene} \hl{where} there's no \hl{mention} of indoor \hl{cleaning} and only a man making designs on dirt. 
& NEUTRAL (Expected: Contradiction) \\
\midrule
\textbf{Round 5} 
&
Hypothesis: A man indoors is vacuuming a spotless carpet while another man outside is \hl{making} \hl{intricate} designs on a dirt ground, representing a \hl{total} \hl{contradiction} \hl{as} it juxtaposes \hl{cleaning indoors} with design-making outdoors, and a spotless carpet with a dirt ground. 
& NEUTRAL (Expected: Contradiction) \\
\bottomrule
\end{tabularx}
\caption{Iterative counterfactual refinement on an \data{SNLI} Hypothesis. 
Top-$k$ important words identified by AttnLRP are highlighted.
The classifier briefly flipped to CONTRADICTION in Round 2 but reverted back to NEUTRAL.}
\label{tab:example4}
\end{table*}

\section{Prompts}
\label{app:prompts}


This section presents the full set of prompts employed in our counterfactual generation pipeline. The Base Counterfactual Prompt in Section~\ref{app:base-prompt} is used to initially generate a counterfactual example. The subsequent refinement stage then improves this initial counterfactual using one of our three feedback mechanisms: the Confidence-Based Refinement Prompt in Section~\ref{app:conf-prompt}, the Attribution-Based Refinement Prompt in Section~\ref{app:attr-prompt}, and the Natural-Language Feedback Refinement Prompt in Section~\ref{app:nl-prompt}.

\subsection{Base Counterfactual Prompt}\label{app:base-prompt}
\begin{verbatim}
A classifier has determined that the label of the following text is [ORIGINAL_LABEL]. Please flip it to [TARGET_LABEL] with minimal changes. Wrap your answer in <cf>...</cf>.
Original input:
[INPUT_TEXT]
Hint: [TASK-SPECIFIC HINT]
\end{verbatim}

\subsection{Confidence-based Refinement Prompt}\label{app:conf-prompt}
\begin{verbatim}
The classifier now predicts your text as [PRED_LABEL] ([CONF]% confidence), which is the desired label.
Your task is to minimize the edits compared to the original while preserving the current label: [PRED_LABEL].
Make the revision as close as possible to the original, but do not revert to the original label: [ORIGINAL_LABEL].
Wrap ONLY the final text in <cf>...</cf>.
Original input (classifier label [ORIGINAL_LABEL]):
[INPUT_TEXT]
Current counterfactual:
[CF_TEXT]
Hint: [TASK-SPECIFIC HINT]
\end{verbatim}

\subsection{Attribution-based Refinement Prompt}\label{app:attr-prompt}
\begin{verbatim}
The classifier now predicts your text as [PRED_LABEL], which is the desired label.
Key words influencing this prediction: [TOP_WORDS].
Your task is to minimize the edits compared to the original while preserving the current label: [PRED_LABEL].
Make the revision as close as possible to the original, but do not revert to the original label: [ORIGINAL_LABEL].
Wrap ONLY the final text in <cf>...</cf>.
Original input (classifier label [ORIGINAL_LABEL]):
[INPUT_TEXT]
Current counterfactual:
[CF_TEXT]
Hint: [TASK-SPECIFIC HINT]
\end{verbatim}

\subsection{Natural-Language Feedback Refinement Prompt}\label{app:nl-prompt}

Natural-language refinement first asks the model to generate NL feedback analyzing how to minimally achieve the target label. This feedback is then used to guide a second prompt that produces the counterfactual.

\subsubsection{Feedback Generation Prompt (NL Feedback)}
\begin{verbatim}
Analyze the current counterfactual and suggest improvements to achieve the target label [TARGET_LABEL].
You should make the smallest possible edits to the text while still achieving the target.
Wrap your reasoning inside <think>...</think>.

Original (classifier label [ORIGINAL_LABEL]):
[INPUT_TEXT]

Current counterfactual:
[CF_TEXT]
\end{verbatim}

\subsubsection{Counterfactual Editing Prompt (Using NL Feedback)}

\begin{verbatim}
Based on the feedback below, revise the text so that it flips to  [TARGET_LABEL].

Feedback:
[FEEDBACK_TEXT]

Wrap ONLY the final counterfactual in <cf>...</cf>.

Original text (classifier label: [ORIGINAL_LABEL]):
[INPUT_TEXT]

Current counterfactual:
[CF_TEXT]
\end{verbatim}

\clearpage
\onecolumn     
\begin{figure*}[p]  
\centering
\section{User Study Guidelines}
\label{sec:user_study_guidelines}

\begin{tcolorbox}[
  colback=gray!5,
  colframe=black!40,
  boxrule=0.5pt,
  arc=2pt,
  width=\textwidth,
  title=\textbf{User Study Guidelines}
]

\small

Dear participants,\\[1mm]

Thank you for participating in our user study. This study focuses on evaluating model-generated explanations, specifically counterfactual explanations (definitions and examples follow). We generate counterfactuals using three approaches in total, and you will evaluate 10 counterfactuals generated by each approach. You will subjectively rate each explanation based on the following three dimensions: \textbf{completeness}, \textbf{understandability}, and \textbf{cohesiveness}.\\[2mm]

\textbf{Counterfactual Example (CFE):} the smallest change to an input that would cause a model to output a different prediction.  
It is important to note that ``change in prediction'' refers to the model’s predicted label, not the ground truth label.\\

\textbf{Simple Example (Sentiment Classification):}
\vspace{-0.75\baselineskip}
\begin{itemize}
  \setlength{\itemsep}{1pt}
  \setlength{\parskip}{0pt}
  \setlength{\parsep}{0pt} 
  \item Original: ``The movie was boring.'' → Model predicts \textbf{negative}.
  \item CFE: ``The movie was great.'' → Model predicts \textbf{positive}.
\end{itemize}

In our questionnaire, the task is more complex: the model modifies news sentences from AG News, which has four categories: Business, Sports, World, Sci/Tech. You will manually rate CFEs using the following metrics. Higher scores indicate higher quality.\\

\textbf{1. Completeness (1–6):}  
The explanation is sufficient in explaining the outcome.
In other words, does the CFE provide enough information to clearly support the new predicted label?
\vspace{-0.75\baselineskip} 
\begin{itemize}
  \setlength{\itemsep}{1pt}
  \setlength{\parskip}{0pt}
  \setlength{\parsep}{0pt} 
  \item \textbf{1} – The CFE is not enough to justify the new label; explanation is insufficient  
  \item \textbf{3} – Partially justifies the new label; some key information is missing
  \item \textbf{6} – Fully justifies the new label; enough information to clearly support the prediction shift
\end{itemize}

\hspace{1em} We give two special cases to help you rate:
\vspace{-0.75\baselineskip} 
\begin{itemize}
  \setlength{\itemsep}{1pt}
  \setlength{\parskip}{0pt}
  \setlength{\parsep}{0pt} 
  \item \textbf{Case I}:  \\
    Original: ``The company reported lower profits this quarter.''  $\rightarrow$
 Model predicts Business. \\
  CFE:\hspace{0.55cm} ``The company reported lower profits this sports quarter.'' $\rightarrow$
 Model predicts Sports. \\
\textit{  We may rate this CFE a low completeness score (e.g., 1–2). Because the topic of CFE is still Business, though it did mention sports quarter.}

  \item \textbf{Case II}: \\
  Original: ``The president met with foreign leaders to discuss regional security.'' $\rightarrow$ Model predicts World.\\
    CFE:\hspace{0.55cm} ``The president met with foreign leaders to discuss new technology.'' $\rightarrow$ Model predicts Sci/Tech.\\
    \textit{We may rate this CFE a low completeness score (e.g., 1–2). The sentence still mainly describes political or diplomatic meetings; the small mention of ``technology'' is not enough to justify Sci/Tech.}
\end{itemize}

\textbf{2. Overall satisfaction (1–6):} This scenario effectively explains how to reach a different outcome.
\vspace{-0.75\baselineskip} 
\begin{itemize}
  \setlength{\itemsep}{1pt}
  \setlength{\parskip}{0pt}
  \setlength{\parsep}{0pt} 
  \item \textbf{1} – Does not explain how to reach a different outcome
  \item \textbf{3} – Explains it partially but lacks clarity
  \item \textbf{6} – Clearly and fully explains how to reach a different outcome
\end{itemize}

\hspace{1em}We give one special case to help you rate:
\vspace{-0.75\baselineskip} 
\begin{itemize}
  \item \textbf{Case I}:  \\
    Original: "Leaders met to discuss international security and trade sanctions."  $\rightarrow$
 Model predicts World. \\
  CFE:\hspace{0.55cm} ``Leaders met to discuss international sports security and trade sanctions.'' $\rightarrow$
 Model predicts World. \\
\textit{We may rate this CFE a low Overall Satisfaction score (e.g., 1–2), because the change from ``security'' to ``sports security'' is too minor to meaningfully alter the sentence’s topic.}
\end{itemize}

\textbf{3. Feasibility (1–6):} The actions suggested by the explanation are practical, realistic to implement and actionable.
\vspace{-0.75\baselineskip} 
\begin{itemize}
  \setlength{\itemsep}{1pt}
  \setlength{\parskip}{0pt}
  \setlength{\parsep}{0pt} 
  \item \textbf{1} – unrealistic or impossible
  \item \textbf{3} – partly feasible but unclear or difficult
  \item \textbf{6} – fully realistic and actionable
\end{itemize}

\hspace{1em}We give one special case to help you rate:
\vspace{-0.75\baselineskip} 
\begin{itemize}
  \item \textbf{Case I}:  \\
    Original: "Apple Inc. won an international innovation award."  $\rightarrow$
 Model predicts Business. \\
  CFE:\hspace{0.55cm} ``Apple Inc. won the World Cup'' $\rightarrow$
 Model predicts Sports. \\
\textit{We may rate this CFE a low score (e.g., 1–2), because the change is unrealistic. A company cannot win the World Cup.}
\end{itemize}

\end{tcolorbox}

\end{figure*}
\twocolumn  

\section{LLM-as-a-Judge Evaluation}
We adopt an LLM-as-a-judge protocol to assess the explanatory quality of counterfactual from three subjective perspectives: \textit{Completeness}, \textit{Overall Satisfaction}, and \textit{Feasibility}.

\subsection{Prompt}
\label{subsec::prompt_llm:judge}
The following prompt instructs the judge model to rate each CFE using integer scores from 1 to 6.
\begin{verbatim}
You are an evaluation model (LLM-as-a-judge). You will be given:
- Original text
- Counterfactual example (CFE): A minimally edited version of the input text that results in a change in the model’s prediction. 
- Original prediction and counterfactual prediction
- Whether the prediction flips (isFlip)

Your task is to score the CFE scenario’s explanatory quality on three metrics. Use integer scores from 1 to 6 (1 = lowest, 6 = highest).
Metrics:
1) Completeness (1-6): The explanation is sufficient in explaining the outcome.
2) Overall satisfaction (1-6): This scenario effectively explains how to reach a different outcome.
3) Feasibility (1-6): The actions suggested by the explanation are practical, realistic to implement and actionable.

Output MUST be exactly in the following format (three tags, integers only):
<completeness>score</completeness><satisfaction>score</satisfaction><feasibility>score</feasibility>

Do not output anything else.
\end{verbatim}

\subsection{Evaluation Across Judge Models}
\label{subsec::eval_llms}
Table~\ref{tab:judge_models_eval} reports the mean ratings for all counterfactuals across three judge models.

\begin{table}[htbp]
  \centering
  \resizebox{0.8\columnwidth}{!}{%
  \begin{tabular}{c|l|ccc|c}
    \toprule
    \textbf{Model} & \textbf{Method} & \textbf{Complet.} & \textbf{Overall Sat.} & \textbf{Feasib.} & \textbf{Kripp. \(\alpha\)} \\
    \midrule

    \multirow{3}{*}{%
      \rotatebox[origin=c]{90}{\parbox{1.8cm}{\centering \textbf{Gemma3}\\\textbf{-27B}}}%
    }
    & \cellcolor[HTML]{D8D8D8}CGG
      & \cellcolor[HTML]{D8D8D8}3.4
      & \cellcolor[HTML]{D8D8D8}3.4
      & \cellcolor[HTML]{D8D8D8}4.5
      & \multirow{3}{*}{0.649} \\
    & \cellcolor[HTML]{D8D8D8}FIZLE
      & \cellcolor[HTML]{D8D8D8}3.2
      & \cellcolor[HTML]{D8D8D8}3.2
      & \cellcolor[HTML]{D8D8D8}3.3
      &  \\
    & \cellcolor[HTML]{FFEACC}\textit{i}\texttt{Flip}-NL
      & \textbf{5.2}
      & \textbf{5.3}
      & \textbf{5.1}
      &  \\
    \midrule

    \multirow{3}{*}{%
      \rotatebox[origin=c]{90}{\parbox{1.8cm}{\centering \textbf{GPT-OSS}\\\textbf{-120B}}}%
    }
    & \cellcolor[HTML]{D8D8D8}CGG
      & \cellcolor[HTML]{D8D8D8}2.7
      & \cellcolor[HTML]{D8D8D8}2.6
      & \cellcolor[HTML]{D8D8D8}2.9
      & \multirow{3}{*}{0.666} \\
    & \cellcolor[HTML]{D8D8D8}FIZLE
      & \cellcolor[HTML]{D8D8D8}2.7
      & \cellcolor[HTML]{D8D8D8}2.7
      & \cellcolor[HTML]{D8D8D8}3.3
      &  \\
    & \cellcolor[HTML]{FFEACC}\textit{i}\texttt{Flip}-NL
      & \textbf{3.4}
      & \textbf{3.4}
      & \textbf{3.5}
      &  \\
    \midrule

    \multirow{3}{*}{%
      \rotatebox[origin=c]{90}{\parbox{1.8cm}{\centering \textbf{DeepSeek}\\\textbf{-R1}}}%
    }
    & \cellcolor[HTML]{D8D8D8}CGG
      & \cellcolor[HTML]{D8D8D8}3.5
      & \cellcolor[HTML]{D8D8D8}3.3
      & \cellcolor[HTML]{D8D8D8}4.4
      & \multirow{3}{*}{0.571} \\
    & \cellcolor[HTML]{D8D8D8}FIZLE
      & \cellcolor[HTML]{D8D8D8}3.4
      & \cellcolor[HTML]{D8D8D8}3.3
      & \cellcolor[HTML]{D8D8D8}4.8
      &  \\
    & \cellcolor[HTML]{FFEACC}\textit{i}\texttt{Flip}-NL
      & \textbf{5.0}
      & \textbf{4.7}
      & \textbf{4.6}
      &  \\
    \bottomrule
  \end{tabular}%
  }
  \caption{Evaluation results on \textit{Completeness}, \textit{Overall Satisfaction}, and \textit{Feasibility} across three judge models. Best results are bolded. Krippendorff's \(\alpha\) denotes Krippendorff's alpha measuring agreement between the judge model ratings and human ratings.}
  \label{tab:judge_models_eval}
\end{table}

\newpage

\section{Error Analysis}
\label{sec:error_analysis}
Subsection~\ref{subsec:incomplete_cfe} and \ref{subsec:unrealistic_cfe} discusses Incomplete and Unrealistic CFE Candidates, 
and Subsection~\ref{subsec:good_iter_refine} shows how iterative refinement can enhance the completeness and feasibility of CFE candidates.

\subsection{Incomplete CFE Candidates}
\label{subsec:incomplete_cfe}
As illustrated in Table~\ref{tab:error_analysis_incomplete}, a common failure case occurs when only partial entity mentions are edited, resulting in incomplete modifications that fail to sufficiently alter the model's predictions. 

\definecolor{scitech}{RGB}{220, 237, 200}
\definecolor{business}{RGB}{255, 240, 200}
\definecolor{world}{RGB}{200, 220, 255}
\definecolor{sport}{RGB}{255, 220, 220}

\definecolor{headergray}{gray}{0.95}

\begin{table}[htbp]
\centering
\small
\resizebox{\columnwidth}{!}{
\begin{tabular}{|c|c|c|}
\hline
\rowcolor{headergray}
\textbf{Dataset} & \multicolumn{2}{c|}{AG News} \\
\hline
\textbf{Method} & \multicolumn{2}{c|}{CGG} \\
\hline
\textbf{Original} & \multicolumn{2}{p{7cm}|}{Senate approves tax relief bill for manufacturers The Senate today passed a far-reaching, \$136 billion corporate tax package that cuts taxes for businesses ranging from film companies to bow-and-arrow makers while closing tax loopholes and bringing US exporters in line with\newline

\hfill\textbf{Prediction:}
\colorbox{business} {business}
} \\
\hline
\textbf{CFE} & \multicolumn{2}{p{7cm}|}{Senate approves \hl{research funding} bill for innovators The Senate today passed a far-reaching, \$136 billion \hl{scientific research} package that cuts \hl{funding} for \hl{projects} ranging from \hl{space companies to robotics makers} while closing \hl{research gaps} and bringing US \hl{scientists} in line with\newline

\hfill\textbf{Prediction:}
\colorbox{business} {business}
} \\
\hline
\textbf{Issue} & \multicolumn{2}{p{7cm}|}{Although many business-related entities are replaced with scientific terms, enough business cues remain and the overall prediction does not change.} \\
\hline
\end{tabular}
}
\caption{Error analysis for incomplete CFE candidates}
\label{tab:error_analysis_incomplete}
\end{table}

\subsection{Unnatural CFE Candidates}
\label{subsec:unrealistic_cfe}

As shown in Table~\ref{tab:error_analysis_unrealistic}, another common failure case is when modifications produce counterfactuals that are unnatural. The generated CFE candidates contain contexts that are inconsistent with real-world facts and scenarios.

\begin{table}[!h]
\centering
\small
\resizebox{\columnwidth}{!}{
\begin{tabular}{|c|c|c|}
\hline
\rowcolor{headergray}
\textbf{Dataset} & \multicolumn{2}{c|}{AG News} \\
\hline
\textbf{Method} & \multicolumn{2}{c|}{FIZLE} \\
\hline
\textbf{Original} & \multicolumn{2}{p{7cm}|}{IBM Claims Computing Crown (The Motley Fool) The Motley Fool - IBM (NYSE: IBM - News) has new bragging rights. Press reports indicate that the technology giant has created the world's fastest supercomputer two years after a Japanese computer claimed that title. \newline

\hfill\textbf{Prediction:}
\colorbox{scitech}{sci/tech}
} \\
\hline
\textbf{CFE} & \multicolumn{2}{p{7cm}|}{IBM Claims \hl{Tennis} Crown (The \hl{Sports} Fool) The \hl{Sports} Fool - IBM (NYSE: IBM - News) has new bragging rights. Press reports indicate that the \hl{tennis} giant has created the world's fastest \hl{serve} two years after a Japanese \hl{player} claimed that title. \newline

\hfill\textbf{Prediction:}
\colorbox{scitech}{sci/tech}
} \\
\hline
\textbf{Issue} & \multicolumn{2}{p{7cm}|}{The CFE unrealistically portrays IBM, a technology company, as a tennis giant, creating a scenario that conflicts with real world.} \\
\hline
\end{tabular}
}
\caption{Error analysis for unnatural CFE candidates.}
\label{tab:error_analysis_unrealistic}
\end{table}

\subsection{Per-annotator Human and LLM Ratings}
\label{subsec::example_user_study}

Table~\ref{tab:example_user_study1} presents two concrete examples that report detailed per-annotator ratings from a human user study (H1--H3) and an LLM-as-a-judge evaluation (L1--L3; \lm{Gemma3-27B}, \lm{GPT-OSS-120B}, and \lm{DeepSeek-R1}) on three dimensions: \textit{Completeness}, \textit{Overall Satisfaction}, and \textit{Feasibility}.

\subsection{Enhancing Completeness and Feasibility with Iterative Refinement}
\label{subsec:good_iter_refine}
As shown in Table~\ref{tab:error_analysis_refine}, iterative refinement helps address both incompleteness and unnatural counterfactual scenarios in CFE candidates.

\begin{table}[!h]
\centering
\small
\resizebox{\columnwidth}{!}{
\begin{tabular}{|c|c|c|}
\hline
\rowcolor{headergray}
\textbf{Dataset} & \multicolumn{2}{c|}{AG News} \\
\hline
\textbf{Method} & \multicolumn{2}{c|}{\textit{i}\texttt{Flip}} \\
\hline
\makecell{\\ \textbf{Original}} & \multicolumn{2}{p{7cm}|}{Gold Fields wins appeal to fight takeover Gold Fields Ltd. won an appeal on Friday in its battle to stave off a hostile \$7.1 billion takeover by Harmony Gold Mining Co. that would create the world \#39;s largest gold mining company.\newline

\hfill\textbf{Prediction:}
\colorbox{business} {business}
} \\
\hline
\makecell{\\ \textbf{CFE (a)} \\(\textit{Iter.} $\mathcal{K}=1$)} & \multicolumn{2}{p{7cm}|}{Gold Fields wins global appeal to address international concerns Gold Fields Ltd. won a global appeal on Friday in its effort to address international concerns by Harmony Gold Mining Co. that would impact global stability, sparking concerns at the United Nations and among nations worldwide.\newline

\hfill\textbf{Prediction:}
\colorbox{business} {business}
} \\
\hline
\makecell{\\ \textbf{CFE (b)} \\(\textit{Iter.} $\mathcal{K}=2$)}
  & \multicolumn{2}{p{7cm}|}{Gold Fields wins global appeal to address international concerns, sparking widespread discussion on the world stage about the global implications of mining operations and their potential impact on international stability, prompting concerns at the United Nations and among nations worldwide.\newline

\hfill\textbf{Prediction:}
\colorbox{world} {world}
} \\
\hline
\textbf{Observation} & \multicolumn{2}{p{7cm}|}{Although \textbf{CFE (a)} adds world-related context, it is still mainly about the legal appeal, so the model keeps the \textit{business} label and the claim about affecting global stability is not very realistic.\newline
After one round of iterative refinement, \textbf{CFE (b)} becomes more complete and realistic by highlighting the global debate around mining operations, and consequently the model prediction changes to \textit{world}.} \\
\hline
\end{tabular}
}
\caption{Error analysis for Iterative Refinement.}
\label{tab:error_analysis_refine}
\end{table}

\begin{table*}[!h]
\centering
\setlength{\tabcolsep}{6pt}
\renewcommand{\arraystretch}{1.18}

\begin{tabularx}{\textwidth}{|L|Y|}
\hline
\rowcolor{headergray} \textbf{Dataset} & \texttt{AG News} \\
\hline

\textbf{Original}
&
{\footnotesize\itshape
"Senate approves tax relief bill for manufacturers The Senate today passed a far-reaching, \$136 billion corporate tax package that cuts taxes for businesses ranging from film companies to bow-and-arrow makers while closing tax loopholes and bringing US exporters in line with "
}
\\
\hline

\textbf{CFE by CGG}
&
\begin{minipage}[t]{\linewidth}\vspace{0pt}
{\footnotesize\itshape
"Senate approves research funding bill for innovators The Senate today passed a far-reaching, \$136 billion scientific research package that cuts funding for projects ranging from space companies to robotics makers while closing research gaps and bringing US scientists in line with"
}
\par\noindent\rule{\linewidth}{0.4pt}\par

{\footnotesize\normalfont
\setlength{\tabcolsep}{4pt}
\renewcommand{\arraystretch}{1.08}

\begin{tabular}{@{}p{0.38\linewidth} p{0.60\linewidth}@{}}

\begin{tabular}[t]{@{}l l@{}}
\textbf{Orig. Pred} & {business} \\
\textbf{CFE Pred}   & {business} \\
\textbf{isFlip}     & Yes \\
\end{tabular}

&

\begin{tabular}[t]{@{}l p{0.52\linewidth}@{}}
\textbf{Completeness} &
\begin{tabular}[t]{@{}l r@{}}
H1--H3: 3/2/4 & L1--L3: 3/1/2
\end{tabular}
\\
\textbf{Overall Satisfaction} &
\begin{tabular}[t]{@{}l r@{}}
H1--H3: 2/2/3 & L1--L3: 3/1/2
\end{tabular}
\\
\textbf{Feasibility} &
\begin{tabular}[t]{@{}l r@{}}
H1--H3: 3/2/2 & L1--L3: 4/1/4
\end{tabular}
\\
\end{tabular}

\end{tabular}
}
\end{minipage}
\\
\hline

\textbf{CFE by FIZLE}
&
\begin{minipage}[t]{\linewidth}\vspace{0pt}
{\footnotesize\itshape
"Sports fans are approval of tax relief bill for sports manufacturers The Senate today passed a far-reaching, \$136 billion corporate tax package that cuts taxes for businesses ranging from sports equipment companies to athletic wear makers while closing tax loopholes and bringing US sports exporters in line with international standards, which is expected to boost the sports industry and create new jobs for athletes and sports enthusiasts."
}
\par\noindent\rule{\linewidth}{0.4pt}\par

{\footnotesize\normalfont
\setlength{\tabcolsep}{4pt}
\renewcommand{\arraystretch}{1.08}

\begin{tabular}{@{}p{0.38\linewidth} p{0.60\linewidth}@{}}

\begin{tabular}[t]{@{}l l@{}}
\textbf{Orig. Pred} & {business} \\
\textbf{CFE Pred}   & {sport} \\
\textbf{isFlip}     & No \\
\end{tabular}

&

\begin{tabular}[t]{@{}l p{0.52\linewidth}@{}}
\textbf{Completeness} &
\begin{tabular}[t]{@{}l r@{}}
H1--H3: 6/3/1 & L1--L3: 6/5/5
\end{tabular}
\\
\textbf{Overall Satisfaction} &
\begin{tabular}[t]{@{}l r@{}}
H1--H3: 3/3/5 & L1--L3: 6/5/5
\end{tabular}
\\
\textbf{Feasibility} &
\begin{tabular}[t]{@{}l r@{}}
H1--H3: 3/2/3 & L1--L3: 6/5/5
\end{tabular}
\\
\end{tabular}

\end{tabular}
}
\end{minipage}
\\
\hline

\textbf{CFE by \textit{i}\texttt{Flip}}
&
\begin{minipage}[t]{\linewidth}\vspace{0pt}
{\footnotesize\itshape
"Senate approves research funding bill for sci/tech innovators. The Senate today passed a far-reaching, \$136 billion research package that funds cutting-edge projects in areas like quantum computing, biotechnology, and advanced materials for companies ranging from artificial intelligence startups to renewable energy labs, while supporting collaborations with institutions like NASA, Google, and MIT, and enhancing the global competitiveness of US tech exporters."
}
\par\noindent\rule{\linewidth}{0.4pt}\par

{\footnotesize\normalfont
\setlength{\tabcolsep}{4pt}
\renewcommand{\arraystretch}{1.08}

\begin{tabular}{@{}p{0.38\linewidth} p{0.60\linewidth}@{}}

\begin{tabular}[t]{@{}l l@{}}
\textbf{Orig. Pred} & {business} \\
\textbf{CFE Pred}   & {sci/tech} \\
\textbf{isFlip}     & Yes \\
\end{tabular}

&

\begin{tabular}[t]{@{}l p{0.52\linewidth}@{}}
\textbf{Completeness} &
\begin{tabular}[t]{@{}l r@{}}
H1--H3: 5/5/6 & L1--L3: 6/5/6
\end{tabular}
\\
\textbf{Overall Satisfaction} &
\begin{tabular}[t]{@{}l r@{}}
H1--H3: 6/6/6 & L1--L3: 6/5/6
\end{tabular}
\\
\textbf{Feasibility} &
\begin{tabular}[t]{@{}l r@{}}
H1--H3: 6/5/6 & L1--L3: 6/4/6
\end{tabular}
\\
\end{tabular}

\end{tabular}
}
\end{minipage}
\\
\hline

\end{tabularx}

\caption{Detailed ratings from three human annotators (H1--H3) and three LLM judges (L1--L3), where L1=\lm{Gemma3-27B}, L2=\lm{GPT-OSS-120B}, and L3=\lm{DeepSeek-R1}.}
\label{tab:example_user_study1}
\end{table*}

\end{document}